\documentclass[pra,superscriptaddress,showpacs,amsmath,amssymb,longbibliography,nofootinbib,aps,10pt]{revtex4-2}%{article}%{revtex4-2}%,twocolumn

\renewcommand{\thesection}{\arabic{section}}
\renewcommand{\thesubsection}{\thesection.\arabic{subsection}}
\renewcommand{\thesubsubsection}{\thesubsection.\arabic{subsubsection}}
\renewcommand{\theparagraph}{\thesubsubsection.\arabic{paragraph}}
\renewcommand{\thesubparagraph}{\theparagraph.\arabic{subparagraph}}
\makeatletter
\renewcommand{\p@subsection}{}
\renewcommand{\p@subsubsection}{}
\renewcommand{\p@paragraph}{}
\renewcommand{\p@subparagraph}{}
\makeatother
\usepackage{titlesec}
\titleformat{\subparagraph}[runin]%
  {\normalfont\normalsize\itshape}%
  {\thesubparagraph.} % <-- Dot added directly after the number
  {1em}%
  {}%
\usepackage[super]{nth}%For 1st, 2nd, 3rd, 4th...

\usepackage{graphicx, color}% Include figure files
\usepackage{float}
\usepackage{dcolumn}% Align table columns on decimal point
\usepackage{bm}% bold math
\usepackage{epstopdf}
\usepackage{physics}
\usepackage[dvipsnames]{xcolor}
\usepackage{subfigure}
\usepackage{pifont}

\definecolor{orange}{rgb}{1,0.5,0}
\definecolor{goodgreen}{rgb}{0.1,0.5,0}
\definecolor{goodred}{rgb}{0.7,0,0}
\usepackage[colorlinks,urlcolor=goodgreen,citecolor=blue,linkcolor=goodred]{hyperref}
\usepackage{verbatim}

\usepackage{array}%For charts
\newcolumntype{P}[1]{>{\centering\arraybackslash}m{#1}}
\usepackage{multirow}%For charts with multirow
\usepackage{ amssymb }%For the L of luminosity
\usepackage{svrsymbols}%For neutrinos
\usepackage{appendix}
\usepackage{longtable}

\usepackage{makecell}%For cells of charts

\usepackage{tikz}
\usetikzlibrary{arrows.meta, positioning, fit}

\usepackage{titlesec}
\usepackage{etoolbox}
\preto\appendices{%
  \titleformat{\section}
    {\normalfont\large\bfseries\centering}{\thesection}{1em}{}%
}

\graphicspath{ {./images/} }
\newcommand{\numbertrainingsignalsnumeric}{17}
\newcommand{\numbertrainingbackgroundsnumeric}{94}
\newcommand{\numbergeneralisationsignalsnumeric}{8}
\newcommand{\numbergeneralisationbackgroundsnumeric}{42}
\newcommand{\numberCPconjugates}{three}
\newcommand{\rewardthreshold}{0.05}
\newcommand{\lowestrewardln}{0.052}
\newcommand{\lowestreward}{0.053}
\newcommand{\highestrewardln}{2.588}
\newcommand{\highestreward}{12.303}
\newcommand{\lowestnumberbackgrounds}{1}
\newcommand{\highestnumberbackgrounds}{15}
\newcommand{\GAperformance}{87}
\newcommand{\trainingperformance}{94}
\newcommand{\generalisationperformance}{40.5}
\newcommand{\finetunedtrainingperformance}{94}
\newcommand{\finetunedgeneralisationperformance}{42}

\begin{document}
\title{Leveraging Reinforcement Learning, Genetic Algorithms and Transformers for background determination in particle physics}

\begin{abstract}

Experimental studies of beauty hadron decays face significant challenges due to a wide range of backgrounds arising from the numerous possible decay channels with similar final states. For a particular signal decay, the process for ascertaining the most relevant background processes necessitates a detailed analysis of final state particles, potential misidentifications, and kinematic overlaps, which, due to computational limitations, is restricted to the simulation of only the most relevant backgrounds. Moreover, this process typically relies on the physicist’s intuition and expertise, as no systematic method exists.\\

This paper has two primary goals. First, from a particle physics perspective, we present a novel approach that utilises Reinforcement Learning (RL) to overcome the aforementioned challenges by systematically determining the critical backgrounds affecting beauty hadron decay measurements. While beauty hadron physics serves as the case study in this work, the proposed strategy is broadly adaptable to other types of particle physics measurements. Second, from a Machine Learning perspective, we introduce a novel algorithm which exploits the synergy between RL and Genetic Algorithms (GAs) for environments with highly sparse rewards and a large trajectory space. This strategy leverages GAs to efficiently explore the trajectory space and identify successful trajectories, which are used to guide the RL agent's training. Our method also incorporates a transformer architecture for the RL agent to process token sequences that represent particle decays. 

\end{abstract}

\author{Guillermo Hijano Mendizabal}
\affiliation{Physik-Institut, Universität Zürich, 8057 Zurich, Switzerland}

\author{Davide Lancierini}
\affiliation{Imperial College London, London, United Kingdom}

\author{Alex Marshall}
\affiliation{H.H. Wills Physics Laboratory, University of Bristol, Bristol, United Kingdom}

\author{Andrea Mauri}
\affiliation{Imperial College London, London, United Kingdom}

\author{Patrick Haworth Owen}
\affiliation{Physik-Institut, Universität Zürich, 8057 Zurich, Switzerland}

\author{Mitesh Patel}
\affiliation{Imperial College London, London, United Kingdom}

\author{Konstantinos Petridis}
\affiliation{H.H. Wills Physics Laboratory, University of Bristol, Bristol, United Kingdom}

\author{Shah Rukh Qasim}
\affiliation{Physik-Institut, Universität Zürich, 8057 Zurich, Switzerland}

\author{Nicola Serra}
\affiliation{Physik-Institut, Universität Zürich, 8057 Zurich, Switzerland}

\author{William Sutcliffe}
\affiliation{Physik-Institut, Universität Zürich, 8057 Zurich, Switzerland}

\author{Hanae Tilquin}
\affiliation{Imperial College London, London, United Kingdom}

\maketitle

\section{Introduction}

In particle physics, measurements are often framed in terms of a signal and its accompanying backgrounds. The signal corresponds to the specific process under study, such as the production of a new particle or a rare interaction, while backgrounds are all other processes that can mimic the same experimental signature but are not of direct interest. An important area is heavy flavor physics, where beauty hadron decays serve as a case study for precision measurements and searches for rare processes to test the Standard Model and probe for signs of new physics, with experiments such as LHCb \cite{LHCb} and B-factories like Belle II \cite{Belle2} playing a leading role. Beauty hadrons can decay into a large number of final states, often through cascades involving several intermediate particles, and given the vast number of possible decay channels and their overlapping kinematic signatures, identifying the most relevant backgrounds for a given signal process under study is both critical and highly non-trivial.\\

To date, there exists no general, systematic method that, given an arbitrary signal decay, can automatically determine the set of relevant backgrounds. In the absence of such a method, background determination relies heavily on the physicist’s expertise. This involves evaluating the final-state particle composition, estimating possible misidentifications, and considering kinematic overlaps. Moreover, due to the significant computational cost associated with generating and simulating events, it is only feasible to simulate the subset of backgrounds considered most relevant. This constraint makes the accuracy of background selection essential, as failing to account for a relevant background not only biases the measurement but also complicates the physical interpretation of the results, creating uncertainty on whether the observed effect originates from genuine signs of new physics, incomplete background modeling, or detector or physics mismodelling. This highlights the importance of having a systematic and reliable method to identify all relevant backgrounds, ensuring that measurements are both accurate and correctly interpreted. Furthermore, an algorithmic approach would substantially accelerate the workflow by reducing the manual effort required to determine background processes, while simultaneously mitigating human error.\\

In this work, we present a novel algorithm which exploits the synergy between RL and GAs to determine the set of relevant backgrounds for a given signal decay. This RL-GA synergy is of interest not only for particle physics applications but also from an RL perspective, providing an effective solution for environments with a highly sparse reward configuration and a large trajectory space, which are particularly problematic for RL algorithms. The core of the algorithm is an RL agent trained to construct token sequences representing background processes. Complementarily, we utilise GAs to identify successful trajectories in the environment, helping to overcome the aforementioned challenges by guiding the training of the agent. Additionally, to process decay sequences, we employ a transformer architecture \cite{transformer}, which excels at capturing complex dependencies across the entire sequence.\\

\section{Related works}

Recent years have seen a growing interest in leveraging advanced Artificial Intelligence (AI) methods to address complex problems in particle physics, with RL, transformers, and GAs emerging as prominent tools. RL has found important applications in the field of particle physics. Examples include instrument design \cite{instrument}, particle tracking \cite{RLTracking}, and particle physics model building \cite{model_building,model_building2,Satsuki1,Satsuki2}. In our prior work, we introduced the DL Advocate framework \cite{DLAdvocate}, which applies RL to quantitatively assess the impact of hidden systematic uncertainties in precision measurements. Concurrently, the field of RL has witnessed significant algorithmic developments. A particularly relevant advancement is the AlphaZero algorithm \cite{AlphaZero}, which outperformed world champion programs in the games of chess, shogi and Go. This algorithm starts the learning process with no domain knowledge except the game rules, and employs self-play to generate the training data. It combines deep neural networks with Monte Carlo Tree Search (MCTS) \cite{MCTS1,MCTS2,MCTS3} to evaluate game positions and possible moves. This strategy facilitates the discovery of promising action trajectories, which is particularly valuable in environments with large trajectory spaces and delayed rewards, where feedback is only obtained at the end of an episode, as in the problem addressed in this work. While AlphaZero has seen limited use in physics so far, recent works have begun to explore its application, for instance, in symbolic regression to find analytical methods within the field \cite{symbolic_regression}, and in the optimisation of quantum dynamics \cite{quantum_dynamics}. However, to the best of our knowledge, the present work constitutes the first application of the AlphaZero algorithm within the domain of particle physics.\\

In parallel, transformer architectures, first introduced in the context of sequence modeling \cite{transformer} and later popularised through large-scale natural language processing models such as GPT-2 and GPT-3 \cite{GPT2,GPT3}, have been successfully adapted to various particle physics tasks. For example, transformers have been used for jet tagging \cite{tagging1, tagging2}. Additionally, they have proven effective in particle track reconstruction \cite{track_reconstruction,track_reconstruction2}. In the problem addressed in this work, decays are represented as token sequences. Therefore, we chose a transformer architecture for the agent due to its proven effectiveness in modeling sequential data and capturing contextual relationships between tokens.\\

Additionally, GAs have found extensive application in particle physics for optimisation and search problems, including detector setup optimisation \cite{detection_setup_optimisation} and efficient exploration of parameter spaces in particle physics models \cite{efficient_exploration, efficient_exploration2,efficient_exploration3}. GAs are optimization techniques inspired by natural evolution and selection. A population of candidate solutions, represented by their genes, evolves through selection, crossover, and mutation across generations, gradually converging toward optimal solutions. This Evolutionary Algorithm (EA) demonstrates strong robustness against local optima, and excels in problems that are not well-suited to traditional optimization techniques, for example, when gradient-based methods are inapplicable due to non-differentiable objective functions, as is the case in this work.\\

Beyond their standalone applications, EAs have also been successfully combined with RL algorithms in broader AI research. Such hybrid approaches have often been employed to optimise the agent’s policy network weights. For instance, Evolutionary Reinforcement Learning (ERL) \cite{ERL} is a hybrid algorithm that leverages the policies in an EA's population to generate diverse experience for training an RL agent, while periodically reinserting the agent into the population to inject gradient-based information into the EA. Another example is Deep Neuroevolution \cite{DeepNeuroevolution}, an approach that applies GAs to directly evolve the weights of deep neural networks, and which has demonstrated competitive performance as an alternative to gradient-based RL methods in training agents for challenging problems.\\

The approach most closely related to the one proposed in this work is Go-Explore \cite{Go_Explore}, an RL algorithm designed to address hard-exploration problems which operates in two phases: in Phase 1, the agent systematically explores the environment by returning to previously visited states stored in an archive and expanding from them to discover high-performing trajectories. In Phase 2, these trajectories are used to train a robust policy via Imitation Learning (IL). IL is a technique in which an agent learns to perform tasks by mimicking expert demonstrations, making it particularly valuable in complex environments where manually programming behaviour or designing an appropriate reward function is challenging \cite{ImitationLearning}.\\

In this work, we propose a novel algorithm which exploits the RL-GA synergy to address the challenges of a large trajectory space and highly sparse rewards. Specifically, GAs are employed to identify successful trajectories in the environment, which are subsequently utilised as high quality training data for the RL agent. Unlike Go-Explore, where high-performing trajectories are obtained through simple policies such as random action sequences, we leverage GAs in Phase 1 of the algorithm because of their effective exploration capabilities and their robustness to convergence in local optima.\\ 

Furthermore, our method is more suitable for environments with purely terminal rewards. In Go-Explore, a state is stored in the archive either when it is ``interestingly'' \footnote{Go-Explore employs a lower-dimensionality space in which similar states are conflated, but meaningfully different states are not.} different (novelty) or when a higher cumulative reward than an existing trajectory to the same state is achieved. In terminal-reward environments, this ``better-trajectory'' criterion offers little guidance before a terminal trajectory is found, because all partial trajectories have identical return of zero. Moreover, a terminal state itself cannot be expanded further, so the discovery of new solutions will only benefit indirectly from storing these trajectories in the archive: by restarting and branching from earlier archived states that lead up to it. Thus, archive growth is driven primarily by novelty, and the cumulative-reward signal is only intermittently informative in this setting. In contrast, in our approach, where GA individuals represent trajectories, this is not the case: through crossover and mutation, new successful trajectories can be discovered directly from already known trajectories with positive reward, allowing the algorithm to exploit previously found solutions more effectively.\\

Another key difference between the approach presented in this work and Go-Explore is that the latter uses one or more high performing trajectories discovered during Phase 1 to robustify a policy via IL, so that the agent can handle stochastic environments while reproducing the discovered behaviors. More specifically, they propose the use of Backward Algorithm \cite{backward_algorithm}, which provides a curriculum over trajectory segments, allowing the policy to master later portions first and thus robustly handle stochastic transitions throughout the entire sequence. In contrast, in this work we propose Policy Gradient with Supervised Updates (PGSU) for Phase 2. PGSU is a hybrid approach that combines Policy Gradient RL with IL, where the IL component corresponds to supervised learning from expert demonstrations. This is often done by incorporating a supervised imitation loss to the policy gradient loss: 

\begin{equation}\label{PGwSU}
    \mathcal{L}_{\text{total}} = \mathcal{L}_{\text{PG}} + \lambda \cdot \mathcal{L}_{\text{supervised}}
\end{equation}

The reason behind the choice of PGSU is the change of perspective in the background finder problem: the agent is not trained to play the best possible trajectory, but to learn how to generate all successful sequences (that is, all relevant backgrounds) for a given signal. While the Backward Algorithm tends to lock the policy toward the demonstrations if no additional exploration is injected, PGSU, combined with other strategies to enhance diversity further such as sampling from a softened MCTS policy distribution, encourages the policy's probability mass to spread over many good solutions. This strategy allows the agent to explore solutions beyond the expert demonstrations\footnote{``Expert demonstrations'' in this work correspond to algorithmically discovered high-performing sequences from Phase 1, rather than human-generated demonstrations.}, a crucial feature as the GA may fail to identify all relevant backgrounds for the training signals.\\

Additionally, we studied an alternative to PGSU for Phase 2, which we named Periodic Expert Guidance (PEG), consisting of adjusting the action masking so that the agent is forced to follow expert trajectories during certain episodes of training. Employing this strategy only for a fraction of episodes allows the agent to explore beyond these trajectories and achieve beyond-expert performance\footnote{In this work, performance is measured as the number of backgrounds correctly determined.}.\\

In the literature, other synergies using GAs to optimise trajectories have also been proposed. For example, the Evolutionary Augmentation Mechanism (EAM) \cite{GA_RL_paper2} generates solutions for combinatorial optimisation problems, refines them using GAs, and reinjects them into policy training. Additionally, during the preparation of this manuscript, we became aware of \cite{GA_RL_paper}, which independently explores an approach similar to ours, in which GA-generated expert demonstrations are used to enhance policy learning. The authors considered two approaches: one involved training a Deep Q-Network (DQN) with an experience replay buffer filled with expert demonstrations, and the other training a neural network through Behavioural Cloning (BC), a common IL approach that directly trains a policy via supervised learning on observation–action pairs from expert demonstrations, which was then used to initialise a Proximal Policy Optimization (PPO) agent. Our work was conducted in parallel and focuses on utilising the GA-discovered expert trajectories to apply PGSU.\\

\section{Methodology}

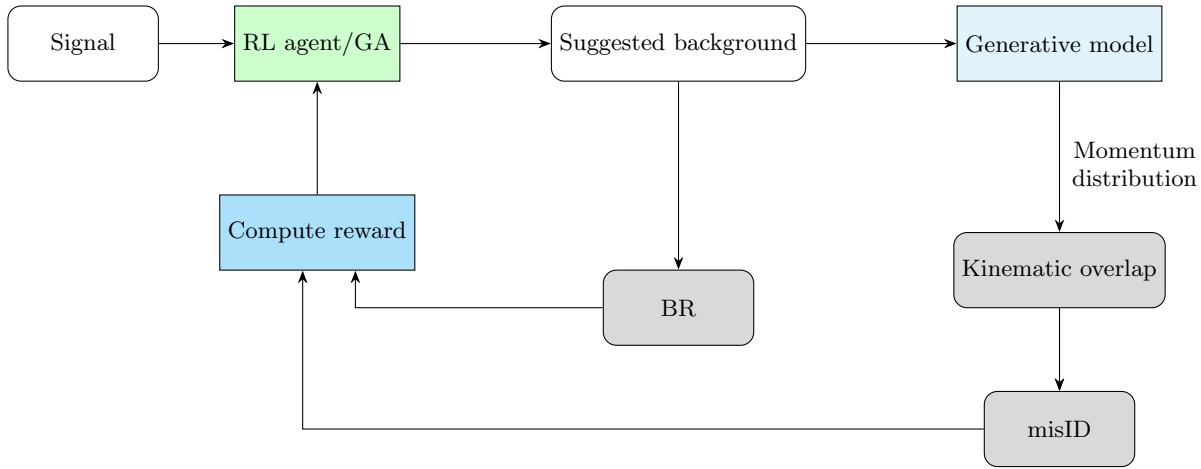
\begin{figure}[H]
\centering
\begin{tikzpicture}[
    block/.style = {rectangle, draw, minimum width=2cm, minimum height=1cm, align=center},
    >=Stealth
]
% Nodes
\node[block, rounded corners=5pt] (B1) {Signal};
\node[block, right=1cm of B1, fill=green!20] (B2) {RL agent/GA};
\node[block, right=2cm of B2, rounded corners=5pt] (B3) {Suggested background};
\node[block, right=2cm of B3, fill=cyan!10] (B4) {Generative model};
\node[block, below=2cm of B4, fill=gray!30, rounded corners=5pt] (B5) {Kinematic overlap};
\node[block, below=2.5cm of B3, fill=gray!30, rounded corners=5pt] (B6) {BR};
\node[block, below=1.1cm of B5, fill=gray!30, rounded corners=5pt] (B7) {misID};
\node[block, below=1.5cm of B2, fill=cyan!30] (B8) {Compute reward};
% Arrows
\draw[->] (B1) -- (B2);
\draw[->] (B2) -- (B3);
\draw[->] (B3) -- (B4);
\draw[->] (B4) -- node[midway, above, xshift=1cm, yshift=-0.5cm, align=center] {Momentum \\ distribution} (B5);
\draw[->] (B3) -- (B6);
\draw[->] (B5) -- (B7);
\draw[->] (B5) -- (B7);
\draw[->] (B7) -| ([xshift=-1cm]B8);
\draw[->] (B8) -- (B2);
\draw[->] (B6) -| ([xshift=1cm]B8);
\end{tikzpicture}
\caption{Diagram showing the workflow of the algorithm.}
\label{GlobalPicture}
\end{figure}

As previously described, in this work we propose an AI-driven method to systematically determine the critical backgrounds affecting a given signal decay. The central component of the algorithm is an RL agent responsible for determining the backgrounds. Additionally, we employ GAs to efficiently explore the trajectory space and identify relevant backgrounds, which serve as expert demonstrations to guide the agent's training. The core workflow of the algorithm is shown in Figure \ref{GlobalPicture}. A signal is specified by a user, for which backgrounds are suggested by the \textit{background finder agent}. Associated to each background, a reward is defined to describe how significant the background is for the given signal. Consequently, the agent's objective is to identify the backgrounds that maximise this reward.\\

%%%%%%%%%%%%%%%%%%%%%%%%%%%%%%%%%%%%%%%%%
%%%%%%%%%%%%%%%%%%%%%%%%%%%%%%%%%%%%%%%%%

The reward function (see Equation \ref{reward_definition}) accounts for several factors: the BRs of the signal $BR_{s}$ and the background $BR_{b}$ (to incorporate the probabilities of these certain decays occurring into the equation), a $\mathcal{M}$ factor associated to possible misidentifications of particles, and a factor $\mathcal{K}$ that quantifies the kinematic overlap between signal and background. These factors can be extracted as follows: in particular, the Particle Data Group (PDG) \cite{PDG} provides a database compiling information on measured decays including their BR\footnote{For the experiments in this work, a set of decays has been considered for training and testing the algorithm. However, the algorithm is designed with a broader use in mind, in order to account for all potential backgrounds that might arise for a given signal.}. In addition, the kinematic overlap factor could be obtained by training a generative model to estimate the momentum distribution of the daughter particles. Following the approach in \cite{Heterogeneous_GNNs}, where a heterogeneous Graph Neural Network (GNN) is employed for replacing detector simulation, this distribution could then be used to compute the overlap between signal and background detector responses.

\begin{equation}\label{reward_definition}
Reward = \frac{BR_{b}}{BR_{s}} \cdot \mathcal{M} \cdot \mathcal{K}
\end{equation}

A straightforward but impractical approach to extract all relevant backgrounds for a given signal would involve recursively iterating over the Particle Data Group (PDG) database using nested loops to construct all possible decay chains. This method is not computationally feasible due to the vast number of decays. Furthermore, the fact that the PDG is incomplete also means that this approach is unable to capture all possible backgrounds. Consequently, a more sophisticated strategy like the one proposed in this work is required.\\

\subsection{RL for background determination}\label{RL_background_determination}

The task of determining the most relevant background processes can be effectively addressed using RL. In this context, an RL agent is trained on a set of signal processes to learn associations with their corresponding backgrounds. Once trained, its generalisation ability permits it to predict the relevant backgrounds for previously unseen signal processes. In the approach explored in this work, the environment state is represented as a sequence of tokens encoding both the signal and the constructed background. The agent's actions correspond to selecting the next token to append to the sequence. The initial state encodes just the signal, and the agent iteratively constructs the background, adding one token per environment step.\\

\subsubsection{Agent architecture}

We adopt a transformer architecture \cite{transformer} for the agent due to its effectiveness in modeling sequential data. Although decay chains are naturally graph-structured, the agent's task of generating candidate backgrounds is framed as an autoregressive decision process, constructing the decay chain one token at a time. For this reason, a transformer architecture is more suitable than a Graph Neural Network (GNN). At inference time, the token sequence representing the environment state is partitioned into two segments: the signal-related tokens are provided as input to the encoder, while the background-related tokens are supplied to the decoder. Following the AlphaZero framework \cite{AlphaZero}, the model consists of two output heads: a policy head, which predicts the probability distribution over possible tokens for a given state, and a value head, which estimates the expected return from that state. In our implementation, both heads share a common transformer backbone before branching into their respective output layers.\\

\subsubsection{Tokenisation}

\begin{figure}[H]
  \centering
  \includegraphics[width=\textwidth]{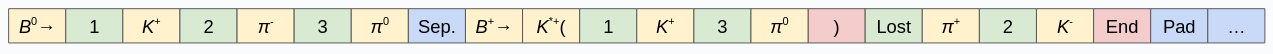}
  \caption{Example token sequence representing a state of the RL environment. Tokens have been highlighted in colors: particle tokens in yellow, tokens referencing the next element in the sequence in green, ``)'' and ``End'' tokens in red, and tokens not representing actions in blue. The signal is $B^0\to K^+  \pi^- \pi^0$ and the background generated by the agent $B^+\to K^{*+}( \ K^+ \pi^0 ) \ \pi^+ K^-$.}
  \label{State}
\end{figure}

The tokenisation utilised in this work is described next: one token is employed to represent each particle. For mother particles, the token encapsulates both the particle and the decay arrow, as separating them into distinct tokens would be redundant (each mother particle is always followed by a decay arrow). Similarly, tokens representing intermediate particles incorporate the opening parenthesis. Additionally, a set of index tokens is introduced. These tokens reference the final state particle immediately following them in the sequence, and allow for comparisons between corresponding particles in the signal and background. For example, in Figure \ref{State}, the particle following index token ``2'' is $\pi^-$ in the signal and $K^-$ in the background, indicating a $K^- \to \pi^-$ misidentification.\\

In addition, we incorporated a ``Lost'' token to indicate that the particle represented by the subsequent token was not reconstructed and does not correspond to a signal particle. A separate token is also employed to represent the closing parenthesis of an intermediate resonance. Furthermore, an ``End'' token is included to indicate the termination of the generated background sequence. All of these tokens constitute valid actions that the agent may select during the generation process. However, two additional tokens which do not correspond to agent actions are incorporated: The ``Sep.'' token marks the boundary between signal and background tokens, and also serves as the start token for the decoder input of the transformer-based agent. Lastly, a ``Pad'' token is used to denote empty or unassigned positions within the token sequence.\\

The proposed tokenisation allows for efficient handling of intermediate resonances, particle misidentifications and partially reconstructed backgrounds.

\subsubsection{Action masking}

Without action masking, the size of the trajectory space in the described RL problem is upper-bounded by $(N_A)^{S_L}$, where $N_A$ denotes the number of actions and $S_L$ the maximum sequence length allowed to represent a background. This results in an intractably large space of possible trajectories, except in extremely simplified scenarios. However, by employing action masking, which filters out invalid actions at each decision step, the effective trajectory space can be substantially reduced. This is possible due to the structured nature of the tokenization scheme and constraints imposed by physical laws, which allow for the elimination of invalid or redundant actions. For instance, when generating an intermediate resonance, the closing parenthesis token is masked unless charge, lepton flavour, and other conservation laws are satisfied. Similarly, the ``End'' token is masked unless the number of intermediate particle tokens in the sequence matches the number of closing parenthesis, ensuring syntactic consistency. Action masking not only enhances computational efficiency but also improves the agent’s exploration and learning capabilities.

\subsubsection{RL algorithm}

The RL problem described presents several key challenges: despite the application of action masking, the trajectory space remains large, contributing to the complexity of exploration. Additionally, the reward is only provided at the end of the episode, complicating the learning process of the agent, as it receives no immediate feedback when selecting an action. Moreover, the reward structure is highly sparse: only a small fraction of sequences correspond to valid background processes, making successful trajectories rare. In this context, the RL algorithm that we chose is AlphaZero \cite{AlphaZero}, which has demonstrated its effectiveness in games like chess, shogi and Go, all of which involve large trajectory spaces and delayed rewards. The algorithm is originally designed for two-player adversarial games. In this work, however, we adapt an open-source AlphaZero implementation \cite{AlphaZero_implementation} to operate in a single-player setting, specifically for solving the background finder problem. The traditional AlphaZero loss, defined over the entire set of states in the self-play dataset $\mathcal{D}$, is given by Equation \ref{alphazero_loss}.

\begin{equation}\label{alphazero_loss}
    \mathcal{L}_{\mathrm{AZ}}=\mathbb{E}_{s \sim \mathcal{D}}\left[(R-v_\theta(s))^2-\pi^\intercal \log p_\theta(\cdot \mid s)\right] +c||\theta||^2
\end{equation}

Here, $s$ denotes a state, $v_\theta(s)$ the network's prediction of the expected outcome for state $s$, $R$ the episode outcome, $p_\theta(\cdot \mid s)$ the model's predicted action probabilities for state $s$, and $\pi$ the policy obtained from MCTS. The value loss measures how accurately the model predicts the final episode outcome, while the policy loss encourages the network's policy head to match the policy obtained from MCTS. Additionally, the term $c||\theta||^2$ denotes an $L_2$ regularization term applied to the network parameters, which prevents overfitting and stabilises training.\\

\subsubsection{Synergy of RL and GAs}

In addition to the challenges of a large trajectory space and the absence of intermediate rewards that games like chess, shogi and Go face, the background finder problem meets the additional complication of a highly sparse reward configuration \footnote{In games such as chess, shogi and Go the terminal reward is typically non-zero: a win corresponds to a positive outcome, a loss to a negative outcome, and a draw to zero. In contrast, in the background finder problem, the majority of trajectories terminate without producing physically valid background processes, resulting in a reward of zero (or negative if penalties are applied).}. This constitutes a major problem in RL, as the agent struggles to find action configurations that lead to episodes from which useful information can be learnt.\\

To address this, we propose a hybrid strategy that leverages GAs to explore and identify successful trajectories in the environment (see Section \ref{GAs}), which are subsequently employed as expert demonstrations to guide the training of the RL agent. More specifically, we apply PGSU, extending the AlphaZero loss by incorporating a supervised term, where trajectory losses are weighted by reward to emphasise backgrounds of greater significance:

\begin{equation}\label{supervised_loss_equation}
    \mathcal{L}_{\mathrm{supervised}}(\tau)=\sum_{i=1}^{N_\tau} \frac{R_i}{T_i} \sum_{t=1}^{T_i} -\log p_\theta(a_t^{(i)}|s_t)
\end{equation}

In Equation \ref{supervised_loss_equation}, $i$ indexes an expert trajectory, and $N_\tau$ denotes the number of expert trajectories associated with the trajectory $\tau$, which depends on the corresponding training signal. $T_i$ is the number of timesteps for which the agent followed the expert trajectory $i$, $R_i$ the reward associated with trajectory $i$, $t$ a timestep, $p_\theta$ the policy head's predicted probability (under parameters $\theta$), $a_t^{(i)}$ the expert action at timestep $t$ for trajectory $i$, and $s_t$ the corresponding state. The negative log-likelihood penalises the model whenever it assigns low probability to expert actions. This supervised loss is applied only to states along expert trajectories, and the loss of each trajectory is normalised by the number of states for which the agent followed the trajectory, rather than the full trajectory length. This ensures that the loss decreases as training progresses: without this normalisation, the loss could initially increase as the agent begins to follow longer portions of expert trajectories, since more terms would be added to the loss. Finally, it is important to note that GAs may not discover all relevant backgrounds for the training signals. For this reason, we adopt the PGSU approach rather than BC, as the former enables exploration beyond expert demonstrations and allows achieving beyond-expert performance.\\

The supervised loss in Equation \ref{supervised_loss_equation} is defined over a single trajectory $\tau$. Equation \ref{total_loss} specifies the total loss, which corresponds to the sum of the traditional AlphaZero loss \cite{AlphaZero} and the supervised term, and extends this definition to the entire set of trajectories in the self-play dataset $\mathcal{D}$. The expectation $\mathbb{E}_{s \sim \mathcal{D}}$ in the AlphaZero term averages over all states encountered during self-play, whereas the expectation $\mathbb{E}_{\tau \sim \mathcal{D}}$ in the supervised term averages over all self-play trajectories.\\

\begin{equation}\label{total_loss}
\begin{split}
    \mathcal{L}&=\mathcal{L}_{\mathrm{AZ}}+\lambda\cdot\mathcal{L}_{\mathrm{supervised}}\\
    &=\mathbb{E}_{s \sim \mathcal{D}}\left[(R-v_\theta(s))^2-\pi^\intercal \log p_\theta(\cdot \mid s)\right] +c||\theta||^2 +\lambda\cdot\mathbb{E}_{\tau \sim \mathcal{D}}\left[\sum_{i=1}^{N_\tau} \frac{R_i}{T_i} \sum_{t=1}^{T_i} -\log p_\theta(a_t^{(i)}|s_t)\right]
\end{split}
\end{equation}

Given that training is performed on batches of individual states rather than full trajectories, it is more convenient to express the supervised loss as an expectation over states. In Equation \ref{total_loss_final}, $N_s$ denotes the number of expert trajectories associated with the training signal for state $s$, $a_s^{(i)}$ the expert action prescribed by trajectory $i$ for state $s$, and $\delta_s^{(i)}=1$ if state $s$ appears in expert trajectory $i$, and $\delta_s^{(i)}=0$ otherwise \footnote{Note that the numerical value of $\lambda$ differs between Equations \ref{total_loss} and \ref{total_loss_final}, as the former averages over the number of trajectories in the dataset, whereas the latter averages over the number of states.}.

\begin{equation}\label{total_loss_final}
\begin{split}
    \mathcal{L}=\mathbb{E}_{s \sim \mathcal{D}}\left[(R-v_\theta(s))^2-\pi^\intercal \log p_\theta(\cdot \mid s) +\lambda\cdot \left(\sum_{i=1}^{N_s} -\delta_s^{(i)} \frac{R_i}{T_i} \log p_\theta(a_s^{(i)}|s)\right)\right] +c||\theta||^2 
\end{split}
\end{equation}

We also propose an alternative to applying PGSU on the expert demonstrations, which we introduce as Periodic Expert Guidance (PEG), consisting of periodically using the GA-discovered backgrounds to enhance the RL agent's training: with a certain frequency, the agent is forced via action masking to play an episode in a trajectory identified as successful by the GAs, creating a mechanism to guide the agent's learning. Crucially, the agent is only periodically exposed to these configurations: this design ensures that the agent can learn from meaningful episodes while still being encouraged to explore and generalise beyond them, thereby achieving beyond-expert performance. PEG may be particularly interesting in RL problems involving very long trajectories, where, in the PGSU approach, the agent must first learn how to reach a state before receiving supervision for that state, which could result in a slower training. A comparison between the two approaches is presented in Section \ref{Results}.

\subsubsection{Determination of the relevant backgrounds}

Once the training of the agent is complete, the backgrounds corresponding to a new signal are determined by sampling actions from the probability distribution provided by the model until the ``End'' token is selected. This sampling process is repeated multiple times to obtain a set of token sequences. If training has been effective and the agent has achieved good generalisation, actions leading to a relevant background sequence will be assigned high probabilities, and the decay will likely be part of the sampled set. To promote diversity, a high sampling temperature ($T=2.5$) is used (see Equation \ref{temperature}, where $p_i$ is the original probability for action $i$, and $q_i$ is the adjusted probability after applying temperature scaling), which encourages the selection of actions beyond those with the highest probability, and thus target the determination of all relevant backgrounds.

\begin{equation}\label{temperature}
    q_i=\frac{p_i^{1/T}}{\sum_j p_j^{1/T}}
\end{equation}

\subsubsection{Reward shaping strategy}

Equation \ref{reward_definition} defines the reward in terms of the BRs of the signal and background processes, misidentifications of particles, and partial reconstruction of backgrounds. However, to address several practical challenges during training, we designed a reward shaping strategy. In particular, the raw rewards computed from Equation \ref{reward_definition} can span several orders of magnitude, potentially leading to instability and inefficiency in learning. To handle this, a transformed reward function is introduced in Equation \ref{reward_shaping1}, where $k$ is a scaling factor, and the logarithmic transformation serves to compresses the reward range.

\begin{equation}\label{reward_shaping1}
    r=\ln\left(1 + k \cdot \frac{BR_{b}}{BR_{s}} \cdot \mathcal{M} \cdot \mathcal{K}\right)
\end{equation}

A second challenge arises from the inherent tendency of RL algorithms to prioritise high-reward configurations while dedicating less attention to those associated with lower rewards. While this behaviour promotes the learning of the most relevant backgrounds, it may also lead to the neglect of backgrounds that, despite yielding smaller rewards, remain significant. To address this issue, the reward function employed in this work is defined in Equation \ref{reward_shaping2}, where $0\leq\alpha\leq1$. In this formulation, $r_\varepsilon$ defines a threshold that determines when a decay is considered sufficiently important to qualify as a relevant background within the toy scenario used for the experiments in this work.

\begin{equation}\label{reward_shaping2}
    R=\alpha r + (1-\alpha) \theta(r-r_\varepsilon)
\end{equation}

Equation \ref{reward_shaping2} represents a weighted sum of two components: the transformed reward in Equation \ref{reward_shaping1}, and an additional term based on the Heaviside function, which provides a fixed bonus to backgrounds considered relevant according to the threshold parameter $r_\varepsilon$. Two limiting cases are illustrative: setting $\alpha=1$ reduces Equation \ref{reward_shaping2} into Equation \ref{reward_shaping1}, while $\alpha=0$ results in a reward function that equally rewards only the relevant backgrounds. Additionally, we applied a penalisation when episodes are truncated due to the agent reaching invalid states with no available actions, thereby discouraging trajectories that lead to them.

\subsubsection{Model fine tuning}\label{fine_tuning}

After completing the training phase, we implemented a fine-tuning stage to further improve the model’s performance. During training, the algorithm records the relevant background configurations encountered by the agent. In the fine-tuning stage, the agent employs these trajectories to specialise on this knowledge. As in training, we designed two distinct strategies: for the agent trained with PGSU, fine tuning consists of performing a second PGSU employing all relevant background sequences built during training as expert demonstrations (rather than only the GA-discovered trajectories). In the fine tuning of the agent trained with PEG, the agent is required to play multiple episodes focusing solely on these specific configurations.

\subsubsection{Parallelisation of AlphaZero implementation}

During the training of the agent, each epoch of the AlphaZero algorithm begins with a self-play phase to generate training data. In this phase, $N_G$ different episodes are played, and each step within an episode involves performing a MCTS, which requires a model inference to estimate action probabilities and state values for each newly expanded node. The aforementioned open-source AlphaZero implementation \cite{AlphaZero_implementation} parallelises model inference across episodes, executing a single batched forward pass for all games. However, all other computations in the algorithm are performed sequentially, which results in a bottleneck for problems where environment-related computations are time consuming.\\

To address this limitation, we introduce additional parallelisation to the implementation in \cite{AlphaZero_implementation} using multiprocessing, in order to distribute the workload across multiple processes and accelerate computation. The set of episodes is partitioned across processes, with each process managing a subset of the games and performing simultaneous inference within that subset. To prevent contention for the model during inference, which would introduce a new bottleneck, each process employs its own deep-copied instance of the model. These copies are updated at the beginning of each training epoch to incorporate the latest weight changes. Our implementation also supports distributing these copies across multiple GPUs, enabling efficient parallel execution and improved scalability.\\

\subsection{GAs for background identification}\label{GAs}

As described earlier, the large trajectory space of the environment of this work and the highly sparse reward configuration present a major challenge for RL algorithms. GAs are optimisation techniques that efficiently explore complex parameter spaces and exhibit strong robustness against local optima. Consequently, leveraging GAs to identify successful trajectories in the environment enables the use of these trajectories as expert demonstrations for training the RL agent, thereby addressing the aforementioned challenges. From the GA perspective, Equation \ref{reward_definition} defines the fitness function to be optimised for identifying the most relevant backgrounds associated with a given signal. In this context, the individuals in the population represent potential backgrounds. Consequently, the objective is not only to identify the absolute maximum of the fitness function, but to find all solutions whose fitness value exceeds a threshold $r_\varepsilon$ that qualifies them as relevant backgrounds. In GA terminology, the set of best found solutions often receives the name of \textit{hall of fame}.\\

\begin{figure}[H]
\centering
\vspace{-2cm} 
\begin{tikzpicture}[
  level distance=15mm,
  every node/.style={circle, draw, minimum size=8mm, inner sep=1pt},
  every edge/.style={->, thick},
  level 1/.style={sibling distance=25mm},
  level 2/.style={sibling distance=25mm},
  level 3/.style={sibling distance=25mm}
]
% Left part: 
\node[draw=none, align=center] (expr) at (-9,2) {\Large Signal:};
\node[draw=none, align=center] (expr) at (-9,1) {\Large $M^S \to I^S(\to F_1 F_2)\,F_3$};
\node[circle, minimum size=1cm, draw, fill=red!30, font=\large] (MS) at (-9,-0.5) {$M^S$}
  child { node[circle, minimum size=1cm, draw, fill=cyan!30, font=\large] (IS) {$I^S$}
    child { node[circle, minimum size=1cm, draw, fill=green!20, font=\large] (F1S) {$F_1$} }
    child { node[circle, minimum size=1cm, draw, fill=green!20, font=\large] (F2S) {$F_2$} }
  }
  child { node[circle, minimum size=1cm, draw, fill=green!20, font=\large] (F3S) {$F_3$} };
% Right part: 
\node[draw=none, align=center] (expr) at (-1.5,2) {\Large Background candidate:};
\node[draw=none, align=center] (expr) at (-1.5,1) {\Large $M^B \to I_1^B(\to I_2^B(\to F_1 F_3) \ F_2)\,F_4$};
\node[circle, minimum size=1cm, draw, fill=red!30, font=\large] (MB) at (-1.5,-0.5) {$M^B$}
  child { node[circle, minimum size=1cm, draw, fill=cyan!30, font=\large] (I1B) {$I_1^B$}
    child { node[circle, minimum size=1cm, draw, fill=cyan!30, font=\large] (I2B) {$I_2^B$}
      child { node[circle, minimum size=1cm, draw, fill=green!20, font=\large] (F1) {$F_1$} }
      child { node[circle, minimum size=1cm, draw, fill=green!20, font=\large] (F3) {$F_3$} }
    }
    child { node[circle, minimum size=1cm, draw, fill=green!20, font=\large] (F2) {$F_2$} }
  }
  child { node[circle, minimum size=1cm, draw, fill=green!20, font=\large] (F4) {$F_4$} };
\end{tikzpicture}
\caption{Tree-based representation of the genes of the background candidate $M^B \to I_1^B(\to I_2^B(\to F_1 F_3) \ F_2)\,F_4$, for the signal $M^S \to I^S(\to F_1 F_2)\,F_3$. Mother particles $M^S$ and $M^B$ are coloured in red ($S$ and $B$ here denote signal and background respectively), intermediate particles $I^S$, $I_1^B$ and $I_2^B$ in blue, and final state particles $F_1$, $F_2$, $F_3$, and $F_4$ in green. In this case, the background candidate has two intermediate resonances and a non-reconstructed particle $F_4$. A physically meaningful example is given by the signal $B^0\to K^{*+}(\to K^+ \pi^0) \pi^-$ and the background $B^0\to K^{*}\mathrm{(1680)}^0(\to K^{*0}(\to K^+ \pi^-) \pi^0) K^0$, where $M^S=M^B=B^0$, $I^S=K^{*+}$, $I_1^B=K^{*}\mathrm{(1680)}^0$, $I_2^B=K^{*0}$, $F_1=K^+$, $F_2=\pi^0$, $F_3=\pi^-$ and $F_4=K^0$.}
\label{tree_structure}
\end{figure}
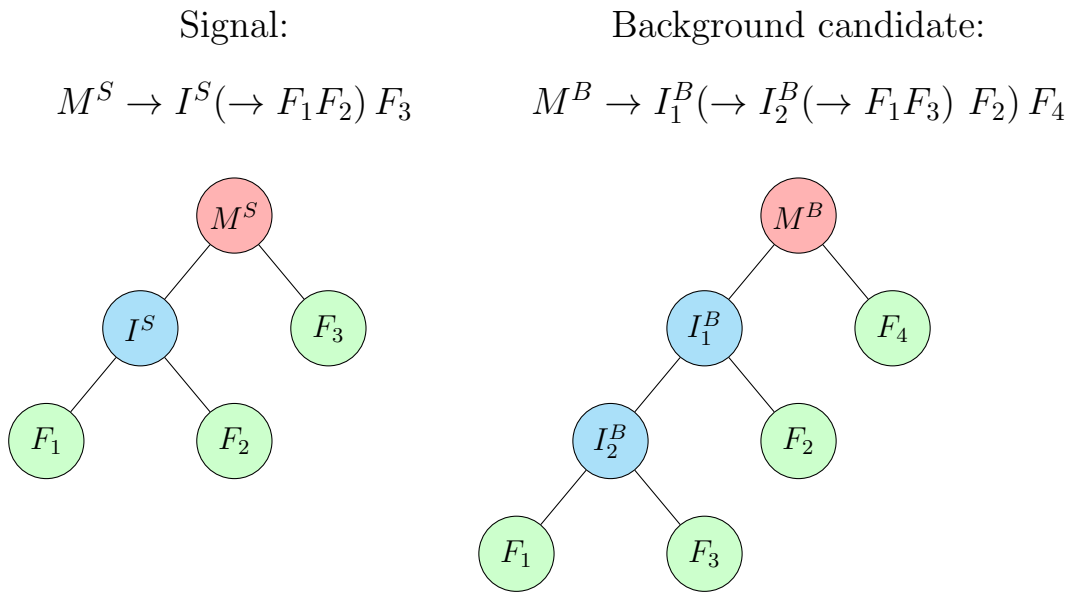

The authors developed the GA implementation used in this work from scratch. Unlike traditional GAs, which typically represent individuals as vectors, we adopted a tree-based representation to naturally encode decay chains, as illustrated in Figure \ref{tree_structure}. In this representation, a gene corresponds to a branch of the tree, namely, a node directly connected to the root (the mother particle of the decay) and all of its descendant nodes. For example, the individual depicted in Figure \ref{tree_structure} consists of two genes: $F_4$ and $I_1^B(\to I_2^B(\to F_1 F_3) \ F_2)$.

\subsubsection{Crossover and mutation processes}

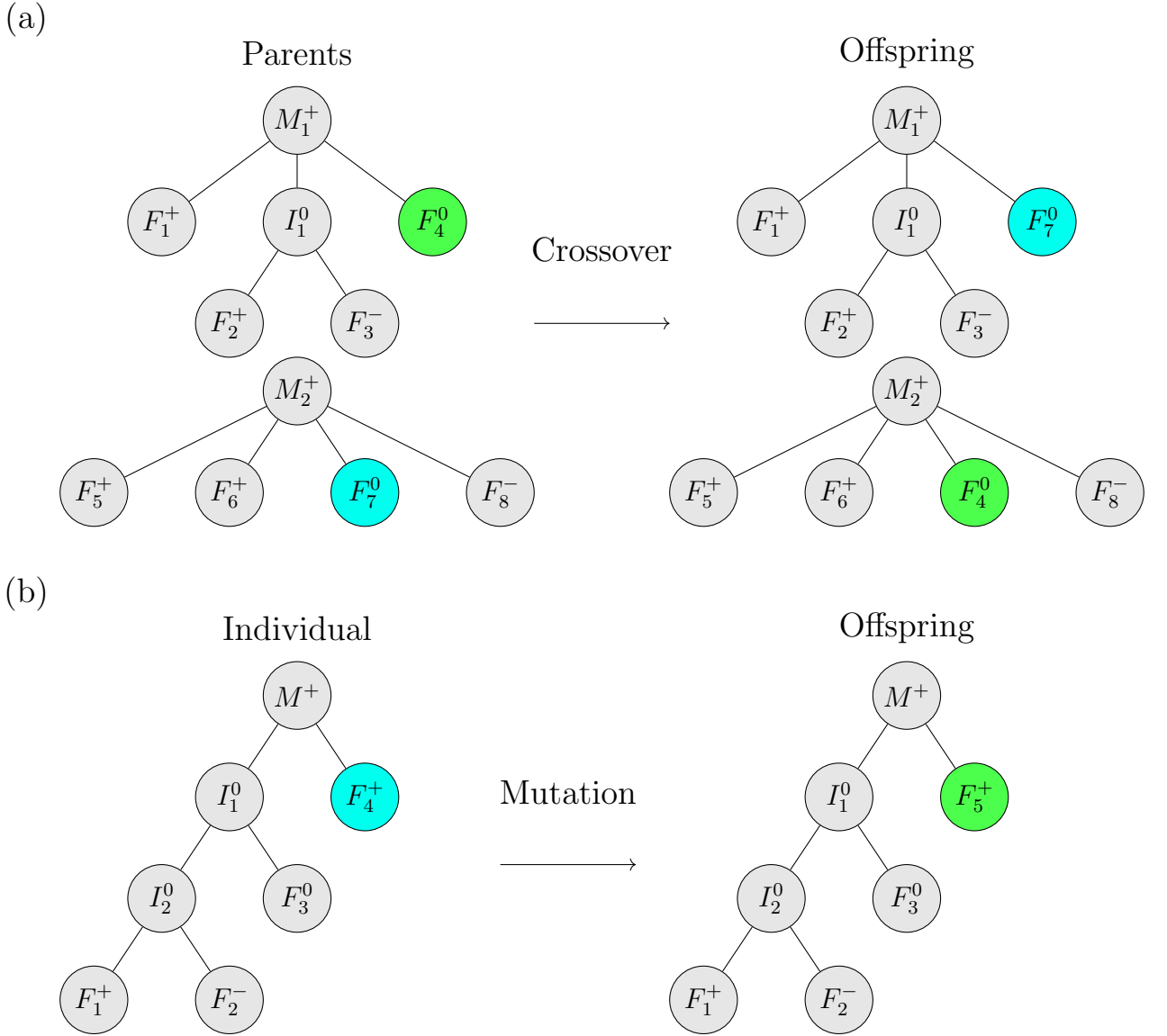
\begin{figure}[H]
\centering
\vspace{-0.3cm} 
\begin{tikzpicture}[
  level distance=15mm,
  every node/.style={circle, draw, minimum size=8mm, inner sep=1pt},
  every edge/.style={->, thick},
  level 1/.style={sibling distance=20mm},
  level 2/.style={sibling distance=20mm},
  level 3/.style={sibling distance=20mm}
]
\definecolor{turquoise}{RGB}{0, 255, 239}%Classic turquoise
\node[draw=none, align=center] (expr) at (-16,0.5) {\Large (a)};
% Left part: 
\node[draw=none, align=center] (expr) at (-12,0) {\Large Parents};
\node[circle, minimum size=1cm, draw, fill=gray!20, font=\large] (MS) at (-12,-1) {$M_1^+$}
  child { node[circle, minimum size=1cm, draw, fill=gray!20, font=\large] (F3S) {$F_1^+$} }
  child { node[circle, minimum size=1cm, draw, fill=gray!20, font=\large] (IS) {$I_1^0$}
    child { node[circle, minimum size=1cm, draw, fill=gray!20, font=\large] (F1S) {$F_2^+$} }
    child { node[circle, minimum size=1cm, draw, fill=gray!20, font=\large] (F2S) {$F_3^-$} }
  }
  child { node[circle, minimum size=1cm, draw, fill=green!70, font=\large] (F3S) {$F_4^0$} };
\node[circle, minimum size=1cm, draw, fill=gray!20, font=\large] (MS) at (-12,-5) {$M_2^+$}
  child { node[circle, minimum size=1cm, draw, fill=gray!20, font=\large] (F3S) {$F_5^+$} } 
  child { node[circle, minimum size=1cm, draw, fill=gray!20, font=\large] (IS) {$F_6^+$} }
  child { node[circle, minimum size=1cm, draw, fill=turquoise, font=\large] (IS) {$F_7^0$} }
  child { node[circle, minimum size=1cm, draw, fill=gray!20, font=\large] (IS) {$F_8^-$} }
    ;
\draw[->] (-8.5,-4) -- (-6.5,-4) node[midway, above=0pt, draw=none] {\Large Crossover}; 
% Right part: 
\node[draw=none, align=center] (expr) at (-3,0) {\Large Offspring};
\node[circle, minimum size=1cm, draw, fill=gray!20, font=\large] (MS) at (-3,-1) {$M_1^+$}
  child { node[circle, minimum size=1cm, draw, fill=gray!20, font=\large] (F3S) {$F_1^+$} }
  child { node[circle, minimum size=1cm, draw, fill=gray!20, font=\large] (IS) {$I_1^0$}
    child { node[circle, minimum size=1cm, draw, fill=gray!20, font=\large] (F1S) {$F_2^+$} }
    child { node[circle, minimum size=1cm, draw, fill=gray!20, font=\large] (F2S) {$F_3^-$} }
  }
  child { node[circle, minimum size=1cm, draw, fill=turquoise, font=\large] (F3S) {$F_7^0$} };
\node[circle, minimum size=1cm, draw, fill=gray!20, font=\large] (MS) at (-3,-5) {$M_2^+$}
  child { node[circle, minimum size=1cm, draw, fill=gray!20, font=\large] (F3S) {$F_5^+$} } 
  child { node[circle, minimum size=1cm, draw, fill=gray!20, font=\large] (IS) {$F_6^+$} }
  child { node[circle, minimum size=1cm, draw, fill=green!70, font=\large] (IS) {$F_4^0$} }
  child { node[circle, minimum size=1cm, draw, fill=gray!20, font=\large] (IS) {$F_8^-$} }
    ;
\node[draw=none, align=center] (expr) at (-16,-8) {\Large (b)};
\node[draw=none, align=center] (expr) at (-12,-8.5) {\Large Individual};
\node[draw=none, align=center] (expr) at (-3,-8.5) {\Large Offspring};
\node[circle, minimum size=1cm, draw, fill=gray!20, font=\large] (MB) at (-12,-9.5) {$M^+$}
  child { node[circle, minimum size=1cm, draw, fill=gray!20, font=\large] (I1B) {$I_1^0$}
    child { node[circle, minimum size=1cm, draw, fill=gray!20, font=\large] (I2B) {$I_2^0$}
      child { node[circle, minimum size=1cm, draw, fill=gray!20, font=\large] (F1) {$F_1^+$} }
      child { node[circle, minimum size=1cm, draw, fill=gray!20, font=\large] (F3) {$F_2^-$} }
    }
    child { node[circle, minimum size=1cm, draw, fill=gray!20, font=\large] (F2) {$F_3^0$} }
  }
  child { node[circle, minimum size=1cm, draw, fill=turquoise, font=\large] (F4) {$F_4^+$} };
  \draw[->] (-9,-12) -- (-7,-12) node[midway, above=0pt, draw=none] {\Large Mutation};
\node[circle, minimum size=1cm, draw, fill=gray!20, font=\large] (MB) at (-3,-9.5) {$M^+$}
  child { node[circle, minimum size=1cm, draw, fill=gray!20, font=\large] (I1B) {$I_1^0$}
    child { node[circle, minimum size=1cm, draw, fill=gray!20, font=\large] (I2B) {$I_2^0$}
      child { node[circle, minimum size=1cm, draw, fill=gray!20, font=\large] (F1) {$F_1^+$} }
      child { node[circle, minimum size=1cm, draw, fill=gray!20, font=\large] (F3) {$F_2^-$} }
    }
    child { node[circle, minimum size=1cm, draw, fill=gray!20, font=\large] (F2) {$F_3^0$} }
  }
  child { node[circle, minimum size=1cm, draw, fill=green!70, font=\large] (F4) {$F_5^+$} };
\end{tikzpicture}
\caption{Illustration of crossover (a) and mutation (b) processes for a tree-based gene representation. The relevant genes have been highlighted in colours. Mother particles are denoted by $M$, intermediate particles by $I$, and final state particles by $F$. Here, superscripts indicate the electric charge: $+$ for positively charged particles, $-$ for negatively charged particles and $0$ for neutral ones. Physically meaningful examples would be a crossover between $B^+\to K^{*0}(\to K^+ \pi^-) \pi^+ \gamma$ and $B^+\to K^+ \pi^+ \pi^0 \pi^-$, which produces an offspring of $B^+\to K^{*0}(\to K^+ \pi^-) \pi^+ \pi^0$ and $B^+\to K^+ \pi^+ \gamma \ \pi^-$, and a mutation transforming $B^+\to K^*(1680)^0(\to K^{*0}(\to K^+ \pi^-) \pi^0) \pi^+$ into $B^+\to K^*(1680)^0(\to K^{*0}(\to K^+ \pi^-) \pi^0) K^+$.}
\label{combination_mutation_tree}
\end{figure}

Crossover and mutation processes for this type of gene representation are depicted in Figure \ref{combination_mutation_tree}. In a crossover process, an exact copy of the parents is first generated, followed by the possibility of gene exchange. For a given gene in the first offspring individual, with probability of $P=0.5$, all equivalent genes in the second offspring individual are identified, and one is randomly selected. The selected genes are then exchanged, after which they become unavailable for further exchanges in the crossover process. Here, \textit{equivalent}, refers to genes containing the same number of positive, negative, and neutral final state particles. This approach ensures that the total count of each particle type remains unchanged in the individuals.\\

For a mutation, with a certain probability, a gene representing a final state particle is replaced by another particle of the same charge. In the example illustrated in Figure \ref{combination_mutation_tree}(b), $F_4^+$ was replaced by $F_5^+$. We chose the genes representing an intermediate resonance to be immutable: they can be transmitted to the offspring in a crossover process, but a mutation cannot affect them. This way, the already formed decay structures are preserved.\\

\subsubsection{Offspring generation and natural selection}\label{offspring_generation_natural_selection}

In GAs, different strategies can be employed for generating offspring, integrating them with the current population, and selecting individuals for the next generation. The strategy adopted in this work involves performing the following steps iteratively:

\begin{enumerate}
    \item The entire population is first cloned, after which cloned individuals undergo genetic variations according to predefined probabilities.
    \item The original population is then merged with the cloned population that has undergone the variation processes.
    \item The next generation is obtained through natural selection. First, elite individuals\footnote{The elite refers to the best individuals in the current population of a specific generation, typically preserved for the next generation to ensure that high-quality solutions are not lost. On the other hand, the hall of fame is a mechanism to track and preserve the best individuals across multiple generations, not just within a single generation.} are identified. These are individuals with fitness values sufficiently high to be considered relevant backgrounds, and they are directly carried over to the next generation. Additionally, tournament selection is employed within the population. In each tournament, three individuals are randomly chosen, and the individual with the highest fitness value is selected to advance to the next generation. This selection continues until the new generation matches the size of the previous population.
\end{enumerate}

\subsubsection{Random immigration}

\begin{figure}[H]
\centering
\vspace{-1cm} 
\begin{tikzpicture}[
  level distance=12mm,
  every node/.style={circle, draw, minimum size=8mm, inner sep=1pt},
  every edge/.style={->, thick},
  level 1/.style={sibling distance=12mm},
  level 2/.style={sibling distance=12mm},
  level 3/.style={sibling distance=12mm}
]
\definecolor{turquoise}{RGB}{0, 255, 239}%Classic turquoise
\definecolor{dark_blue}{RGB}{0, 127, 255}%Classic turquoise
\definecolor{gray_blue}{RGB}{204, 229, 255}%Classic turquoise
\node[draw=none, align=center] (expr) at (-12,0) {\Large Current population};
\node[draw=none, align=center] (expr) at (-5.5,0) {\Large Offspring before \\ \Large random immigration};
\node[draw=none, align=center] (expr) at (1,0) {\Large Offspring after \\ \Large random immigration};
\node[circle, draw, fill=turquoise, font=\large] (MB) at (-12,-1) {}
  child { node[circle, draw, fill=turquoise, font=\large] (I1B) {} }
  child { node[circle, draw, fill=turquoise, font=\large] (I1B) {} }
  child { node[circle, draw, fill=turquoise, font=\large] (I1B) {} }
  child { node[circle, draw, fill=turquoise, font=\large] (F4) {} };
\node[circle, draw, fill=dark_blue, font=\large] (MB) at (-12,-3) {}
  child { node[circle, draw, fill=dark_blue, font=\large] (I1B) {} }
  child { node[circle, draw, fill=dark_blue, font=\large] (I1B) {} }
  child { node[circle, draw, fill=dark_blue, font=\large] (I1B) {} }
  child { node[circle, draw, fill=dark_blue, font=\large] (F4) {} };
\node[circle, draw, fill=gray_blue, font=\large] (MB) at (-12,-5) {}
  child { node[circle, draw, fill=gray_blue, font=\large] (I1B) {} }
  child { node[circle, draw, fill=gray_blue, font=\large] (I1B) {} }
  child { node[circle, draw, fill=gray_blue, font=\large] (I1B) {} }
  child { node[circle, draw, fill=gray_blue, font=\large] (F4) {} };
\node[circle, draw, fill=turquoise, font=\large] (MB) at (-5.5,-1) {}
  child { node[circle, draw, fill=dark_blue, font=\large] (I1B) {} }
  child { node[circle, draw, fill=turquoise, font=\large] (I1B) {} }
  child { node[circle, draw, fill=dark_blue, font=\large] (I1B) {} }
  child { node[circle, draw, fill=turquoise, font=\large] (F4) {} };
\node[circle, draw, fill=dark_blue, font=\large] (MB) at (-5.5,-3) {}
  child { node[circle, draw, fill=dark_blue, font=\large] (I1B) {} }
  child { node[circle, draw, fill=gray_blue, font=\large] (I1B) {} }
  child { node[circle, draw, fill=dark_blue, font=\large] (I1B) {} }
  child { node[circle, draw, fill=green!70, font=\large] (F4) {} };
\node[circle, draw, fill=gray_blue, font=\large] (MB) at (-5.5,-5) {}
  child { node[circle, draw, fill=gray_blue, font=\large] (I1B) {} }
  child { node[circle, draw, fill=gray_blue, font=\large] (I1B) {} }
  child { node[circle, draw, fill=gray_blue, font=\large] (I1B) {} }
  child { node[circle, draw, fill=gray_blue, font=\large] (F4) {} };
\node[circle, draw, fill=turquoise, font=\large] (MB) at (1,-1) {}
  child { node[circle, draw, fill=dark_blue, font=\large] (I1B) {} }
  child { node[circle, draw, fill=turquoise, font=\large] (I1B) {} }
  child { node[circle, draw, fill=dark_blue, font=\large] (I1B) {} }
  child { node[circle, draw, fill=turquoise, font=\large] (F4) {} };
\node[circle, draw, fill=dark_blue, font=\large] (MB) at (1,-3) {}
  child { node[circle, draw, fill=dark_blue, font=\large] (I1B) {} }
  child { node[circle, draw, fill=gray_blue, font=\large] (I1B) {} }
  child { node[circle, draw, fill=dark_blue, font=\large] (I1B) {} }
  child { node[circle, draw, fill=green!70, font=\large] (F4) {} };
\node[circle, draw, fill=red!30, font=\large] (MB) at (1,-5) {}
  child { node[circle, draw, fill=red!30, font=\large] (I1B) {} }
  child { node[circle, draw, fill=red!30, font=\large] (I1B) {} }
  child { node[circle, draw, fill=red!30, font=\large] (I1B) {} }
  child { node[circle, draw, fill=red!30, font=\large] (F4) {} };
\draw[->, very thick] (-9.5,-3.5) -- (-8,-3.5) node[midway, above=0pt, draw=none] {};
\draw[->, very thick] (-3,-3.5) -- (-1.5,-3.5) node[midway, above=0pt, draw=none] {};
\end{tikzpicture}
\caption{Illustration of the random immigration variation process. A redundant individual in the offspring is replaced by a newly generated individual.}
\label{random_immigration}
\end{figure}
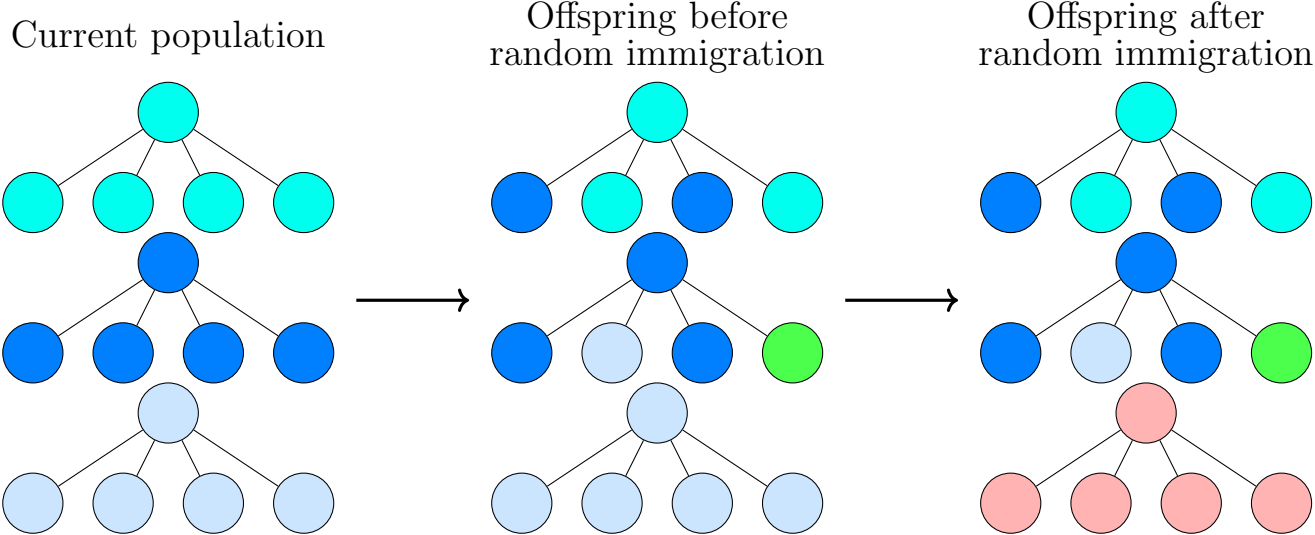

As described in Section \ref{offspring_generation_natural_selection}, the offspring is generated by cloning the original population and applying variation processes to it. Since these processes occur probabilistically, duplicates of original individuals that have not undergone genetic variation may appear in the offspring. To address this redundancy, we employ a common technique known as \textit{random immigration}, which replaces duplicated individuals with newly generated random ones. This approach increases genetic diversity within the population, enhancing exploration without enlarging the population size. Additionally, it reduces dependence on the initialisation of the population. Figure \ref{random_immigration} illustrates this concept.

\subsubsection{Custom variation processes}

In addition to the traditional variation processes, we designed custom variation processes to address specific objectives, such us constructing intermediate resonances, enhancing exploration efficiency by incorporating physical properties, and reusing learnt information in future optimisation problems.\\

\paragraph{Intermediate resonance construction}

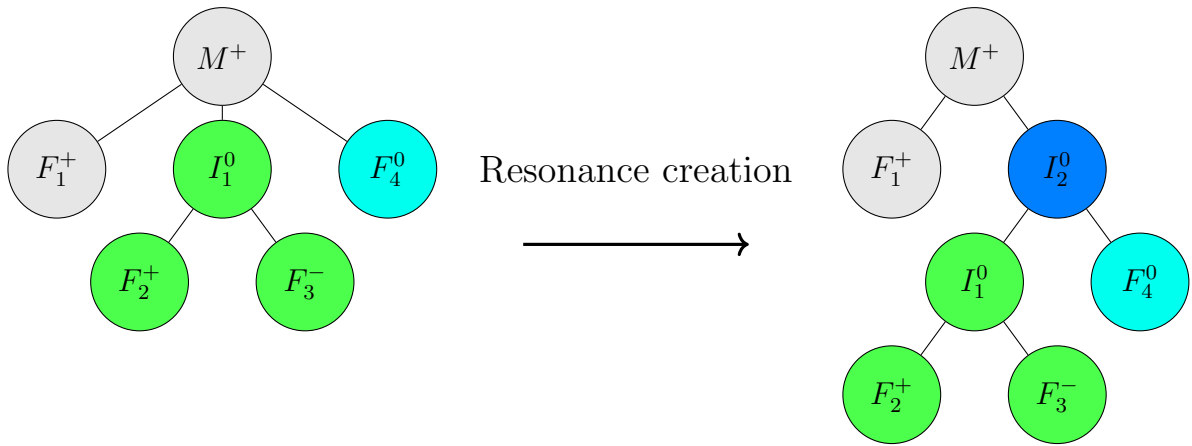
\begin{figure}[H]
\centering
\vspace{-0.3cm} 
\begin{tikzpicture}[
  level distance=15mm,
  every node/.style={circle, draw, minimum size=8mm, inner sep=1pt},
  every edge/.style={->, thick},
  level 1/.style={sibling distance=22mm},
  level 2/.style={sibling distance=22mm},
  level 3/.style={sibling distance=22mm}
]
\definecolor{turquoise}{RGB}{0, 255, 239}%Classic turquoise
\definecolor{dark_blue}{RGB}{0, 127, 255}%Classic turquoise
\node[circle, minimum size=1.3cm, draw, fill=gray!20, font=\large] (MB) at (-12,-9.5) {$M^+$}
  child { node[circle, minimum size=1.3cm, draw, fill=gray!20, font=\large] (F4) {$F_1^+$} }
  child { node[circle, minimum size=1.3cm, draw, fill=green!70, font=\large] (I1B) {$I_1^0$}
    child { node[circle, minimum size=1.3cm, draw, fill=green!70, font=\large] (I2B) {$F_2^+$}
    }
    child { node[circle, minimum size=1.3cm, draw, fill=green!70, font=\large] (F2) {$F_3^-$} }
  }
  child { node[circle, minimum size=1.3cm, draw, fill=turquoise, font=\large] (F4) {$F_4^0$} };
  \draw[->, very thick] (-8,-12) -- (-5,-12) node[midway, draw=none] {};
\node[draw=none, align=center] (expr) at (-6.5,-11) {\Large Resonance creation};
\node[circle, minimum size=1.3cm, draw, fill=gray!20, font=\large] (MB) at (-2,-9.5) {$M^+$}
  child { node[circle, minimum size=1.3cm, draw, fill=gray!20, font=\large] (F) {$F_1^+$} }
  child { node[circle, minimum size=1.3cm, draw, fill=dark_blue, font=\large] (F4) {$I_2^0$}  
    child { node[circle, minimum size=1.3cm, draw, fill=green!70, font=\large] (I1B) {$I_1^0$}
      child { node[circle, minimum size=1.3cm, draw, fill=green!70, font=\large] (I2B) {$F_2^+$} }
      child { node[circle, minimum size=1.3cm, draw, fill=green!70, font=\large] (F2) {$F_3^-$} }
    }
    child { node[circle, minimum size=1.3cm, draw, fill=turquoise, font=\large] (F5) {$F_4^0$} }
  } ;
\end{tikzpicture}
\caption{Illustration of the custom variation process to construct an intermediate resonance. Mother particles are denoted by $M$, intermediate particles by $I$, and final state particles by $F$. Here, superscripts indicate the electric charge: $+$ for positively charged particles, $-$ for negatively charged particles and $0$ for neutral ones. In the example, genes $I_1^0(\to F_2^+ F_3^-)$ and $F_4^0$ are combined to construct the intermediate resonance $I_2^0(\to I_1^0(\to F_2^+ F_3^-) F_4^0 )$. A physically meaningful case would be the creation of the background candidate $B^+\to K^*(1680)^0 (\to K^{*0}(\to K^+ \pi^-) \pi^0) \pi^+$ from the GA individual $B^+\to K^{*0}(\to K^+ \pi^-) \pi^+ \pi^0$.}
\label{resonance_creation}
\end{figure}

By default, we chose to initialise the individuals in the population without intermediate resonances: the tree structure includes only the mother particle and its directly connected final-state particles. To introduce intermediate resonances within the gene representation, we implemented a custom variation process. In this process, a subset of genes is randomly selected, the required charge for the corresponding intermediate particle is computed, and an intermediate particle with the appropriate charge is then randomly sampled. In future work, this procedure could be extended to generate intermediate resonances that also satisfy additional conservation laws, such as lepton number or energy conservation, thereby producing physically more constrained decay structures and enhancing the efficiency of the GA by reducing the search space. An illustration of this resonance construction process is provided in Figure \ref{resonance_creation}.\\

We designed two different strategies for selecting the intermediate particle: a naive approach, in which the particle is randomly sampled from the set of intermediate particles with the appropriate charge, and a more sophisticated method that relies on decay channels already incorporated into the algorithm’s internal knowledge base. The designed GAs collect knowledge about existing decays in a dictionary structure. Each key in the dictionary corresponds to a particle, and the associated value is the set of allowed decay channels for that particle. This dictionary is initialised with decay information from the PDG. However, if a decay not already present in the dictionary is encountered during optimisation, the GA \textit{learns}\footnote{Not to be confused with the meaning of learning in ML.} and incorporates this new information. In contrast to conventional GA applications, where problems are typically independent by nature (the optimisation of a function $f_1$ does not affect the optimisation of a different function $f_2$), this property of independence does not hold in the context of the background finder: the GA may identify a background involving a decay not listed in the PDG, and this newly discovered information could be relevant for other signals beyond the one currently under consideration. After the optimisation for a given signal is completed, the dictionary can be saved and reused for background identification in subsequent signals, avoiding redundant rediscovery of decays.\\

Although charge conservation is ensured in the naive resonance construction, other conservation laws are not necessarily satisfied. Nevertheless, this approach allows exploration beyond the current knowledge in the PDG. In contrast, decays stored in the dictionary are guaranteed to be physically valid, allowing for more efficient construction of intermediate resonances. To balance exploration and exploitation, these two approaches are combined: when constructing an intermediate resonance, the naive method is applied with probability $P_{\mathrm{Naive}}$, while with probability $1-P_{\mathrm{Naive}}$, a decay from the dictionary is suggested. For the latter, all intermediate resonances compatible with the selected genes are identified, and one is sampled with a probability proportional to its BR.\\

\paragraph{Inheriting from the signal}

\begin{figure}[H]
\centering
%\vspace{-1cm} 
\begin{tikzpicture}[
  level distance=12mm,
  every node/.style={circle, draw, minimum size=8mm, inner sep=1pt},
  every edge/.style={->, thick},
  level 1/.style={sibling distance=12mm},
  level 2/.style={sibling distance=12mm},
  level 3/.style={sibling distance=12mm}
]
\definecolor{turquoise}{RGB}{0, 255, 239}%Classic turquoise
\definecolor{dark_blue}{RGB}{0, 127, 255}%Classic turquoise
\definecolor{gray_blue}{RGB}{204, 229, 255}%Classic turquoise
\node [fill=gray!10, rectangle, fit={(-14.4,-7) (-9.6,0.5)}] {};
\node [fill=gray!10, rectangle, fit={(-7.9,-7) (-3.1,0.5)}] {};
\node [fill=gray!10, rectangle, fit={(-9,0.7) (-4,4)}] {};
\node[draw=none, align=center] (expr) at (-12,0) {\Large Current population};
\node[draw=none, align=center] (expr) at (-5.5,0) {\Large Cloned population};
\node[draw=none, align=center] (expr) at (-6.5,3.5) {\Large Signal};
\node[draw=none, align=center] (expr) at (1,-3.5) {\Large Apply variation \\ \Large processes to cloned \\ \Large population to \\ \Large obtain offspring};
\node[circle, draw, fill=yellow!90, font=\large] (MB) at (-6.5,2.5) {}
  child { node[circle, draw, fill=yellow!90, font=\large] (I1B) {} }
  child { node[circle, draw, fill=yellow!90, font=\large] (I1B) {} }
  child { node[circle, draw, fill=yellow!90, font=\large] (I1B) {} }
  child { node[circle, draw, fill=yellow!90, font=\large] (F4) {} };
\node[circle, draw, fill=turquoise, font=\large] (MB) at (-12,-1) {}
  child { node[circle, draw, fill=turquoise, font=\large] (I1B) {} }
  child { node[circle, draw, fill=turquoise, font=\large] (I1B) {} }
  child { node[circle, draw, fill=turquoise, font=\large] (I1B) {} }
  child { node[circle, draw, fill=turquoise, font=\large] (F4) {} };
\node[circle, draw, fill=dark_blue, font=\large] (MB) at (-12,-3) {}
  child { node[circle, draw, fill=dark_blue, font=\large] (I1B) {} }
  child { node[circle, draw, fill=dark_blue, font=\large] (I1B) {} }
  child { node[circle, draw, fill=dark_blue, font=\large] (I1B) {} }
  child { node[circle, draw, fill=dark_blue, font=\large] (F4) {} };
\node[circle, draw, fill=gray_blue, font=\large] (MB) at (-12,-5) {}
  child { node[circle, draw, fill=gray_blue, font=\large] (I1B) {} }
  child { node[circle, draw, fill=gray_blue, font=\large] (I1B) {} }
  child { node[circle, draw, fill=gray_blue, font=\large] (I1B) {} }
  child { node[circle, draw, fill=gray_blue, font=\large] (F4) {} };
\node[circle, draw, fill=turquoise, font=\large] (MB) at (-5.5,-1) {}
  child { node[circle, draw, fill=turquoise, font=\large] (I1B) {} }
  child { node[circle, draw, fill=turquoise, font=\large] (I1B) {} }
  child { node[circle, draw, fill=turquoise, font=\large] (I1B) {} }
  child { node[circle, draw, fill=turquoise, font=\large] (F4) {} };
\node[circle, draw, fill=yellow!90, font=\large] (MB) at (-5.5,-3) {}
  child { node[circle, draw, fill=yellow!90, font=\large] (I1B) {} }
  child { node[circle, draw, fill=yellow!90, font=\large] (I1B) {} }
  child { node[circle, draw, fill=yellow!90, font=\large] (I1B) {} }
  child { node[circle, draw, fill=yellow!90, font=\large] (F4) {} };
\node[circle, draw, fill=gray_blue, font=\large] (MB) at (-5.5,-5) {}
  child { node[circle, draw, fill=gray_blue, font=\large] (I1B) {} }
  child { node[circle, draw, fill=gray_blue, font=\large] (I1B) {} }
  child { node[circle, draw, fill=gray_blue, font=\large] (I1B) {} }
  child { node[circle, draw, fill=gray_blue, font=\large] (F4) {} };
\draw[->, very thick] (-9.5,-3.5) -- (-8,-3.5) node[midway, above=0pt, draw=none] {};
\draw[->, very thick] (-3,-3.5) -- (-1.5,-3.5) node[midway, above=0pt, draw=none] {};
\end{tikzpicture}
\caption{Strategy to enhance exploration of the region in search space surrounding the signal: with a certain probability, genes are cloned directly from the signal instead of from the current population.}
\label{inherit_from_the_signal}
\end{figure}
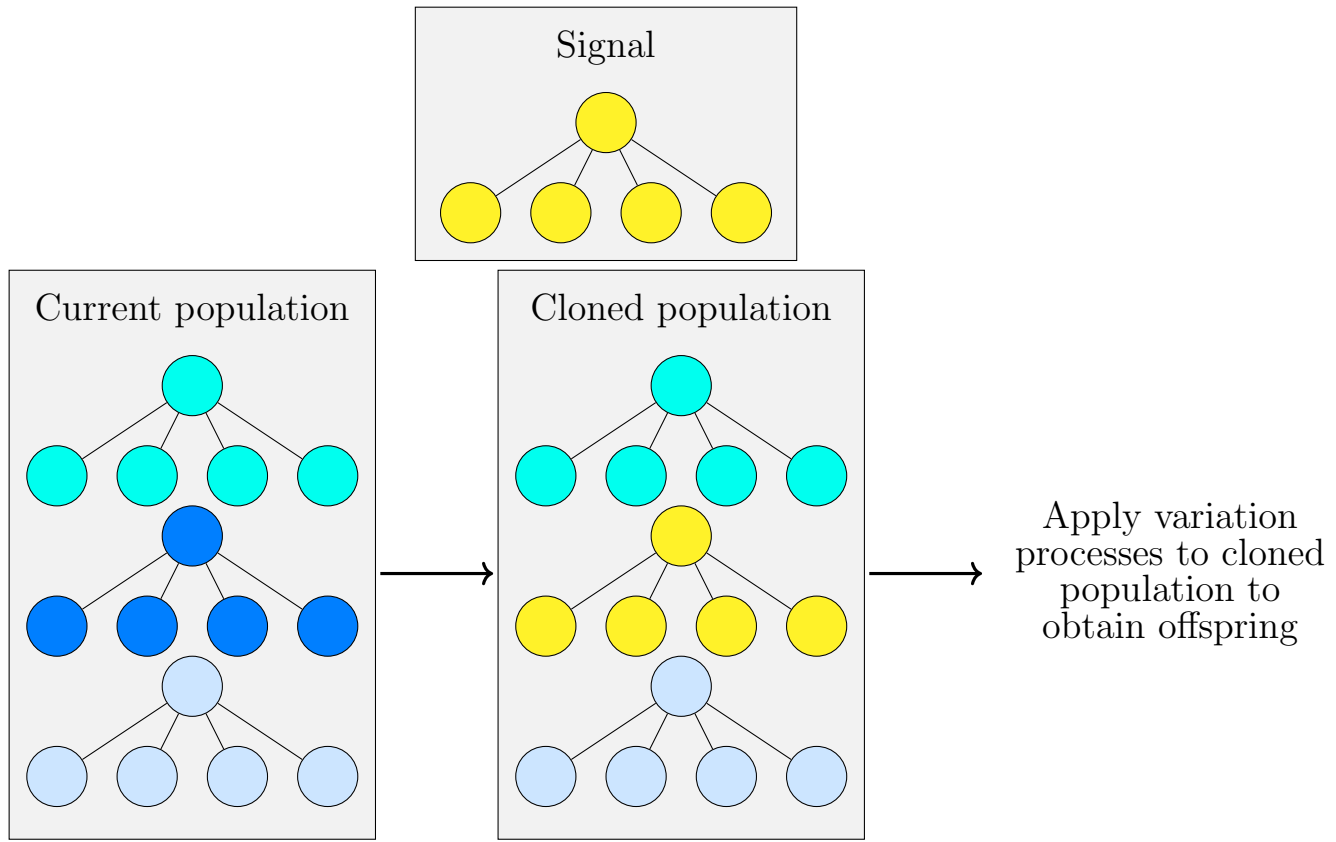

Relevant backgrounds are decays that closely mimic the detector response of the signal. Therefore, significant similarities are expected between the decay chains of the background and the signal. To exploit this, we have developed a strategy that, with a specified probability, clones genes from the signal itself rather than from individuals in the current population. This method enhances exploration around the signal’s region in the search space, where a higher concentration of relevant backgrounds is expected. To further enhance the search efficiency, cloning was not limited to the exact signal decay. We also included the CP-conjugate of the signal, the signal decay without intermediate resonances (retaining only the final-state particles), as well as variants obtained by interchanging first and second generation leptons in the signal. This diverse set of clones increases the likelihood of generating individuals that are relevant backgrounds. Figure \ref{inherit_from_the_signal} illustrates this approach.\\

\section{Environment}\label{environment}

We developed a toy scenario to assess the performance of the designed algorithm. A set of 40 particles is considered in this scenario, as presented in Table \ref{particle_chart}. We selected B-mesons as mother particles due to their central role in flavour physics and the complexity of their decay topologies, which provide a realistic scenario for the proposed method. Similarly, the choice of intermediate particles was guided by the need to include experimentally well-established intermediate states, such as kaon and D-meson excitations, that frequently appear in B-meson decays and yield sufficiently rich decay chains to challenge the algorithm.

{
\renewcommand\arraystretch{2.0} % only affects this table
\begin{table}[htbp]
    \centering
    \begin{tabular}{|P{5cm}|P{12cm}|}%{|c|c|c|}
        \hline
         Particle type & Particles\\ \hline \hline
         Neutrinos & $\neutrino_e,\neutrino_{\mu},\neutrino_{\tau},\antineutrino_e,\antineutrino_{\mu},\antineutrino_{\tau}$ \\ \hline
         Detectable final state particles & $e^+,e^-,\mu^+,\mu^-,K^+,K^-,K^0,\bar{K}^0,\pi^+,\pi^-,\pi^0,\gamma$ \\ \hline
         Intermediate particles & \makecell[c]{$K^{*+},K^{*-},K^{*0},\bar{K}^{*0},K^*_0(1430)^0,\bar{K}^*_0(1430)^0,K^*(1680)^0,\bar{K}^*(1680)^0,$ \\ $D^+,D^-,D^0,\bar{D}^0,
    D^{*+},D^{*-},D^{*0},\bar{D}^{*0},\tau^+,\tau^-$} \\ \hline
         Mother particles & $B^+,B^-,B^0,\bar{B}^0$ \\ \hline
    \end{tabular}
    \caption{Particles used in the toy scenario.}
    \label{particle_chart}
\end{table}
}

We selected \numbertrainingsignalsnumeric\ distinct signals to train the agent, while \numbergeneralisationsignalsnumeric\ additional signals are used to evaluate the agent's generalisation ability in determining the relevant backgrounds for signals not encountered during training. Due to the limited size of the training signal set, the generalisation signals are chosen to be similar to the training signals, so that the agent can achieve strong generalisation ability. For example, \numberCPconjugates\ of the generalisation signals are CP conjugates of \numberCPconjugates\ of the training signals, where CP refers to the combined charge conjugation and parity transformation. We imposed a constraint on the backgrounds: only decay processes involving a maximum of two intermediate resonances and no more than one non-reconstructed particle are allowed. We set the threshold to qualify a background as relevant to $r_\varepsilon$=\rewardthreshold. The selected signals, along with their corresponding relevant backgrounds are presented in Appendix \ref{signals_and_backgrounds}. The number of relevant backgrounds varies significantly, ranging from \lowestnumberbackgrounds\ to \highestnumberbackgrounds\ depending on the signal. The rewards corresponding to these backgrounds, after logarithmic transformation, range from \lowestrewardln\ to \highestrewardln. This implies that the raw rewards range approximately from \lowestreward\ to \highestreward, spanning more than two orders of magnitude. Together, these wide ranges constitute a considerable challenge for determining all relevant backgrounds.\\

We utilised a simplified form for each of the terms in Equation \ref{reward_definition}: 
\begin{itemize}
    \item The BRs from the PDG database are assumed to represent the true values.
    \item While the misidentification rate depends on the particle species and the experiment in question, we assume a global misidentification rate of 1\% to validate our algorithmic approach, and write the misidentification penalty $\mathcal{M}$ as $\mathcal{M}=0.01^{N_{misID}}$, where $N_{misID}$ denotes the number of misidentified particles.
    \item The kinematic overlap factor, $\mathcal{K}$, is set to $\mathcal{K}=0.1^{|N_{miss}-N_{\neutrino}^s|}$, where $N_{miss}$ is the number of non-reconstructed particles in the background, and $N_{\neutrino}^s$ the number of neutrinos in the signal.\\
\end{itemize}

At the level of complexity considered in the toy scenario, recursively parsing the PDG database is still computationally feasible, allowing for the obtainment of the true most relevant backgrounds and the evaluation of the algorithm's performance.\\

\section{Results}\label{Results}

\subsection{GA performance}

\begin{figure*}[htbp]
  \centering
  \includegraphics[width=0.7\textwidth]{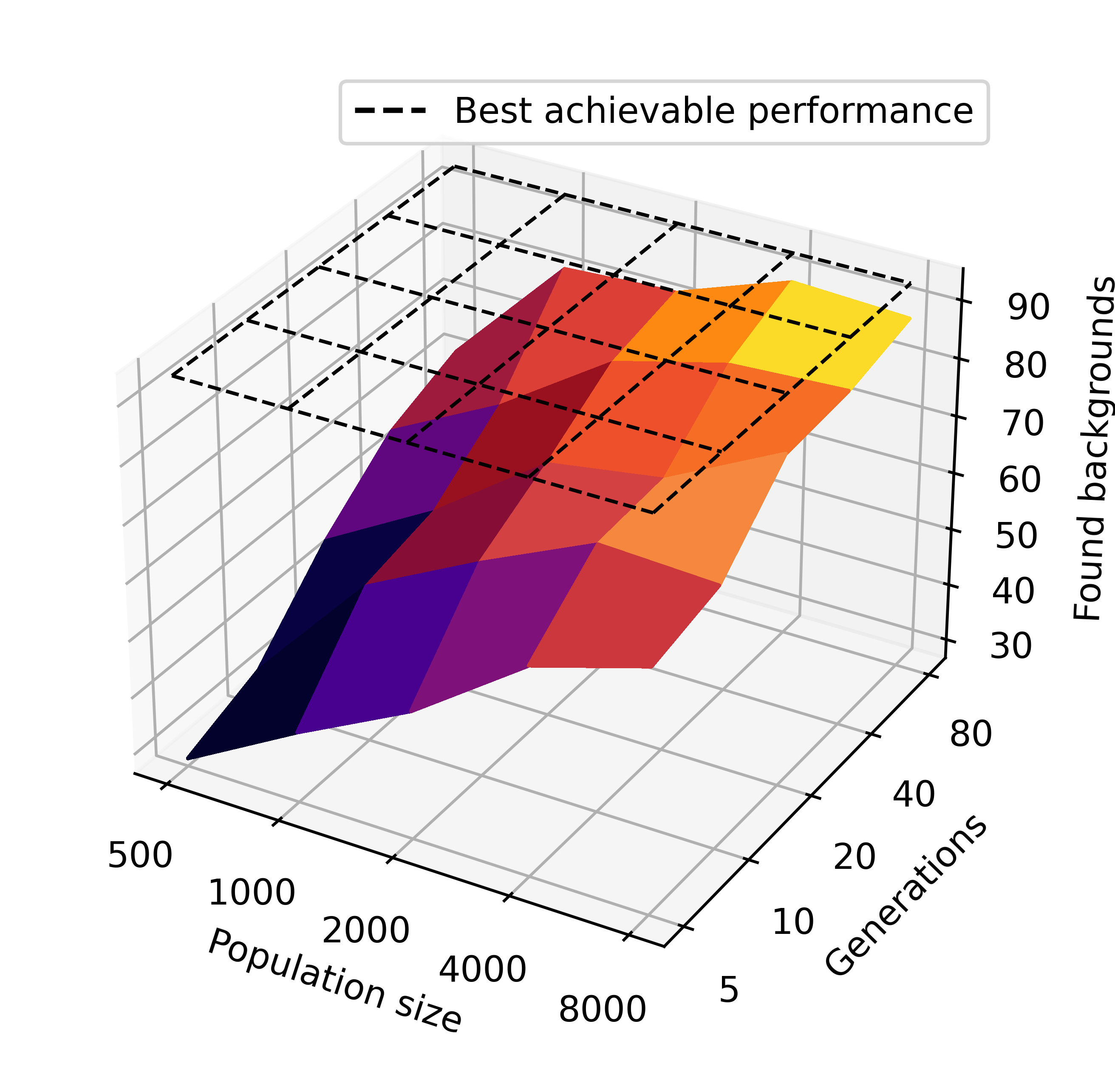}
  \caption{GA performance as a function of population size and number of generations. The results show a consistent increase in performance with larger populations and longer evolutionary runs.}
  \label{GA_performance}
\end{figure*}

We conducted an initial experiment to evaluate the effectiveness of Genetic Algorithms (GAs) to identify the backgrounds associated with the training signals. Based on the toy model reward function, \numbertrainingbackgroundsnumeric\ backgrounds are considered relevant. To assess GA performance, defined as the number of relevant backgrounds identified, we performed multiple runs using varying population sizes and numbers of generations. The results, presented in Figure \ref{GA_performance}, demonstrate a clear upward trend in performance as either hyperparameter increases\footnote{Background processes may contain more final state particles than the signal if they involve non-reconstructed particles. For a more efficient search, instead of allowing the number of positive, negative and neutral final state particles in individuals to vary freely, we fixed these counts for the entire population. As a consequence, the process of finding the relevant backgrounds for a signal will in reality involve running multiple GA evolutions, each with a different configuration of these counts. For the signals in the considered training set, the number of evolutions is 3 or 4 depending on the case. The reported population sizes refer to those used in each of these evolutions.}. This trend is expected: larger populations enhance genetic diversity, while more generations give the algorithm additional opportunities to explore the space. In Appendix \ref{GA_space_size}, we compute the GA space size for an example signal to be around $3.3\cdot10^8$. The results also indicate that, even with a total number of evaluations that is much smaller than the GA space size, the algorithm successfully identifies the majority of backgrounds relevant to the training signals.\\

\subsection{GA-assisted RL training}\label{GA_RL_training}

\begin{figure}[htbp]
    \centering
    \subfigure[Training performance]{
        \includegraphics[width=0.7\textwidth]{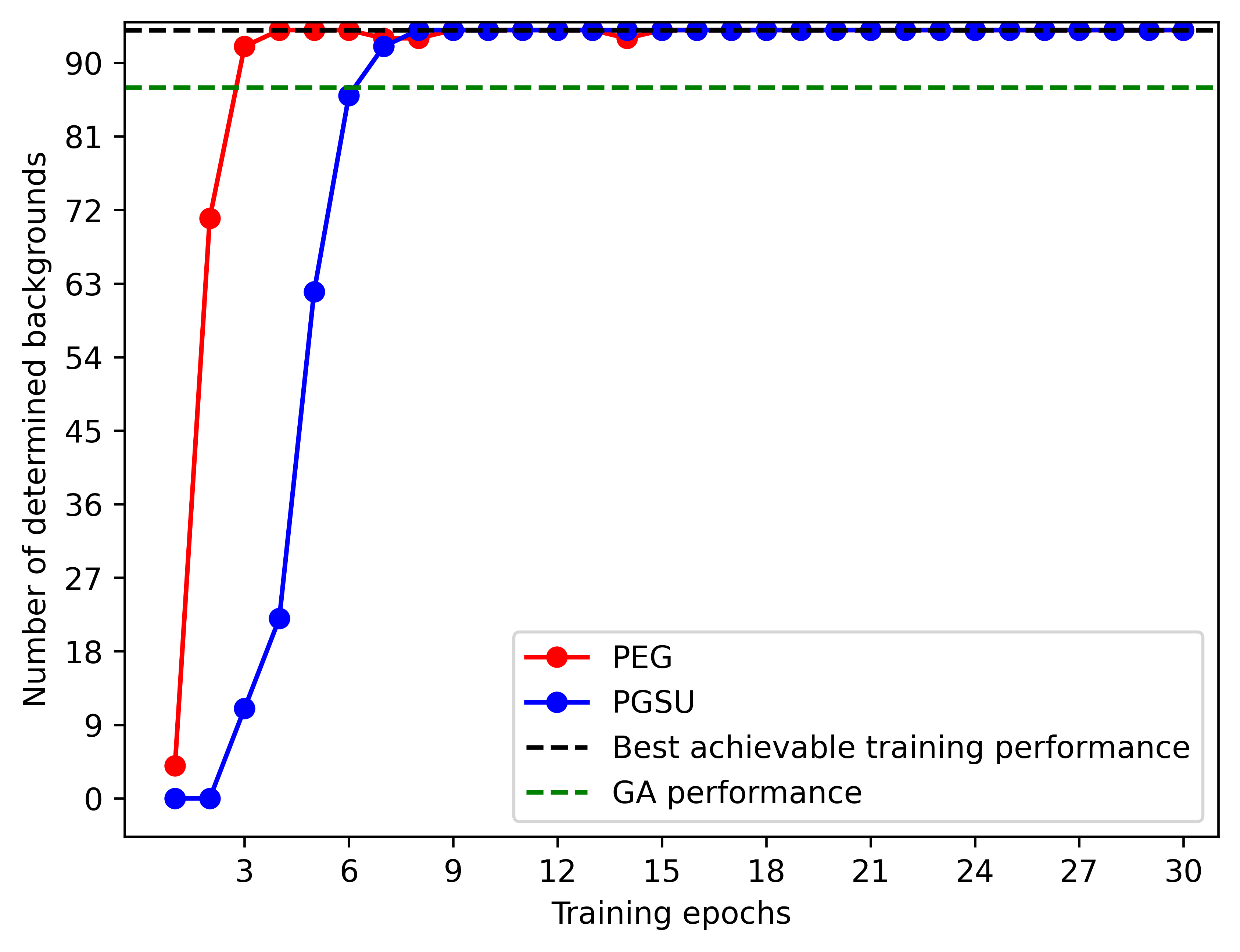}
    }
    \vfill
    \subfigure[Generalisation performance]{
        \includegraphics[width=0.7\textwidth]{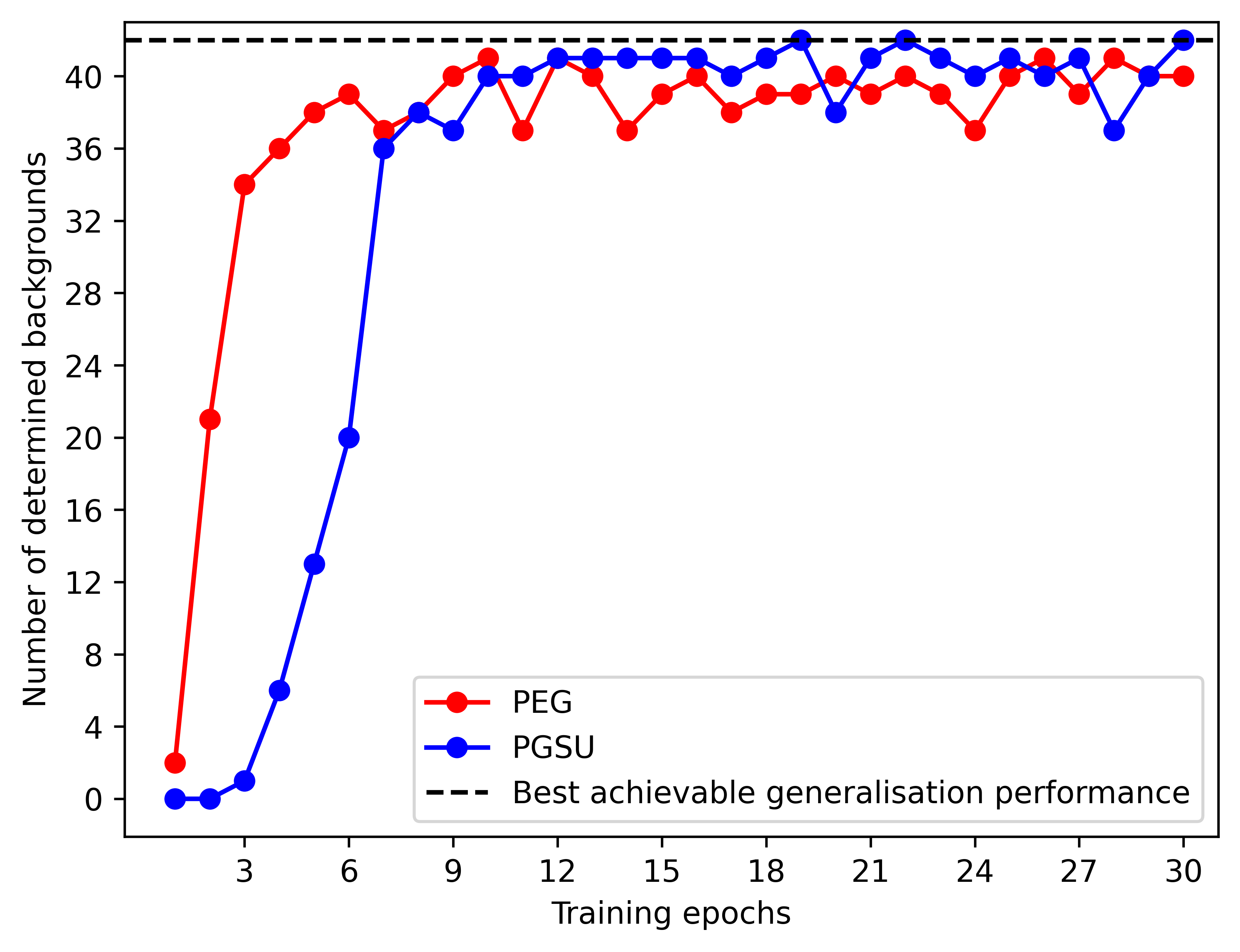}
    }
    \caption{Learning curves illustrating the agent's performance in terms of the number of correctly determined backgrounds for the training (a) and generalisation (b) signals. The blue curve corresponds to the PGSU approach, while the red curve corresponds to PEG, with both methods using GA-discovered trajectories as expert demonstrations. Performance is evaluated over 100,000 episodes. The horizontal dotted green line indicates the number of backgrounds identified by GAs, while the black dotted line represents the best achievable performance.}
    \label{training_generalisation_performance}
\end{figure}

We performed a second experiment following the GA-assisted RL strategies described in Section \ref{RL_background_determination}. We performed two different trainings, one following the PEG approach and the other using PGSU. The hyperparameters utilised to train the agent are summarised in Table \ref{hyperparameters}. For the GAs, we employed a population of 6,000 individuals and 40 generations of evolution. Regarding the RL hyperparameters, in both tests we trained the agent for 30 epochs, each consisting of 4,500 episodes and 5 training iterations per epoch. We used a small transformer model with 360,372 parameters for the agent. In the PEG approach, we assigned a probability of 5\% for the agent to play episodes on GA-discovered expert trajectories, whereas for PGSU we used a value of $\lambda=12.8$. The learning curves, presented in Figure \ref{training_generalisation_performance}, reflect performance evaluated over 100,000 episodes, measuring how many relevant backgrounds associated with the training and generalisation signals are correctly determined. To enhance the diversity of solutions during background determination, we applied a temperature higher than that used during training.\\   

The results demonstrate that, although GAs identify \GAperformance\ out of \numbertrainingbackgroundsnumeric\ training backgrounds, in both approaches the RL agent is able to explore beyond these solutions and achieve beyond-expert training performance. This scenario reflects the philosophy behind the synergy: without GA assistance, the agent's performance remains flat at 0 out of \numbertrainingbackgroundsnumeric\ backgrounds due to the highly sparse reward configuration and the difficulty in discovering relevant backgrounds (results omitted for brevity). In contrast, GA assistance effectively guides the agent during training without limiting its ability to discover novel solutions. Both PEG and PGSU show an increasing trend in training and generalisation performance across epochs. In both cases, training performance plateaus at approximately \trainingperformance\ out of \numbertrainingbackgroundsnumeric\ backgrounds. For generalisation, while PEG achieves comparable performance, PGSU achieves superior performance, plateauing at an average of \generalisationperformance\ out \numbergeneralisationbackgroundsnumeric\ backgrounds. Nevertheless, as expected, PEG exhibits a faster initial performance improvement, as discussed in Section \ref{RL_background_determination}.\\

{
\renewcommand\arraystretch{1.3} % only affects this table
\begin{table}[htbp]
    \centering
    \begin{tabular}{|P{1.5cm}|P{8cm}|P{2.5cm}|}%{|c|c|c|}
        \hline
        & Hyperparameter & Value\\ \hline \hline
        \multirow{2}{*}{\rotatebox[origin=c]{90}{GA}} & Population size & 6,000 \\ 
        & Evolutions & 40 \\ \hline
        \multirow{18}{*}{\rotatebox[origin=c]{90}{RL}} & Model parameters & 360,372 \\
        & $\alpha$ & 0.5 \\
        & $k$ & 1 \\
        & Truncation penalisation & -0.3 \\
        & Epochs & 30 \\
        & Training episodes/epoch & 4,500 \\
        & Training iterations/epoch & 5 \\
        & Batch size & 128 \\ 
        & Optimiser & Adam \\
        & Learning rate & 0.001 \\
        & Weight decay & 0.0001 \\
        & Searches (MCTS) & 1,000 \\ 
        & Dirichlet noise weight (MCTS) & 0.25 \\
        & Dirichlet alpha (MCTS) & 0.3 \\
        & Exploration constant (MCTS) & 2 \\
        & Temperature in training & 1.25 \\
        & Episodes for background determination & 100,000 \\
        & Temperature in background determination & 2.5 \\ \hline
        PEG & GA assistance probability & 5\% \\ \hline
        PGSU & $\lambda$ & 12.8 \\ \hline

    \end{tabular}
    \caption{Hyperparameters employed for the training of the agents.}
    \label{hyperparameters}
\end{table}
}

\subsection{Fine tuning of agent}

\begin{figure}[htbp]
    \centering
    \subfigure[Training performance]{
        \includegraphics[width=0.7\textwidth]{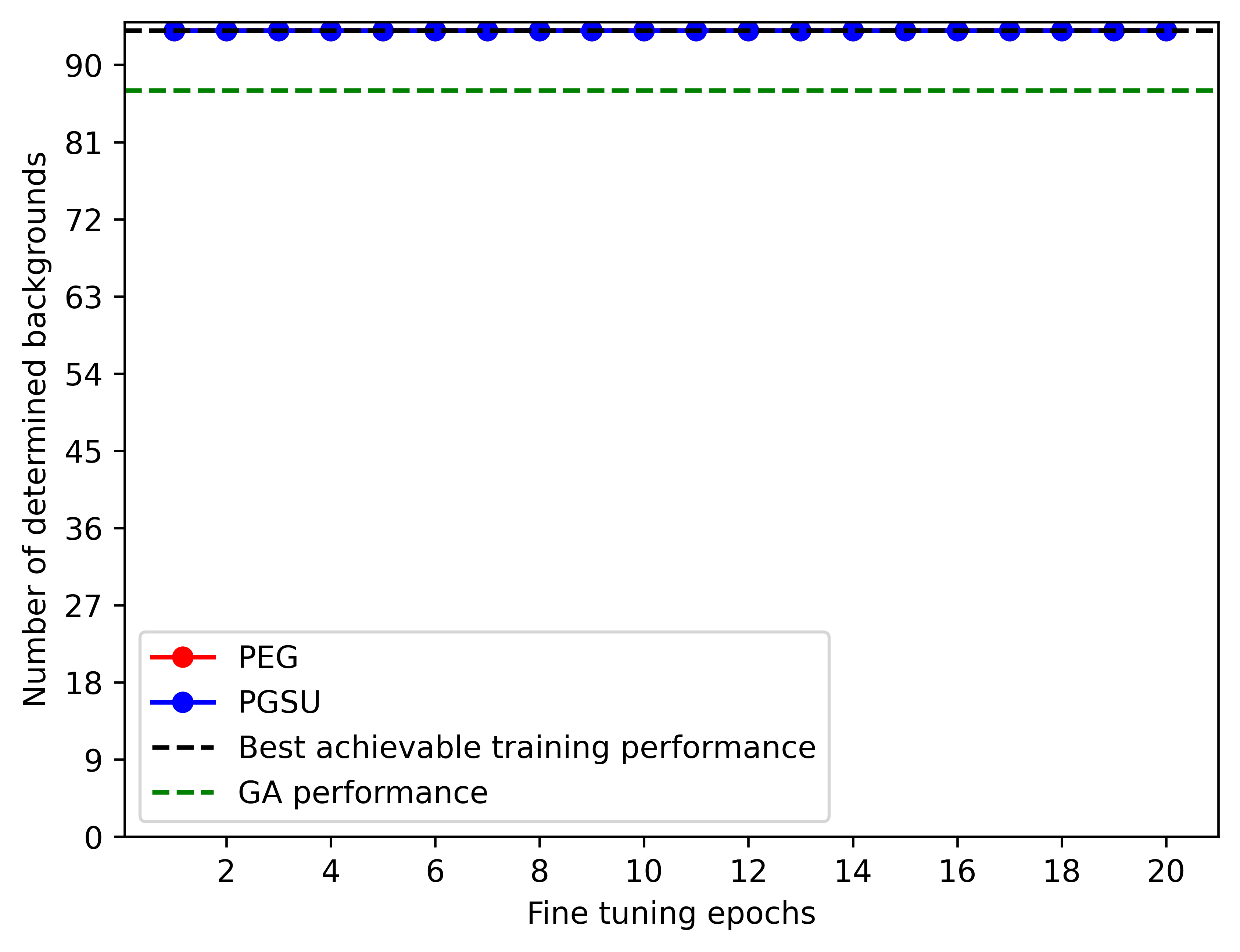}
    }
    \vfill
    \subfigure[Generalisation performance]{
        \includegraphics[width=0.7\textwidth]{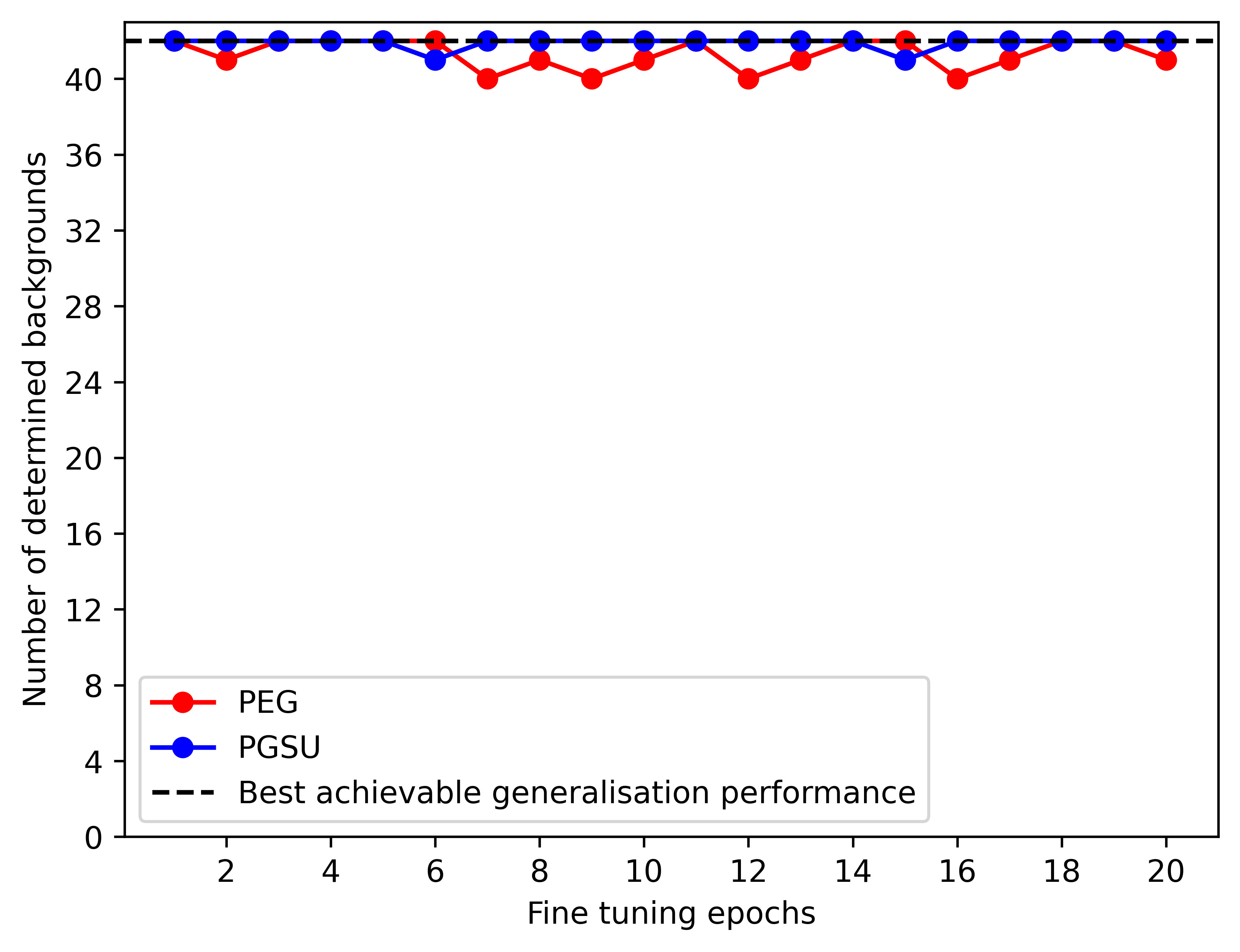}
    }
    \caption{Learning curves illustrating the agent's performance in terms of the number of correctly determined backgrounds for the training (a) and generalisation (b) signals after the agent is fine tuned. The blue curve corresponds to the PGSU approach, while the red curve corresponds to PEG, with both methods using relevant backgrounds identified during training as expert demonstrations. Performance is evaluated over 100,000 episodes. The horizontal dotted green line indicates the number of backgrounds identified by GAs, while the black dotted line represents the best achievable performance.}
    \label{finetuned_training_generalisation_performance}
\end{figure}

To further improve the performance of the agents trained in Section \ref{GA_RL_training}, we performed fine tuning following the strategies described in Section \ref{fine_tuning}. The hyperparameters employed are summarised in Table \ref{hyperparameters_finetuning}, and the results are presented in Figure \ref{finetuned_training_generalisation_performance}. After fine-tuning, the agent fine tuned with PGSU exhibits robust results, achieving a training performance of \finetunedtrainingperformance\ out of \numbertrainingbackgroundsnumeric\ backgrounds, and a generalisation performance of approximately \finetunedgeneralisationperformance\ out of \numbergeneralisationbackgroundsnumeric\ backgrounds. For the agent fine tuned with PEG, all relevant training backgrounds are also determined (both PEG and PGSU curves overlap), although its generalisation performance is slightly lower.\\

{
\renewcommand\arraystretch{1.3} % only affects this table
\begin{table}[htbp]
    \centering
    \begin{tabular}{|P{2cm}|P{8cm}|P{2.5cm}|}%{|c|c|c|}
        \hline
         & Hyperparameter & Value\\ \hline \hline
         \multirow{14}{*}{\rotatebox[origin=c]{90}{RL}} & Truncation penalisation & -0.3 \\
         & Epochs & 20 \\
         & Training episodes/epoch & 80 \\
         & Training iterations/epoch & 5 \\
         & Batch size & 128 \\ 
         & Optimiser & Adam \\
         & Learning rate & 0.001 \\
         & Weight decay & 0.0001 \\
         & Searches (MCTS) & 1,000 \\ 
         & Dirichlet noise weight (MCTS) & 0.25 \\
         & Dirichlet alpha (MCTS) & 0.3 \\
         & Exploration constant (MCTS) & 2 \\
         & Episodes for background determination & 100,000 \\
         & Temperature in background determination & 2.5 \\ \hline
         PEG & Assistance probability & 100\% \\ \hline 
         \multirow{2}{*}{\rotatebox[origin=c]{90}{PGSU}} & $\lambda$ & 12.8 \\ 
         & Temperature in fine tuning & 1.25 \\ \hline
    \end{tabular}
    \caption{Hyperparameters employed for the fine tuning of the agents.}
    \label{hyperparameters_finetuning}
\end{table}
}

\subsection{Visualisation of learnt embeddings}

\begin{figure*}[htbp]
  \centering
  \includegraphics[width=0.8\textwidth]{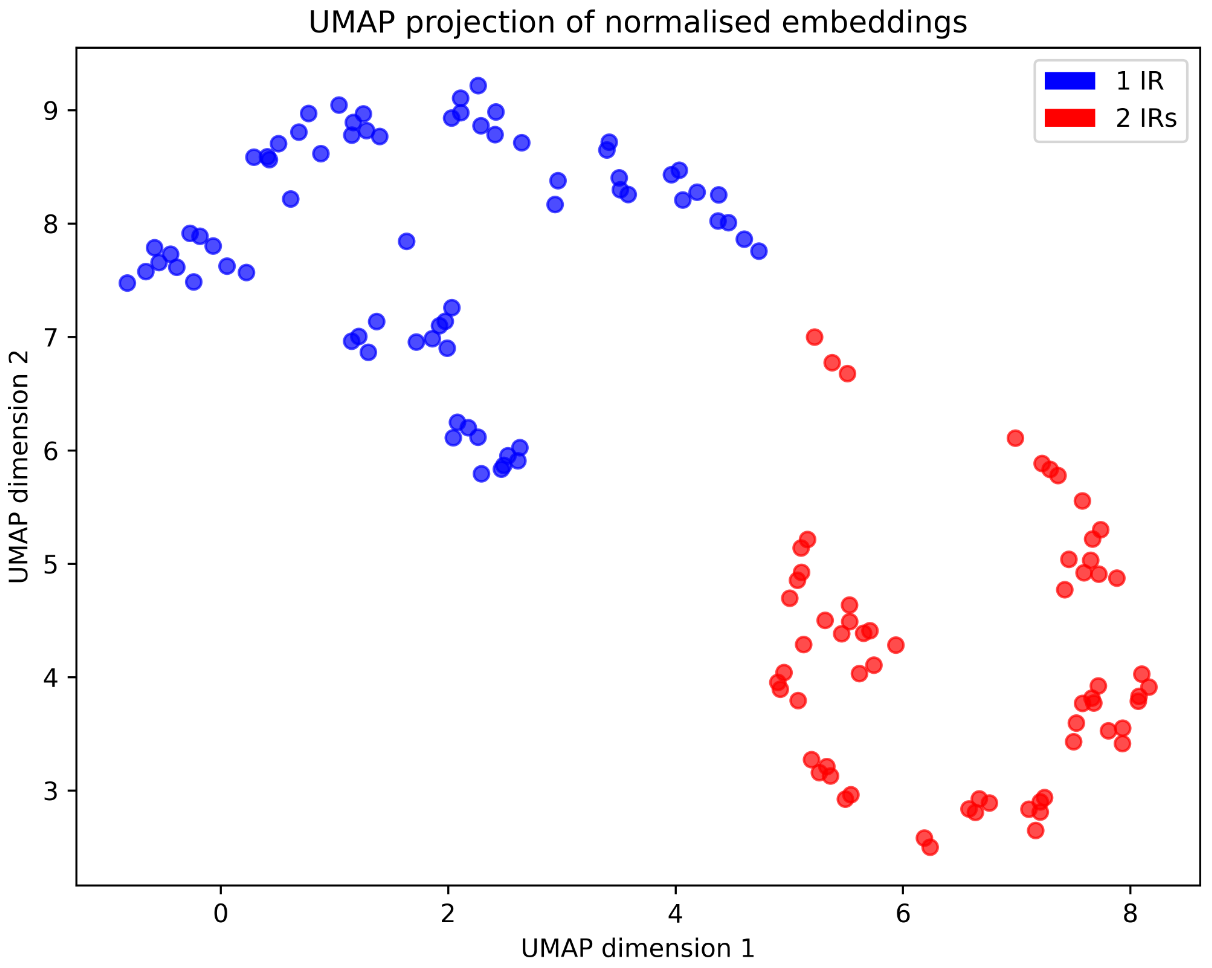}
  \caption{UMAP visualization of the L2-normalised transformer embeddings, where each point corresponds to a signal–background pair. Points are coloured according to the number of intermediate resonances in the background: blue for 1, and red for 2. In the toy scenario considered, no relevant backgrounds with zero intermediate resonances are present. The model learnt distinct embeddings for backgrounds with 1 and 2 intermediate resonances, with blue points clustering on the left side of the plot and red points grouping on the right.}
  \label{UMAP_IRs}
\end{figure*}

\begin{figure*}[htbp]
  \centering
  \includegraphics[width=1.0\textwidth]{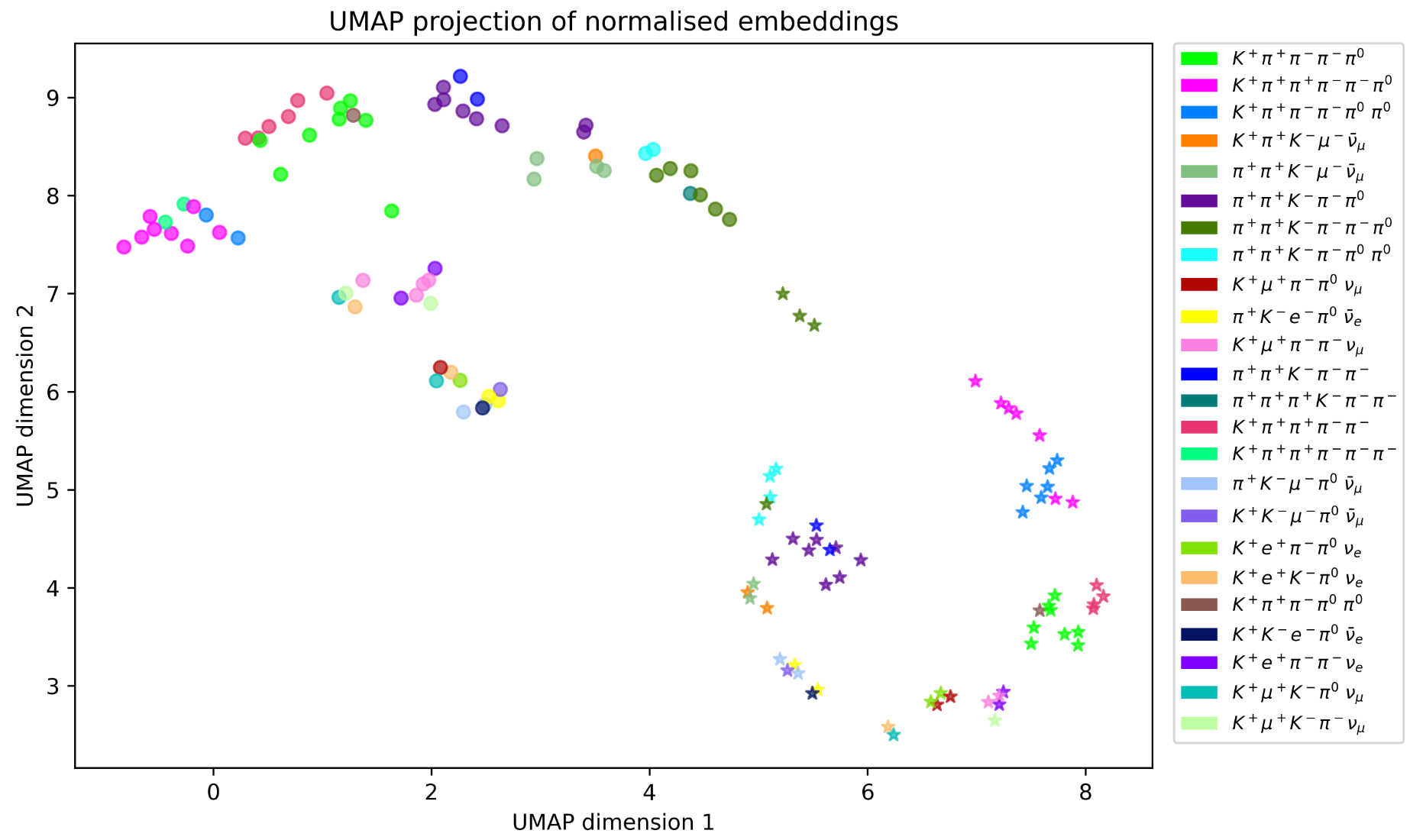}
  \caption{UMAP visualization of the L2-normalised transformer embeddings, where each point corresponds to a signal–background pair. Points are coloured by the configuration of final state particles present in the background. Points with the same final state particles tend to form distinct clusters, with one cluster corresponding to backgrounds containing 1 intermediate resonance (represented with dots) and another corresponding to backgrounds with 2 (represented with stars).}
  \label{UMAP_FSPs}
\end{figure*}

To gain insight into the internal representations learnt by the model, we analyse the embeddings produced by the shared backbone of the transformer model. For each signal–background pair under study, we define the embedding as the output of the transformer layer immediately before the network branches into the policy and value heads. We visualise these embeddings using Uniform Manifold Approximation and Projection (UMAP) \cite{UMAP}, a dimensionality reduction technique designed to preserve the structure of the data as much as possible. Prior to projection, the embeddings are L2-normalised to ensure uniform scaling. Results are presented in Figures \ref{UMAP_IRs} and \ref{UMAP_FSPs}.\\

Figure \ref{UMAP_IRs} shows the embedding space coloured by the number of intermediate resonances in the background. For the signals considered in the toy scenario, the relevant backgrounds contain either 1 or 2 intermediate resonances. The model learnt distinct embeddings for these two cases: backgrounds with 1 intermediate resonance (blue) form a cluster predominantly on the left, while those with 2 (red) group on the right.\\

Figure \ref{UMAP_FSPs} shows the same embedding space, now coloured according to the final state particles present in the background. Backgrounds with the same particle configuration tend to form clusters, with these clusters aligning with resonance structure: backgrounds with 1 intermediate resonance (represented with dots) group separately from those with 2 (represented with stars). This suggests that the model captures both structural and compositional differences in its learnt representations.\\

\section{Discussion}

For the experiments in this work, we developed a toy scenario with 40 particles, using B-mesons as mother particles and experimentally well-established intermediate states, to provide realistic and sufficiently complex decay chains for evaluating the performance of the proposed algorithm. We conducted a first experiment to assess the performance of the GA, where results demonstrate that the GA can successfully identify most of the backgrounds for the training signals, despite using a much smaller number of evaluations than the size of the GA search space.\\

By employing the GA-discovered backgrounds as expert demonstrations, we trained and fine-tuned two agents, one following the PGSU approach and the other using PEG. The results indicate that both methods surpass expert performance and achieve perfect training performance\footnote{In this context, perfect performance corresponds to correctly determining every background.}. PGSU exhibits robust performance with slightly superior generalisation ability. In contrast, PEG achieves comparable results while providing faster initial performance improvement, as we only added a supervised term to the loss in Equation \ref{total_loss} for states along expert trajectories. This requires the agent to learn how to reach a state before it can receive supervision for that state. Consequently, PEG could be an interesting alternative for environments with very long episodes, where PGSU may result in a slower learning.\\

Finally, we analysed the embeddings produced by the transformer to gain insight into the internal representations learnt by the model. We visualised the embeddings using UMAP \cite{UMAP}. The results show that backgrounds sharing the same final state particles tend to form clusters, with backgrounds containing one intermediate resonance grouping separately from those with two. The results imply that the model's internal representations distinguish backgrounds based on both their structure and particle composition.\\

One of the most significant outcomes of this work is that it opens the door to a systematic strategy for accelerating and automating background determination in particle physics analyses. Traditionally, this task requires the physicist's expert knowledge and manual inspection. However, by introducing an algorithmic approach based on ML, the procedure can be standardised, scaled to a wide range of analyses, and potentially integrated into large experimental pipelines.\\

The results presented in this work should be regarded as a proof of concept, demonstrating the feasibility of employing different AI and ML techniques to automate background determination. While the toy scenario employed here provides a controlled environment for testing the method, the encouraging performance of the proposed framework highlights its potential to be extended and adapted to real-world applications.\\

Nevertheless, several key challenges remain before such a framework can be deployed in realistic scenarios. First, in our experiments the training was performed using a small set of signal decays, and focused on a subset of 40 particles with constraints on the number of allowed intermediate resonances and misidentifications. While this setup provides a manageable and sufficiently complex environment, it limits the coverage of the space of possible background processes. In practice, training should be conducted with a much larger and more diverse set of signals to allow the agent to learn generalisable representations across the full decay space. Second, the toy model reward employed in our experiments was sufficient to validate the concept, but represents only a simplification of the complexities present in a realistic physics analysis. Future developments should explore more sophisticated, data-driven approaches to extract the reward, enabling the framework to handle realistic physics observables.\\

Furthermore, the GA component of our framework can be refined to further enhance its efficiency. In particular, the incorporation of additional conservation laws beyond charge conservation would constrain the search space and more effectively guide the exploration toward physically valid decay processes.\\

Taken together, these considerations highlight both the promise and the necessary next steps of this research direction. The methodology proposed here demonstrates that combining RL and GAs provides a powerful mechanism for addressing the highly-sparse-reward and large-search-space challenges inherent in background determination. With further refinement, including scaling to larger datasets, employing a more realistic reward function than the simplified one used here, and incorporating additional physics constraints into the algorithm, this approach could develop into a practical tool for high-energy physics analyses.\\

\section{Conclusions}

In this work, we have introduced a novel AI-driven method to systematically determine relevant background processes in particle physics analyses, with beauty hadron decays serving as a case study. This problem, traditionally addressed through expert intuition and manual inspection, is both critical and complex due to the vast number of potential background channels, their overlapping kinematic signatures, and the computational cost of simulating each process. Our approach aims to replace this procedure with an algorithmic approach that can be generalised and scaled to a wide range of analyses. By automating the background determination process, the framework not only accelerates the analysis workflow but also reduces the reliance on subjective judgment, thereby mitigating the risk of human error.\\

The core idea of the approach is to formulate the problem as a sequential decision-making task, enabling the use of RL to explore the space of possible background processes. Background decays are represented as sequences of tokens, and each action taken by the RL agent corresponds to selecting a token, allowing the agent to construct candidate background decays token by token. To effectively handle the symbolic nature of particle decays, we employ a transformer-based architecture \cite{transformer} for the RL agent, enabling it to capture dependencies within decay sequences. The agent is trained on a set of signal decays, where it learns to infer the most relevant background processes through interaction with the environment. Once trained, the agent employs its generalisation ability to determine relevant backgrounds for previously unseen signal decays.\\

To address the challenges of a large trajectory space and episodes where feedback is provided only at the end, we utilise AlphaZero algorithm \cite{AlphaZero}, which we adapt to a single-player setting. We selected AlphaZero specifically because it has demonstrated remarkable performance in two-player games such as chess, shogi, and Go, which share these same challenges. Despite its success in such domains, AlphaZero has seen limited application in physics research and, to our knowledge, no prior application in particle physics. This work thus represents a novel application of AlphaZero-inspired techniques to this field.\\

A key additional challenge in applying RL to this domain is the high sparsity of rewards, meaning that successful trajectories in which valid background processes are constructed are rare. To address this, we propose a novel algorithm that exploits the RL-GA synergy, in which GAs are employed to discover successful action trajectories in the environment, which serve as expert demonstrations for the RL agent, significantly improving learning efficiency and exploration. In contrast to traditional GAs, which typically employ a vector-based representation for individual genes, we adopt a tree-based structure that more naturally captures the hierarchical nature of decay chains. Beyond the standard variation operators (crossover and mutation), we introduced customised variation mechanisms tailored to specific objectives, including the construction of intermediate resonances, the improvement of exploration efficiency through the integration of physical properties, and the ability to reuse previously acquired knowledge for future optimisation problems.\\

We explored two distinct approaches for leveraging GA-discovered demonstrations. The first, based on PGSU, combines the policy gradient loss with a supervised imitation loss derived from expert trajectories. The second, which we refer to as PEG, applies action masking to periodically constrain the agent to follow full expert trajectories during training. Since GA-generated trajectories can fail to capture all relevant backgrounds for the training signals, both approaches are designed to encourage exploration beyond the demonstrations, allowing the agent to achieve performance that surpasses the expert level. Finally, we have also employed other techniques such as action masking, reward shaping, and a fine-tuning of the agent to enhance the algorithm’s performance.\\

The experiments conducted in this work demonstrate the effectiveness of the proposed algorithm. The GA efficiently explores and optimises the vast space of possible background processes, successfully identifying the majority of relevant backgrounds while requiring significantly fewer evaluations than the size of the full search space. By leveraging these GA-discovered backgrounds as expert demonstrations, both the PGSU and PEG techniques surpass expert-level performance, with PGSU achieving slightly superior generalisation ability, and PEG providing faster initial performance improvement. Furthermore, the transformer-based agent learns meaningful internal representations, with embeddings clustering according to particle composition and intermediate resonance structure. Together, these results validate the efficiency and learning capability of the proposed algorithm and highlight its potential to generalise beyond the training signals, providing a promising strategy for background determination in particle physics.\\

While the presented proof-of-concept experiments provide valuable insights, several limitations should be acknowledged. The training was conducted on a restricted set of particles, with constraints on the number of allowed intermediate resonances and misidentifications, and utilising a simplified reward model. This setup constrains the range of background processes considered, and simplifies complexities of real analyses. Nevertheless, the promising results highlight the framework's potential to be expanded and applied to real-world scenarios, and demonstrate the effectiveness of the RL-GA synergy for addressing environments characterised by highly sparse rewards and a large trajectory space.\\

\section{Conflicts of Interest}
On behalf of all the authors, the corresponding author declares that there is no conflict of interest.

\section{Data availability statement}

This work makes use of the $\texttt{DECAY\_LHCB.dec}$ file provided by the DecayLanguage library \cite{DecayLanguage}, which specifies decay channels and their BRs. The BRs included in this file are based on measurements compiled by the PDG \cite{PDG}. No additional datasets were generated or analysed in this study.

\section{Funding statement}

The UZH group acknowledges support from the Swiss National Science Foundation (SNSF) under grants 209263 and 204238, and from Forschubngskredit under the UZH Candoc Grant no. FK-25-086. The Imperial College group acknowledges support from the STFC (United Kingdom) and The Leverhulme Trust (United Kingdom). The University of Bristol group acknowledges support from the STFC under grant ST/W000490/1.\\

\clearpage

\begin{appendices}

\section{Relevant backgrounds by signal}\label{signals_and_backgrounds}

This appendix presents the relevant backgrounds for each of the training and generalisation signals, along with the corresponding rewards after logarithmic transformation (see Equation \ref{reward_shaping1}).\\

\noindent\textbf{\large • Relevant backgrounds for training signals:}\\[1em]
\noindent\textbf{\ding{93}\ Signal: $B^0\rightarrow \pi^+ D^{*-}( \pi^- \bar{D}^0( K^+ \pi^- \pi^0 ) )$}\\[0.5em]
\begin{longtable}{|P{4.0cm}|P{6.5cm}|P{2.5cm}|}
\hline
\rule{0pt}{1.1em}\textbf{Background index} & \textbf{Background} & $r$ \\
\hline
\endfirsthead
\hline
\rule{0pt}{1.1em}\textbf{Background index} & \textbf{Background} & $r$ \\
\hline
\endhead
\rule{0pt}{1.1em} No. 1 & $B^0\rightarrow \pi^+ \pi^0 D^{*-}( \pi^- \bar{D}^0( K^+ \pi^- ) )$ & 0.605 \\ \hline
\rule{0pt}{1.1em} No. 2 & $B^0\rightarrow \pi^+ \pi^- \bar{D}^0( K^+ \pi^- \pi^0 )$ & 0.371 \\ \hline
\rule{0pt}{1.1em} No. 3 & $B^+\rightarrow \pi^+ \pi^+ \pi^- \bar{D}^0( K^+ \pi^- \pi^0 )$ & 0.31 \\ \hline
\rule{0pt}{1.1em} No. 4 & $B^0\rightarrow \pi^+ D^-( \pi^- \pi^0 K^{*0}( K^+ \pi^- ) )$ & 0.281 \\ \hline
\rule{0pt}{1.1em} No. 5 & $B^0\rightarrow \pi^+ D^{*-}( \pi^0 D^-( K^+ \pi^- \pi^- ) )$ & 0.267 \\ \hline
\rule{0pt}{1.1em} No. 6 & $B^0\rightarrow \pi^+ \pi^0 D^{*-}( \pi^- \bar{D}^0( K^+ \pi^- \pi^0 ) )$ & 0.26 \\ \hline
\rule{0pt}{1.1em} No. 7 & $B^0\rightarrow \pi^+ \pi^0 D^-( K^+ \pi^- \pi^- )$ & 0.166 \\ \hline
\rule{0pt}{1.1em} No. 8 & $B^+\rightarrow \pi^+ \pi^+ \pi^0 D^{*-}( \pi^- \bar{D}^0( K^+ \pi^- ) )$ & 0.142 \\ \hline
\rule{0pt}{1.1em} No. 9 & $B^0\rightarrow \pi^+ \pi^- \pi^0 \bar{D}^0( K^+ \pi^- )$ & 0.14 \\ \hline
\rule{0pt}{1.1em} No. 10 & $B^0\rightarrow \pi^+ \pi^0 D^{*-}( \pi^0 D^-( K^+ \pi^- \pi^- ) )$ & 0.087 \\ \hline
\rule{0pt}{1.1em} No. 11 & $B^0\rightarrow \pi^+ D^-( K^+ \pi^- \pi^- \pi^0 )$ & 0.084 \\ \hline
\rule{0pt}{1.1em} No. 12 & $B^+\rightarrow \pi^+ \pi^+ \pi^0 D^-( K^+ \pi^- \pi^- )$ & 0.07 \\ \hline
\rule{0pt}{1.1em} No. 13 & $B^0\rightarrow \pi^+ \pi^- \bar{D}^{*0}( \pi^0 \bar{D}^0( K^+ \pi^- ) )$ & 0.056 \\ \hline
\rule{0pt}{1.1em} No. 14 & $B^0\rightarrow \pi^+ \pi^- \pi^0 \bar{D}^0( K^+ \pi^- \pi^0 )$ & 0.052 \\ \hline
\end{longtable}
\noindent\textbf{\ding{93}\ Signal: $\bar{B}^0\rightarrow \mu^- \bar{\nu}_{\mu} D^+( K^+ \pi^+ K^- )$}\\[0.5em]
\begin{longtable}{|P{4.0cm}|P{6.5cm}|P{2.5cm}|}
\hline
\rule{0pt}{1.1em}\textbf{Background index} & \textbf{Background} & $r$ \\
\hline
\endfirsthead
\hline
\rule{0pt}{1.1em}\textbf{Background index} & \textbf{Background} & $r$ \\
\hline
\endhead
\rule{0pt}{1.1em} No. 1 & $\bar{B}^0\rightarrow \mu^- \bar{\nu}_{\mu} D^{*+}( \pi^+ D^0( K^+ K^- ) )$ & 1.139 \\ \hline
\rule{0pt}{1.1em} No. 2 & $\bar{B}^0\rightarrow \mu^- \bar{\nu}_{\mu} D^+( K^+ \bar{K}^{*0}( \pi^+ K^- ) )$ & 0.414 \\ \hline
\rule{0pt}{1.1em} No. 3 & $\bar{B}^0\rightarrow \mu^- \bar{\nu}_{\mu} D^+( \pi^+ \pi^+ K^- )$ & 0.287 \\ \hline
\rule{0pt}{1.1em} No. 4 & $\bar{B}^0\rightarrow \pi^+ \mu^- \bar{\nu}_{\mu} D^0( K^+ K^- )$ & 0.242 \\ \hline
\rule{0pt}{1.1em} No. 5 & $\bar{B}^0\rightarrow \mu^- \bar{\nu}_{\mu} D^{*+}( \pi^+ D^0( \pi^+ K^- ) )$ & 0.19 \\ \hline
\end{longtable}
\noindent\textbf{\ding{93}\ Signal: $\bar{B}^0\rightarrow \pi^- \pi^0 D^{*+}( \pi^+ D^0( \pi^+ K^- ) )$}\\[0.5em]
\begin{longtable}{|P{4.0cm}|P{6.5cm}|P{2.5cm}|}
\hline
\rule{0pt}{1.1em}\textbf{Background index} & \textbf{Background} & $r$ \\
\hline
\endfirsthead
\hline
\rule{0pt}{1.1em}\textbf{Background index} & \textbf{Background} & $r$ \\
\hline
\endhead
\rule{0pt}{1.1em} No. 1 & $\bar{B}^0\rightarrow \pi^- D^{*+}( \pi^+ D^0( \pi^+ K^- \pi^0 ) )$ & 0.79 \\ \hline
\rule{0pt}{1.1em} No. 2 & $\bar{B}^0\rightarrow \pi^+ \pi^- D^0( \pi^+ K^- \pi^0 )$ & 0.432 \\ \hline
\rule{0pt}{1.1em} No. 3 & $B^-\rightarrow \pi^+ \pi^- \pi^- D^0( \pi^+ K^- \pi^0 )$ & 0.363 \\ \hline
\rule{0pt}{1.1em} No. 4 & $\bar{B}^0\rightarrow \pi^- D^+( \pi^+ \pi^0 \bar{K}^{*0}( \pi^+ K^- ) )$ & 0.329 \\ \hline
\rule{0pt}{1.1em} No. 5 & $\bar{B}^0\rightarrow \pi^- D^{*+}( \pi^0 D^+( \pi^+ \pi^+ K^- ) )$ & 0.314 \\ \hline
\rule{0pt}{1.1em} No. 6 & $\bar{B}^0\rightarrow \pi^- \pi^0 D^{*+}( \pi^+ D^0( \pi^+ K^- \pi^0 ) )$ & 0.306 \\ \hline
\rule{0pt}{1.1em} No. 7 & $\bar{B}^0\rightarrow \pi^- \pi^0 D^+( \pi^+ \pi^+ K^- )$ & 0.197 \\ \hline
\rule{0pt}{1.1em} No. 8 & $B^-\rightarrow \pi^- \pi^- \pi^0 D^{*+}( \pi^+ D^0( \pi^+ K^- ) )$ & 0.168 \\ \hline
\rule{0pt}{1.1em} No. 9 & $\bar{B}^0\rightarrow \pi^+ \pi^- \pi^0 D^0( \pi^+ K^- )$ & 0.166 \\ \hline
\rule{0pt}{1.1em} No. 10 & $\bar{B}^0\rightarrow \pi^- \pi^0 D^{*+}( \pi^0 D^+( \pi^+ \pi^+ K^- ) )$ & 0.104 \\ \hline
\rule{0pt}{1.1em} No. 11 & $\bar{B}^0\rightarrow \pi^- D^+( \pi^+ \pi^+ K^- \pi^0 )$ & 0.1 \\ \hline
\rule{0pt}{1.1em} No. 12 & $B^-\rightarrow \pi^- \pi^- \pi^0 D^+( \pi^+ \pi^+ K^- )$ & 0.083 \\ \hline
\rule{0pt}{1.1em} No. 13 & $\bar{B}^0\rightarrow \pi^+ \pi^- D^{*0}( \pi^0 D^0( \pi^+ K^- ) )$ & 0.067 \\ \hline
\rule{0pt}{1.1em} No. 14 & $\bar{B}^0\rightarrow \pi^+ \pi^- \pi^0 D^0( \pi^+ K^- \pi^0 )$ & 0.062 \\ \hline
\rule{0pt}{1.1em} No. 15 & $B^-\rightarrow \pi^- \pi^- D^{*+}( \pi^+ D^0( \pi^+ K^- \pi^0 ) )$ & 0.057 \\ \hline
\end{longtable}
\noindent\textbf{\ding{93}\ Signal: $\bar{B}^0\rightarrow \pi^- D^{*+}( \pi^+ D^0( \pi^+ K^- \pi^0 ) )$}\\[0.5em]
\begin{longtable}{|P{4.0cm}|P{6.5cm}|P{2.5cm}|}
\hline
\rule{0pt}{1.1em}\textbf{Background index} & \textbf{Background} & $r$ \\
\hline
\endfirsthead
\hline
\rule{0pt}{1.1em}\textbf{Background index} & \textbf{Background} & $r$ \\
\hline
\endhead
\rule{0pt}{1.1em} No. 1 & $\bar{B}^0\rightarrow \pi^- \pi^0 D^{*+}( \pi^+ D^0( \pi^+ K^- ) )$ & 0.605 \\ \hline
\rule{0pt}{1.1em} No. 2 & $\bar{B}^0\rightarrow \pi^+ \pi^- D^0( \pi^+ K^- \pi^0 )$ & 0.371 \\ \hline
\rule{0pt}{1.1em} No. 3 & $B^-\rightarrow \pi^+ \pi^- \pi^- D^0( \pi^+ K^- \pi^0 )$ & 0.31 \\ \hline
\rule{0pt}{1.1em} No. 4 & $\bar{B}^0\rightarrow \pi^- D^+( \pi^+ \pi^0 \bar{K}^{*0}( \pi^+ K^- ) )$ & 0.281 \\ \hline
\rule{0pt}{1.1em} No. 5 & $\bar{B}^0\rightarrow \pi^- D^{*+}( \pi^0 D^+( \pi^+ \pi^+ K^- ) )$ & 0.267 \\ \hline
\rule{0pt}{1.1em} No. 6 & $\bar{B}^0\rightarrow \pi^- \pi^0 D^{*+}( \pi^+ D^0( \pi^+ K^- \pi^0 ) )$ & 0.26 \\ \hline
\rule{0pt}{1.1em} No. 7 & $\bar{B}^0\rightarrow \pi^- \pi^0 D^+( \pi^+ \pi^+ K^- )$ & 0.166 \\ \hline
\rule{0pt}{1.1em} No. 8 & $B^-\rightarrow \pi^- \pi^- \pi^0 D^{*+}( \pi^+ D^0( \pi^+ K^- ) )$ & 0.142 \\ \hline
\rule{0pt}{1.1em} No. 9 & $\bar{B}^0\rightarrow \pi^+ \pi^- \pi^0 D^0( \pi^+ K^- )$ & 0.14 \\ \hline
\rule{0pt}{1.1em} No. 10 & $\bar{B}^0\rightarrow \pi^- \pi^0 D^{*+}( \pi^0 D^+( \pi^+ \pi^+ K^- ) )$ & 0.087 \\ \hline
\rule{0pt}{1.1em} No. 11 & $\bar{B}^0\rightarrow \pi^- D^+( \pi^+ \pi^+ K^- \pi^0 )$ & 0.084 \\ \hline
\rule{0pt}{1.1em} No. 12 & $B^-\rightarrow \pi^- \pi^- \pi^0 D^+( \pi^+ \pi^+ K^- )$ & 0.07 \\ \hline
\rule{0pt}{1.1em} No. 13 & $\bar{B}^0\rightarrow \pi^+ \pi^- D^{*0}( \pi^0 D^0( \pi^+ K^- ) )$ & 0.056 \\ \hline
\rule{0pt}{1.1em} No. 14 & $\bar{B}^0\rightarrow \pi^+ \pi^- \pi^0 D^0( \pi^+ K^- \pi^0 )$ & 0.052 \\ \hline
\end{longtable}
\noindent\textbf{\ding{93}\ Signal: $B^+\rightarrow \mu^+ \nu_{\mu} \bar{D}^0( K^+ \pi^- \pi^0 )$}\\[0.5em]
\begin{longtable}{|P{4.0cm}|P{6.5cm}|P{2.5cm}|}
\hline
\rule{0pt}{1.1em}\textbf{Background index} & \textbf{Background} & $r$ \\
\hline
\endfirsthead
\hline
\rule{0pt}{1.1em}\textbf{Background index} & \textbf{Background} & $r$ \\
\hline
\endhead
\rule{0pt}{1.1em} No. 1 & $B^+\rightarrow \mu^+ \nu_{\mu} \bar{D}^{*0}( \pi^0 \bar{D}^0( K^+ \pi^- ) )$ & 0.361 \\ \hline
\end{longtable}
\noindent\textbf{\ding{93}\ Signal: $B^-\rightarrow e^- \bar{\nu}_e D^0( \pi^+ K^- \pi^0 )$}\\[0.5em]
\begin{longtable}{|P{4.0cm}|P{6.5cm}|P{2.5cm}|}
\hline
\rule{0pt}{1.1em}\textbf{Background index} & \textbf{Background} & $r$ \\
\hline
\endfirsthead
\hline
\rule{0pt}{1.1em}\textbf{Background index} & \textbf{Background} & $r$ \\
\hline
\endhead
\rule{0pt}{1.1em} No. 1 & $B^-\rightarrow e^- \bar{\nu}_e D^{*0}( \pi^0 D^0( \pi^+ K^- ) )$ & 0.361 \\ \hline
\end{longtable}
\noindent\textbf{\ding{93}\ Signal: $B^0\rightarrow \mu^+ \nu_{\mu} D^{*-}( \pi^- \bar{D}^0( K^+ \pi^- ) )$}\\[0.5em]
\begin{longtable}{|P{4.0cm}|P{6.5cm}|P{2.5cm}|}
\hline
\rule{0pt}{1.1em}\textbf{Background index} & \textbf{Background} & $r$ \\
\hline
\endfirsthead
\hline
\rule{0pt}{1.1em}\textbf{Background index} & \textbf{Background} & $r$ \\
\hline
\endhead
\rule{0pt}{1.1em} No. 1 & $B^0\rightarrow \mu^+ \nu_{\mu} D^-( K^+ \pi^- \pi^- )$ & 0.95 \\ \hline
\rule{0pt}{1.1em} No. 2 & $B^0\rightarrow \mu^+ \pi^- \nu_{\mu} \bar{D}^0( K^+ \pi^- )$ & 0.121 \\ \hline
\end{longtable}
\noindent\textbf{\ding{93}\ Signal: $B^-\rightarrow \pi^+ \pi^- \pi^- D^0( \pi^+ K^- )$}\\[0.5em]
\begin{longtable}{|P{4.0cm}|P{6.5cm}|P{2.5cm}|}
\hline
\rule{0pt}{1.1em}\textbf{Background index} & \textbf{Background} & $r$ \\
\hline
\endfirsthead
\hline
\rule{0pt}{1.1em}\textbf{Background index} & \textbf{Background} & $r$ \\
\hline
\endhead
\rule{0pt}{1.1em} No. 1 & $B^-\rightarrow \pi^- \pi^- D^+( \pi^+ \pi^+ K^- )$ & 0.322 \\ \hline
\rule{0pt}{1.1em} No. 2 & $B^-\rightarrow \pi^+ \pi^- \pi^- D^0( \pi^+ K^- \pi^0 )$ & 0.306 \\ \hline
\rule{0pt}{1.1em} No. 3 & $B^-\rightarrow \pi^- D^0( \pi^+ \pi^- \bar{K}^{*0}( \pi^+ K^- ) )$ & 0.256 \\ \hline
\rule{0pt}{1.1em} No. 4 & $B^-\rightarrow \pi^- D^0( \pi^+ \pi^+ K^- \pi^- )$ & 0.218 \\ \hline
\rule{0pt}{1.1em} No. 5 & $B^-\rightarrow \pi^- \pi^- \pi^0 D^{*+}( \pi^+ D^0( \pi^+ K^- ) )$ & 0.139 \\ \hline
\rule{0pt}{1.1em} No. 6 & $B^-\rightarrow \pi^- \pi^- D^{*+}( \pi^+ D^0( \pi^+ K^- ) )$ & 0.126 \\ \hline
\rule{0pt}{1.1em} No. 7 & $\bar{B}^0\rightarrow \pi^+ \pi^- \pi^- D^+( \pi^+ \pi^+ K^- )$ & 0.069 \\ \hline
\rule{0pt}{1.1em} No. 8 & $B^-\rightarrow \pi^- \pi^- \pi^0 D^+( \pi^+ \pi^+ K^- )$ & 0.069 \\ \hline
\end{longtable}
\noindent\textbf{\ding{93}\ Signal: $\bar{B}^0\rightarrow \mu^- \bar{\nu}_{\mu} D^{*+}( \pi^+ D^0( \pi^+ K^- ) )$}\\[0.5em]
\begin{longtable}{|P{4.0cm}|P{6.5cm}|P{2.5cm}|}
\hline
\rule{0pt}{1.1em}\textbf{Background index} & \textbf{Background} & $r$ \\
\hline
\endfirsthead
\hline
\rule{0pt}{1.1em}\textbf{Background index} & \textbf{Background} & $r$ \\
\hline
\endhead
\rule{0pt}{1.1em} No. 1 & $\bar{B}^0\rightarrow \mu^- \bar{\nu}_{\mu} D^+( \pi^+ \pi^+ K^- )$ & 0.95 \\ \hline
\rule{0pt}{1.1em} No. 2 & $\bar{B}^0\rightarrow \pi^+ \mu^- \bar{\nu}_{\mu} D^0( \pi^+ K^- )$ & 0.121 \\ \hline
\end{longtable}
\noindent\textbf{\ding{93}\ Signal: $B^-\rightarrow e^- \bar{\nu}_e D^{*0}( \pi^0 D^0( \pi^+ K^- ) )$}\\[0.5em]
\begin{longtable}{|P{4.0cm}|P{6.5cm}|P{2.5cm}|}
\hline
\rule{0pt}{1.1em}\textbf{Background index} & \textbf{Background} & $r$ \\
\hline
\endfirsthead
\hline
\rule{0pt}{1.1em}\textbf{Background index} & \textbf{Background} & $r$ \\
\hline
\endhead
\rule{0pt}{1.1em} No. 1 & $B^-\rightarrow e^- \bar{\nu}_e D^0( \pi^+ K^- \pi^0 )$ & 1.195 \\ \hline
\end{longtable}
\noindent\textbf{\ding{93}\ Signal: $B^+\rightarrow \pi^+ \pi^+ D^-( K^+ \pi^- \pi^- )$}\\[0.5em]
\begin{longtable}{|P{4.0cm}|P{6.5cm}|P{2.5cm}|}
\hline
\rule{0pt}{1.1em}\textbf{Background index} & \textbf{Background} & $r$ \\
\hline
\endfirsthead
\hline
\rule{0pt}{1.1em}\textbf{Background index} & \textbf{Background} & $r$ \\
\hline
\endhead
\rule{0pt}{1.1em} No. 1 & $B^+\rightarrow \pi^+ \pi^+ \pi^- \bar{D}^0( K^+ \pi^- )$ & 1.289 \\ \hline
\rule{0pt}{1.1em} No. 2 & $B^+\rightarrow \pi^+ \pi^+ \pi^- \bar{D}^0( K^+ \pi^- \pi^0 )$ & 0.663 \\ \hline
\rule{0pt}{1.1em} No. 3 & $B^+\rightarrow \pi^+ \bar{D}^0( \pi^+ \pi^- K^{*0}( K^+ \pi^- ) )$ & 0.57 \\ \hline
\rule{0pt}{1.1em} No. 4 & $B^+\rightarrow \pi^+ \bar{D}^0( K^+ \pi^+ \pi^- \pi^- )$ & 0.495 \\ \hline
\rule{0pt}{1.1em} No. 5 & $B^+\rightarrow \pi^+ \pi^+ \pi^0 D^{*-}( \pi^- \bar{D}^0( K^+ \pi^- ) )$ & 0.331 \\ \hline
\rule{0pt}{1.1em} No. 6 & $B^+\rightarrow \pi^+ \pi^+ D^{*-}( \pi^- \bar{D}^0( K^+ \pi^- ) )$ & 0.303 \\ \hline
\rule{0pt}{1.1em} No. 7 & $B^0\rightarrow \pi^+ \pi^+ \pi^- D^-( K^+ \pi^- \pi^- )$ & 0.171 \\ \hline
\rule{0pt}{1.1em} No. 8 & $B^+\rightarrow \pi^+ \pi^+ \pi^0 D^-( K^+ \pi^- \pi^- )$ & 0.171 \\ \hline
\rule{0pt}{1.1em} No. 9 & $B^+\rightarrow \pi^+ \pi^+ D^{*-}( \pi^- \bar{D}^0( K^+ \pi^- \pi^0 ) )$ & 0.119 \\ \hline
\rule{0pt}{1.1em} No. 10 & $B^+\rightarrow \pi^+ \pi^+ \pi^- K^*_0(1430)^0( K^+ \pi^- )$ & 0.064 \\ \hline
\rule{0pt}{1.1em} No. 11 & $B^+\rightarrow \pi^+ \pi^+ \pi^- K^{*0}( K^+ \pi^- )$ & 0.064 \\ \hline
\rule{0pt}{1.1em} No. 12 & $B^+\rightarrow \pi^+ \bar{D}^0( \pi^+ \pi^- \pi^0 K^{*0}( K^+ \pi^- ) )$ & 0.059 \\ \hline
\end{longtable}
\noindent\textbf{\ding{93}\ Signal: $B^-\rightarrow \mu^- \bar{\nu}_{\mu} D^{*0}( \pi^0 D^0( K^+ K^- ) )$}\\[0.5em]
\begin{longtable}{|P{4.0cm}|P{6.5cm}|P{2.5cm}|}
\hline
\rule{0pt}{1.1em}\textbf{Background index} & \textbf{Background} & $r$ \\
\hline
\endfirsthead
\hline
\rule{0pt}{1.1em}\textbf{Background index} & \textbf{Background} & $r$ \\
\hline
\endhead
\rule{0pt}{1.1em} No. 1 & $B^-\rightarrow \mu^- \bar{\nu}_{\mu} D^0( \pi^+ K^- \pi^0 )$ & 0.205 \\ \hline
\rule{0pt}{1.1em} No. 2 & $B^-\rightarrow \mu^- \bar{\nu}_{\mu} D^0( K^+ K^- \pi^0 )$ & 0.1 \\ \hline
\rule{0pt}{1.1em} No. 3 & $B^-\rightarrow \mu^- \bar{\nu}_{\mu} D^{*0}( \pi^0 D^0( \pi^+ K^- ) )$ & 0.094 \\ \hline
\rule{0pt}{1.1em} No. 4 & $B^-\rightarrow \mu^- \bar{\nu}_{\mu} D^0( K^- K^{*+}( K^+ \pi^0 ) )$ & 0.072 \\ \hline
\end{longtable}
\noindent\textbf{\ding{93}\ Signal: $B^+\rightarrow e^+ \nu_e \bar{D}^{*0}( \pi^0 \bar{D}^0( K^+ K^- ) )$}\\[0.5em]
\begin{longtable}{|P{4.0cm}|P{6.5cm}|P{2.5cm}|}
\hline
\rule{0pt}{1.1em}\textbf{Background index} & \textbf{Background} & $r$ \\
\hline
\endfirsthead
\hline
\rule{0pt}{1.1em}\textbf{Background index} & \textbf{Background} & $r$ \\
\hline
\endhead
\rule{0pt}{1.1em} No. 1 & $B^+\rightarrow e^+ \nu_e \bar{D}^0( K^+ \pi^- \pi^0 )$ & 0.205 \\ \hline
\rule{0pt}{1.1em} No. 2 & $B^+\rightarrow e^+ \nu_e \bar{D}^0( K^+ K^- \pi^0 )$ & 0.1 \\ \hline
\rule{0pt}{1.1em} No. 3 & $B^+\rightarrow e^+ \nu_e \bar{D}^{*0}( \pi^0 \bar{D}^0( K^+ \pi^- ) )$ & 0.094 \\ \hline
\rule{0pt}{1.1em} No. 4 & $B^+\rightarrow e^+ \nu_e \bar{D}^0( K^+ K^{*-}( K^- \pi^0 ) )$ & 0.072 \\ \hline
\rule{0pt}{1.1em} No. 5 & $B^+\rightarrow e^+ \pi^0 \nu_e \bar{D}^0( K^+ K^- )$ & 0.055 \\ \hline
\end{longtable}
\noindent\textbf{\ding{93}\ Signal: $B^-\rightarrow \mu^- \bar{\nu}_{\mu} D^0( \pi^+ K^- \pi^0 )$}\\[0.5em]
\begin{longtable}{|P{4.0cm}|P{6.5cm}|P{2.5cm}|}
\hline
\rule{0pt}{1.1em}\textbf{Background index} & \textbf{Background} & $r$ \\
\hline
\endfirsthead
\hline
\rule{0pt}{1.1em}\textbf{Background index} & \textbf{Background} & $r$ \\
\hline
\endhead
\rule{0pt}{1.1em} No. 1 & $B^-\rightarrow \mu^- \bar{\nu}_{\mu} D^{*0}( \pi^0 D^0( \pi^+ K^- ) )$ & 0.361 \\ \hline
\end{longtable}
\noindent\textbf{\ding{93}\ Signal: $B^+\rightarrow e^+ \nu_e \bar{D}^0( K^+ \pi^- \pi^0 )$}\\[0.5em]
\begin{longtable}{|P{4.0cm}|P{6.5cm}|P{2.5cm}|}
\hline
\rule{0pt}{1.1em}\textbf{Background index} & \textbf{Background} & $r$ \\
\hline
\endfirsthead
\hline
\rule{0pt}{1.1em}\textbf{Background index} & \textbf{Background} & $r$ \\
\hline
\endhead
\rule{0pt}{1.1em} No. 1 & $B^+\rightarrow e^+ \nu_e \bar{D}^{*0}( \pi^0 \bar{D}^0( K^+ \pi^- ) )$ & 0.361 \\ \hline
\end{longtable}
\noindent\textbf{\ding{93}\ Signal: $B^+\rightarrow \pi^+ \bar{D}^{*0}( \pi^0 \bar{D}^0( K^+ \pi^- \pi^0 ) )$}\\[0.5em]
\begin{longtable}{|P{4.0cm}|P{6.5cm}|P{2.5cm}|}
\hline
\rule{0pt}{1.1em}\textbf{Background index} & \textbf{Background} & $r$ \\
\hline
\endfirsthead
\hline
\rule{0pt}{1.1em}\textbf{Background index} & \textbf{Background} & $r$ \\
\hline
\endhead
\rule{0pt}{1.1em} No. 1 & $B^0\rightarrow \pi^+ \pi^0 D^{*-}( \pi^- \bar{D}^0( K^+ \pi^- \pi^0 ) )$ & 0.159 \\ \hline
\rule{0pt}{1.1em} No. 2 & $B^+\rightarrow \pi^+ \pi^0 \bar{D}^0( K^+ \pi^- \pi^0 )$ & 0.145 \\ \hline
\rule{0pt}{1.1em} No. 3 & $B^+\rightarrow \pi^+ \bar{D}^0( \pi^0 \pi^0 K^{*0}( K^+ \pi^- ) )$ & 0.074 \\ \hline
\rule{0pt}{1.1em} No. 4 & $B^0\rightarrow \pi^+ \pi^0 D^{*-}( \pi^0 D^-( K^+ \pi^- \pi^- ) )$ & 0.052 \\ \hline
\end{longtable}
\noindent\textbf{\ding{93}\ Signal: $B^-\rightarrow e^- \bar{\nu}_e D^{*0}( \pi^0 D^0( K^+ K^- ) )$}\\[0.5em]
\begin{longtable}{|P{4.0cm}|P{6.5cm}|P{2.5cm}|}
\hline
\rule{0pt}{1.1em}\textbf{Background index} & \textbf{Background} & $r$ \\
\hline
\endfirsthead
\hline
\rule{0pt}{1.1em}\textbf{Background index} & \textbf{Background} & $r$ \\
\hline
\endhead
\rule{0pt}{1.1em} No. 1 & $B^-\rightarrow e^- \bar{\nu}_e D^0( \pi^+ K^- \pi^0 )$ & 0.205 \\ \hline
\rule{0pt}{1.1em} No. 2 & $B^-\rightarrow e^- \bar{\nu}_e D^0( K^+ K^- \pi^0 )$ & 0.1 \\ \hline
\rule{0pt}{1.1em} No. 3 & $B^-\rightarrow e^- \bar{\nu}_e D^{*0}( \pi^0 D^0( \pi^+ K^- ) )$ & 0.094 \\ \hline
\rule{0pt}{1.1em} No. 4 & $B^-\rightarrow e^- \bar{\nu}_e D^0( K^- K^{*+}( K^+ \pi^0 ) )$ & 0.072 \\ \hline
\end{longtable}
\noindent\textbf{\large • Relevant backgrounds for generalisation signals:}\\[1em]
\noindent\textbf{\ding{93}\ Signal: $B^+\rightarrow \pi^+ \pi^+ \pi^- \bar{D}^0( K^+ \pi^- )$}\\[0.5em]
\begin{longtable}{|P{4.0cm}|P{6.5cm}|P{2.5cm}|}
\hline
\rule{0pt}{1.1em}\textbf{Background index} & \textbf{Background} & $r$ \\
\hline
\endfirsthead
\hline
\rule{0pt}{1.1em}\textbf{Background index} & \textbf{Background} & $r$ \\
\hline
\endhead
\rule{0pt}{1.1em} No. 1 & $B^+\rightarrow \pi^+ \pi^+ D^-( K^+ \pi^- \pi^- )$ & 0.322 \\ \hline
\rule{0pt}{1.1em} No. 2 & $B^+\rightarrow \pi^+ \pi^+ \pi^- \bar{D}^0( K^+ \pi^- \pi^0 )$ & 0.306 \\ \hline
\rule{0pt}{1.1em} No. 3 & $B^+\rightarrow \pi^+ \bar{D}^0( \pi^+ \pi^- K^{*0}( K^+ \pi^- ) )$ & 0.256 \\ \hline
\rule{0pt}{1.1em} No. 4 & $B^+\rightarrow \pi^+ \bar{D}^0( K^+ \pi^+ \pi^- \pi^- )$ & 0.218 \\ \hline
\rule{0pt}{1.1em} No. 5 & $B^+\rightarrow \pi^+ \pi^+ \pi^0 D^{*-}( \pi^- \bar{D}^0( K^+ \pi^- ) )$ & 0.139 \\ \hline
\rule{0pt}{1.1em} No. 6 & $B^+\rightarrow \pi^+ \pi^+ D^{*-}( \pi^- \bar{D}^0( K^+ \pi^- ) )$ & 0.126 \\ \hline
\rule{0pt}{1.1em} No. 7 & $B^0\rightarrow \pi^+ \pi^+ \pi^- D^-( K^+ \pi^- \pi^- )$ & 0.069 \\ \hline
\rule{0pt}{1.1em} No. 8 & $B^+\rightarrow \pi^+ \pi^+ \pi^0 D^-( K^+ \pi^- \pi^- )$ & 0.069 \\ \hline
\end{longtable}
\noindent\textbf{\ding{93}\ Signal: $B^0\rightarrow \pi^+ \pi^0 D^{*-}( \pi^- \bar{D}^0( K^+ \pi^- ) )$}\\[0.5em]
\begin{longtable}{|P{4.0cm}|P{6.5cm}|P{2.5cm}|}
\hline
\rule{0pt}{1.1em}\textbf{Background index} & \textbf{Background} & $r$ \\
\hline
\endfirsthead
\hline
\rule{0pt}{1.1em}\textbf{Background index} & \textbf{Background} & $r$ \\
\hline
\endhead
\rule{0pt}{1.1em} No. 1 & $B^0\rightarrow \pi^+ D^{*-}( \pi^- \bar{D}^0( K^+ \pi^- \pi^0 ) )$ & 0.79 \\ \hline
\rule{0pt}{1.1em} No. 2 & $B^0\rightarrow \pi^+ \pi^- \bar{D}^0( K^+ \pi^- \pi^0 )$ & 0.432 \\ \hline
\rule{0pt}{1.1em} No. 3 & $B^+\rightarrow \pi^+ \pi^+ \pi^- \bar{D}^0( K^+ \pi^- \pi^0 )$ & 0.363 \\ \hline
\rule{0pt}{1.1em} No. 4 & $B^0\rightarrow \pi^+ D^-( \pi^- \pi^0 K^{*0}( K^+ \pi^- ) )$ & 0.329 \\ \hline
\rule{0pt}{1.1em} No. 5 & $B^0\rightarrow \pi^+ D^{*-}( \pi^0 D^-( K^+ \pi^- \pi^- ) )$ & 0.314 \\ \hline
\rule{0pt}{1.1em} No. 6 & $B^0\rightarrow \pi^+ \pi^0 D^{*-}( \pi^- \bar{D}^0( K^+ \pi^- \pi^0 ) )$ & 0.306 \\ \hline
\rule{0pt}{1.1em} No. 7 & $B^0\rightarrow \pi^+ \pi^0 D^-( K^+ \pi^- \pi^- )$ & 0.197 \\ \hline
\rule{0pt}{1.1em} No. 8 & $B^+\rightarrow \pi^+ \pi^+ \pi^0 D^{*-}( \pi^- \bar{D}^0( K^+ \pi^- ) )$ & 0.168 \\ \hline
\rule{0pt}{1.1em} No. 9 & $B^0\rightarrow \pi^+ \pi^- \pi^0 \bar{D}^0( K^+ \pi^- )$ & 0.166 \\ \hline
\rule{0pt}{1.1em} No. 10 & $B^0\rightarrow \pi^+ \pi^0 D^{*-}( \pi^0 D^-( K^+ \pi^- \pi^- ) )$ & 0.104 \\ \hline
\rule{0pt}{1.1em} No. 11 & $B^0\rightarrow \pi^+ D^-( K^+ \pi^- \pi^- \pi^0 )$ & 0.1 \\ \hline
\rule{0pt}{1.1em} No. 12 & $B^+\rightarrow \pi^+ \pi^+ \pi^0 D^-( K^+ \pi^- \pi^- )$ & 0.083 \\ \hline
\rule{0pt}{1.1em} No. 13 & $B^0\rightarrow \pi^+ \pi^- \bar{D}^{*0}( \pi^0 \bar{D}^0( K^+ \pi^- ) )$ & 0.067 \\ \hline
\rule{0pt}{1.1em} No. 14 & $B^0\rightarrow \pi^+ \pi^- \pi^0 \bar{D}^0( K^+ \pi^- \pi^0 )$ & 0.062 \\ \hline
\rule{0pt}{1.1em} No. 15 & $B^+\rightarrow \pi^+ \pi^+ D^{*-}( \pi^- \bar{D}^0( K^+ \pi^- \pi^0 ) )$ & 0.057 \\ \hline
\end{longtable}
\noindent\textbf{\ding{93}\ Signal: $B^-\rightarrow \mu^- \bar{\nu}_{\mu} D^{*0}( \pi^0 D^0( \pi^+ K^- ) )$}\\[0.5em]
\begin{longtable}{|P{4.0cm}|P{6.5cm}|P{2.5cm}|}
\hline
\rule{0pt}{1.1em}\textbf{Background index} & \textbf{Background} & $r$ \\
\hline
\endfirsthead
\hline
\rule{0pt}{1.1em}\textbf{Background index} & \textbf{Background} & $r$ \\
\hline
\endhead
\rule{0pt}{1.1em} No. 1 & $B^-\rightarrow \mu^- \bar{\nu}_{\mu} D^0( \pi^+ K^- \pi^0 )$ & 1.195 \\ \hline
\end{longtable}
\noindent\textbf{\ding{93}\ Signal: $B^0\rightarrow e^+ \nu_e D^-( K^+ \pi^- \pi^- )$}\\[0.5em]
\begin{longtable}{|P{4.0cm}|P{6.5cm}|P{2.5cm}|}
\hline
\rule{0pt}{1.1em}\textbf{Background index} & \textbf{Background} & $r$ \\
\hline
\endfirsthead
\hline
\rule{0pt}{1.1em}\textbf{Background index} & \textbf{Background} & $r$ \\
\hline
\endhead
\rule{0pt}{1.1em} No. 1 & $B^0\rightarrow e^+ \nu_e D^{*-}( \pi^- \bar{D}^0( K^+ \pi^- ) )$ & 0.489 \\ \hline
\rule{0pt}{1.1em} No. 2 & $B^0\rightarrow e^+ \pi^- \nu_e \bar{D}^0( K^+ \pi^- )$ & 0.078 \\ \hline
\end{longtable}
\noindent\textbf{\ding{93}\ Signal: $\bar{B}^0\rightarrow \mu^- \bar{\nu}_{\mu} D^+( \pi^+ \pi^+ K^- )$}\\[0.5em]
\begin{longtable}{|P{4.0cm}|P{6.5cm}|P{2.5cm}|}
\hline
\rule{0pt}{1.1em}\textbf{Background index} & \textbf{Background} & $r$ \\
\hline
\endfirsthead
\hline
\rule{0pt}{1.1em}\textbf{Background index} & \textbf{Background} & $r$ \\
\hline
\endhead
\rule{0pt}{1.1em} No. 1 & $\bar{B}^0\rightarrow \mu^- \bar{\nu}_{\mu} D^{*+}( \pi^+ D^0( \pi^+ K^- ) )$ & 0.489 \\ \hline
\rule{0pt}{1.1em} No. 2 & $\bar{B}^0\rightarrow \pi^+ \mu^- \bar{\nu}_{\mu} D^0( \pi^+ K^- )$ & 0.078 \\ \hline
\end{longtable}
\noindent\textbf{\ding{93}\ Signal: $B^+\rightarrow \mu^+ \nu_{\mu} \bar{D}^{*0}( \pi^0 \bar{D}^0( K^+ K^- ) )$}\\[0.5em]
\begin{longtable}{|P{4.0cm}|P{6.5cm}|P{2.5cm}|}
\hline
\rule{0pt}{1.1em}\textbf{Background index} & \textbf{Background} & $r$ \\
\hline
\endfirsthead
\hline
\rule{0pt}{1.1em}\textbf{Background index} & \textbf{Background} & $r$ \\
\hline
\endhead
\rule{0pt}{1.1em} No. 1 & $B^+\rightarrow \mu^+ \nu_{\mu} \bar{D}^0( K^+ \pi^- \pi^0 )$ & 0.205 \\ \hline
\rule{0pt}{1.1em} No. 2 & $B^+\rightarrow \mu^+ \nu_{\mu} \bar{D}^0( K^+ K^- \pi^0 )$ & 0.1 \\ \hline
\rule{0pt}{1.1em} No. 3 & $B^+\rightarrow \mu^+ \nu_{\mu} \bar{D}^{*0}( \pi^0 \bar{D}^0( K^+ \pi^- ) )$ & 0.094 \\ \hline
\rule{0pt}{1.1em} No. 4 & $B^+\rightarrow \mu^+ \nu_{\mu} \bar{D}^0( K^+ K^{*-}( K^- \pi^0 ) )$ & 0.072 \\ \hline
\rule{0pt}{1.1em} No. 5 & $B^+\rightarrow \mu^+ \pi^0 \nu_{\mu} \bar{D}^0( K^+ K^- )$ & 0.055 \\ \hline
\end{longtable}
\noindent\textbf{\ding{93}\ Signal: $B^0\rightarrow e^+ \pi^- \nu_e \bar{D}^0( K^+ \pi^- )$}\\[0.5em]
\begin{longtable}{|P{4.0cm}|P{6.5cm}|P{2.5cm}|}
\hline
\rule{0pt}{1.1em}\textbf{Background index} & \textbf{Background} & $r$ \\
\hline
\endfirsthead
\hline
\rule{0pt}{1.1em}\textbf{Background index} & \textbf{Background} & $r$ \\
\hline
\endhead
\rule{0pt}{1.1em} No. 1 & $B^0\rightarrow e^+ \nu_e D^-( K^+ \pi^- \pi^- )$ & 2.588 \\ \hline
\rule{0pt}{1.1em} No. 2 & $B^0\rightarrow e^+ \nu_e D^{*-}( \pi^- \bar{D}^0( K^+ \pi^- ) )$ & 2.17 \\ \hline
\rule{0pt}{1.1em} No. 3 & $B^0\rightarrow \mu^+ \nu_{\mu} D^-( K^+ \pi^- \pi^- )$ & 0.116 \\ \hline
\rule{0pt}{1.1em} No. 4 & $B^0\rightarrow \mu^+ \nu_{\mu} D^{*-}( \pi^- \bar{D}^0( K^+ \pi^- ) )$ & 0.075 \\ \hline
\end{longtable}
\noindent\textbf{\ding{93}\ Signal: $B^0\rightarrow \mu^+ \nu_{\mu} D^{*-}( \pi^- \bar{D}^0( K^+ K^- ) )$}\\[0.5em]
\begin{longtable}{|P{4.0cm}|P{6.5cm}|P{2.5cm}|}
\hline
\rule{0pt}{1.1em}\textbf{Background index} & \textbf{Background} & $r$ \\
\hline
\endfirsthead
\hline
\rule{0pt}{1.1em}\textbf{Background index} & \textbf{Background} & $r$ \\
\hline
\endhead
\rule{0pt}{1.1em} No. 1 & $B^0\rightarrow \mu^+ \nu_{\mu} D^-( K^+ K^- \pi^- )$ & 0.386 \\ \hline
\rule{0pt}{1.1em} No. 2 & $B^0\rightarrow \mu^+ \nu_{\mu} D^-( K^- K^{*0}( K^+ \pi^- ) )$ & 0.216 \\ \hline
\rule{0pt}{1.1em} No. 3 & $B^0\rightarrow \mu^+ \nu_{\mu} D^-( K^+ \pi^- \pi^- )$ & 0.145 \\ \hline
\rule{0pt}{1.1em} No. 4 & $B^0\rightarrow \mu^+ \pi^- \nu_{\mu} \bar{D}^0( K^+ K^- )$ & 0.121 \\ \hline
\rule{0pt}{1.1em} No. 5 & $B^0\rightarrow \mu^+ \nu_{\mu} D^{*-}( \pi^- \bar{D}^0( K^+ \pi^- ) )$ & 0.094 \\ \hline
\end{longtable}

\newpage

\section{GA space size}\label{GA_space_size}

This appendix presents the different types of decay structures of the backgrounds for an example signal within the toy scenario introduced in Section \ref{environment}, namely $M^+\to F_1^+ F_2^+ F_3^- F_4^0 F_5^0$, which corresponds to the decay of a positively charged mother particle into two positively charged, one negatively charged, and two neutral final-state particles. Table \ref{space_size_chart} organises these structures according to the number of intermediate resonances (IRs), the number of non-reconstructed particles, the charge of the mother particle, and the charges of the IRs present in the background. The GA space size of each category is also included. While we have attempted to enumerate all possible cases, it cannot be excluded that additional subcases exist. Nevertheless, the list of decay structures provided here is intended to be representative rather than exhaustive. The computation of the total GA space size for this example signal yields 333,305,600\footnote{The computed size represents the search space that the GA can explore, not the number of physically allowed background processes.}.\\

\setcounter{table}{3}

{
\LTcapwidth=\textwidth
\renewcommand\arraystretch{1.3} % only affects this table
\begin{longtable}{|P{1.2cm}|P{2cm}|P{1.2cm}|P{1.5cm}|P{5.3cm}|P{5.7cm}|}%[htbp]
    %\centering
    %\begin{tabular}{|P{2cm}|P{2cm}|P{1.5cm}|P{1.5cm}|P{5cm}|P{5.5cm}|}%{|c|c|c|}
        \hline
         No. of IRs & No. of non-reconstructed particles & Mother particle type & IR type & Background type & Size\\ \hline \hline
         \endfirsthead 
         \hline
         No. of IRs & No. of non-reconstructed particles & Mother particle type & IR type & Background type & Size\\ \hline \hline
         \endhead
         \multirow{3}{*}{0} & 0 & $M^+$ & - & $M^+\to F_1^+ F_2^+ F_3^- F_4^0 F_5^0$ & $1\cdot4^2\cdot4\cdot10^2=6{,}400$ \\ \cline{2-6}
         & \multirow{2}{*}{1} & $M^0$ & - & $M^0\to F_1^+ F_2^+ F_3^- F_4^0 F_5^0 F_6^-$ &  $2\cdot4^2\cdot4^2\cdot10^2=51{,}200$ \\
         \cline{3-6}
         & & $M^+$ & - & $M^+\to F_1^+ F_2^+ F_3^- F_4^0 F_5^0 F_6^0$ & $1\cdot4^2\cdot4\cdot10^3=64{,}000$ \\ \hline

         \multirow{39}{*}{1} & \multirow{10}{*}{0} & \multirow{10}{*}{$M^+$} & \multirow{4}{*}{$I_1^+$} & $M^+\to F_1^+ I_1^+( F_2^+ F_4^0 ) F_3^- F_5^0$ & \multirow{4}{*}{$4\cdot(1\cdot4^2\cdot4\cdot10^2\cdot4)=102{,}400$}  \\%25{,}600
         & & & & $M^+\to I_1^+( F_1^+ F_2^+ F_3^-) F_4^0 F_5^0$ & \\
         & & & & $M^+\to I_1^+( F_1^+ F_4^0 F_5^0) F_2^+ F_3^-$  & \\
         & & & & $M^+\to I_1^+( F_1^+ F_2^+ F_3^- F_4^0) F_5^0$  & \\
         \cline{4-6}
         & & & \multirow{4}{*}{$I_1^0$} & $M^+\to F_1^+ I_1^0( F_2^+ F_3^- ) F_4^0 F_5^0$ & \multirow{4}{*}{$4\cdot(1\cdot4^2\cdot4\cdot10^2\cdot10)=256{,}000$}  \\%64{,}000
         & & & & $M^+\to F_1^+ F_2^+ F_3^- I_1^0(F_4^0 F_5^0 )$  & \\
         & & & & $M^+\to I_1^0( F_1^+ F_3^- F_4^0 ) F_2^+ F_5^0$  & \\
         & & & & $M^+\to I_1^0( F_1^+ F_3^- F_4^0 F_5^0 ) F_2^+ $  & \\
         \cline{4-6}
         & & & \multirow{2}{*}{$I_1^-$} & $M^+\to F_1^+ F_2^+ I_1^-( F_3^- F_4^0 ) F_5^0$ & \multirow{2}{*}{$2\cdot(1\cdot4^2\cdot4\cdot10^2\cdot4)=51{,}200$} \\%25{,}600
         & & & & $M^+\to F_1^+ F_2^+ I_1^-( F_3^- F_4^0 F_5^0 )$  & \\
         \cline{2-6}
         & \multirow{29}{*}{1} & \multirow{15}{*}{$M^0$} & \multirow{5}{*}{$I_1^+$} & $M^0\to F_1^+ I_1^+( F_2^+ F_4^0 ) F_3^- F_5^0 F_6^-$ & \multirow{5}{*}{$5\cdot(2\cdot4^2\cdot4^2\cdot10^2\cdot4)=1{,}024{,}000$} \\%204{,}800
         & & & & $M^0\to I_1^+(F_1^+ F_2^+ F_3^-) F_4^0 F_5^0 F_6^-$  & \\
         & & & & $M^0\to I_1^+(F_1^+ F_4^0 F_5^0 ) F_2^+ F_3^- F_6^-$  & \\
         & & & & $M^0\to I_1^+(F_1^+ F_2^+ F_3^- F_4^0 ) F_5^0 F_6^-$  & \\
         & & & & $M^0\to I_1^+(F_1^+ F_2^+ F_3^- F_4^0 F_5^0 ) F_6^-$  & \\
         \cline{4-6}
         & & & \multirow{5}{*}{$I_1^0$} & $M^0\to F_1^+ I_1^0( F_2^+ F_3^- ) F_4^0 F_5^0 F_6^-$ & \multirow{5}{*}{$5\cdot(2\cdot4^2\cdot4^2\cdot10^2\cdot10)=2{,}560{,}000$}\\%512{,}000
         & & & & $M^0\to F_1^+ F_2^+ F_3^- I_1^0( F_4^0 F_5^0 ) F_6^-$  & \\
         & & & & $M^0\to I_1^0(F_1^+ F_3^- F_4^0 ) F_2^+ F_5^0 F_6^-$  & \\
         & & & & $M^0\to I_1^0(F_1^+ F_3^- F_4^0 F_5^0 ) F_2^+  F_6^-$  & \\
         & & & & $M^0\to I_1^0(F_1^+ F_2^+ F_3^- F_4^0 F_6^- ) F_5^0 $ & \\
         \cline{4-6}
         & & & \multirow{5}{*}{$I_1^-$} & $M^0\to F_1^+ F_2^+ I_1^-( F_3^- F_4^0 ) F_5^0 F_6^-$ & \multirow{5}{*}{$5\cdot(2\cdot4^2\cdot4^2\cdot10^2\cdot4)=1{,}024{,}000$} \\%204{,}800
         & & & & $M^0\to F_1^+ F_2^+ I_1^-( F_3^- F_4^0 F_5^0 ) F_6^-$ & \\
         & & & & $M^0\to F_1^+ I_1^-( F_2^+ F_3^- F_6^- ) F_4^0 F_5^0 $ & \\
         & & & & $M^0\to F_1^+ I_1^-( F_2^+ F_3^- F_6^- F_4^0 ) F_5^0 $ & \\
         & & & & $M^0\to F_1^+ I_1^-( F_2^+ F_3^- F_6^- F_4^0 F_5^0 ) $ & \\
         \cline{3-6}
         & & \multirow{5}{*}{$M^+$} & \multirow{5}{*}{$I_1^+$} & $M^+\to I_1^+( F_1^+ F_4^0 ) F_2^+ F_3^- F_5^0 F_6^0$ & \multirow{5}{*}{$5\cdot(1\cdot4^2\cdot4\cdot10^3\cdot4)=1{,}280{,}000$} \\%Adjusted: \multirow{14}{*}{$M^+$} %256{,}000
         & & & & $M^+\to I_1^+( F_1^+ F_2^+ F_3^- ) F_4^0 F_5^0 F_6^0$ & \\
         & & & & $M^+\to I_1^+( F_1^+ F_4^0 F_5^0 ) F_2^+ F_3^- F_6^0$ & \\
         & & & & $M^+\to I_1^+( F_1^+ F_2^+ F_3^- F_4^0 ) F_5^0 F_6^0$ & \\
         & & & & $M^+\to I_1^+( F_1^+ F_2^+ F_3^- F_4^0 F_5^0 ) F_6^0$ & \\
         \cline{4-6}
         & & & \multirow{6}{*}{$I_1^0$} & $M^+\to I_1^0( F_1^+ F_3^- ) F_2^+ F_4^0 F_5^0 F_6^0$ & \multirow{6}{*}{$6\cdot(1\cdot4^2\cdot4\cdot10^3\cdot10)=3{,}840{,}000$} \\%640{,}000
         & & & & $M^+\to F_1^+ F_2^+ F_3^- I_1^0( F_4^0 F_5^0 ) F_6^0$ & \\
         & & & & $M^+\to F_1^+ I_1^0( F_2^+ F_3^- F_4^0 ) F_5^0 F_6^0$ & \\
         & & & & $M^+\to F_1^+ F_2^+ F_3^- I_1^0( F_4^0 F_5^0 F_6^0 )$ & \\
         & & & & $M^+\to F_1^+ I_1^0( F_2^+ F_3^- F_4^0 F_5^0 ) F_6^0$ & \\
         & & & & $M^+\to F_1^+ I_1^0( F_2^+ F_3^- F_4^0 F_5^0 F_6^0 )$ & \\
         \cline{4-6}
         & & & \multirow{3}{*}{$I_1^-$} & $M^+\to  F_1^+ F_2^+ I_1^-( F_3^- F_4^0 ) F_5^0 F_6^0$ & \multirow{3}{*}{$3\cdot(1\cdot4^2\cdot4\cdot10^3\cdot4)=768{,}000$} \\%256{,}000
         & & & & $M^+\to  F_1^+ F_2^+ I_1^-( F_3^- F_4^0 F_5^0 ) F_6^0$ & \\
         & & & & $M^+\to  F_1^+ F_2^+ I_1^-( F_3^- F_4^0 F_5^0 F_6^0 )$ & \\
         \cline{1-6}
         \multirow{60}{*}{2} & \multirow{8}{*}{0} & \multirow{8}{*}{$M^+$} & \multirow{2}{*}{$I_1^+$ \& $I_2^+$} & $M^+\to  I_1^+( F_1^+ F_4^0 ) I_2^+( F_2^+ F_5^0 ) F_3^-$ & \multirow{2}{*}{$2\cdot(1\cdot4^2\cdot4\cdot10^2\cdot4^2)=204{,}800$} \\%Adjusted: \multirow{121}{*}{2} %102{,}400
        & & & & $M^+\to  I_1^+( F_1^+ F_3^- I_2^+( F_2^+ F_4^0 ) ) F_5^0 $ & \\
        \cline{4-6}
        & & & \multirow{4}{*}{$I_1^+$ \& $I_2^0$} & $M^+\to  I_1^+( F_1^+ F_4^0 ) I_2^0( F_2^+ F_3^- ) F_5^0 $ & \multirow{4}{*}{$4\cdot(1\cdot4^2\cdot4\cdot10^2\cdot4\cdot10)=1{,}024{,}000$} \\%256{,}000
         & & & & $M^+\to  I_1^+( F_1^+ F_4^0 ) I_2^0( F_2^+ F_3^- F_5^0 )$ & \\
         & & & & $M^+\to  I_1^+( F_1^+ F_4^0 F_5^0 ) I_2^0( F_2^+ F_3^- )$ & \\
         & & & & $M^+\to  I_1^+( F_1^+ F_2^+ F_3^- ) I_2^0( F_4^0 F_5^0 )$ & \\
         \cline{4-6}
         & & & $I_1^+$ \& $I_2^-$ & $M^+\to  I_1^+( F_1^+ F_4^0 ) I_2^-( F_3^- F_5^0 ) F_2^+ $ & $1\cdot4^2\cdot4\cdot10^2\cdot4^2=102{,}400$ \\
         \cline{4-6}
         & & & $I_1^0$ \& $I_2^0$ & $M^+\to  I_1^0( F_1^+ F_3^- ) I_2^0( F_4^0 F_5^0 ) F_2^+ $ & $1\cdot4^2\cdot4\cdot10^2\cdot10^2=640{,}000$\\
         \cline{2-6}
         & \multirow{50}{*}{1} & \multirow{48}{*}{$M^0$} & \multirow{4}{*}{$I_1^+$ \& $I_2^+$} & $M^0\to I_1^+( F_1^+ F_4^0 ) I_2^+( F_2^+ F_5^0 ) F_3^- F_6^-$ & \multirow{4}{*}{$4\cdot(2\cdot4^2\cdot4^2\cdot10^2\cdot4^2)=3{,}276{,}800 $} \\%Adjusted: \multirow{113}{*}{1} %%%819{,}200
         & & & & $M^0\to I_1^+( F_1^+ F_3^- I_2^+( F_2^+ F_4^0 ) ) F_5^0 F_6^-$ & \\
         & & & & $M^0\to I_1^+( F_1^+ F_3^- I_2^+( F_2^+ F_4^0 F_5^0 ) ) F_6^-$ & \\
         & & & & $M^0\to I_1^+( F_1^+ F_3^- F_5^0 I_2^+( F_2^+ F_4^0 ) ) F_6^-$ & \\
         \cline{4-6}
         & & & \multirow{15}{*}{$I_1^+$ \& $I_2^0$} & $M^0\to I_1^+( F_1^+ F_4^0 ) I_2^0( F_2^+ F_3^- ) F_5^0 F_6^-$ & \multirow{15}{*}{$15\cdot(2\cdot4^2\cdot4^2\cdot10^2\cdot4\cdot10)=30{,}720{,}000$} \\%2{,}048{,}000
         & & & & $M^0\to I_1^+( F_1^+ F_4^0 F_5^0 ) I_2^0( F_2^+ F_3^- )  F_6^-$ & \\
         & & & & $M^0\to I_1^+( F_1^+ F_4^0 ) I_2^0( F_2^+ F_3^- F_5^0 )  F_6^-$ & \\
         & & & & $M^0\to I_1^+( F_1^+ F_2^+ F_3^- ) I_2^0( F_4^0 F_5^0 ) F_6^-$ & \\
         & & & & $M^0\to I_1^+( F_1^+ I_2^0( F_2^+ F_3^- ) ) F_4^0 F_5^0 F_6^-$ & \\
         & & & & $M^0\to I_1^+( F_1^+ I_2^0( F_2^+ F_3^- ) F_4^0 ) F_5^0 F_6^-$ & \\
         & & & & $M^0\to I_1^+( F_1^+ I_2^0( F_2^+ F_3^- F_4^0 ) ) F_5^0 F_6^-$ & \\
         & & & & $M^0\to I_1^+( F_1^+ I_2^0( F_2^+ F_3^- ) F_4^0 F_5^0 ) F_6^-$ & \\
         & & & & $M^0\to I_1^+( F_1^+ I_2^0( F_2^+ F_3^- F_4^0 ) F_5^0 ) F_6^-$ & \\
         & & & & $M^0\to I_1^+( F_1^+ I_2^0( F_2^+ F_3^- F_4^0 F_5^0 ) ) F_6^-$ & \\
         & & & & $M^0\to I_2^0( I_1^+( F_1^+ F_4^0 ) F_3^- ) F_2^+ F_5^0 F_6^-$ & \\
         & & & & $M^0\to I_2^0( I_1^+( F_1^+ F_4^0 ) F_3^- F_5^0 ) F_2^+ F_6^-$ & \\
         & & & & $M^0\to I_2^0( I_1^+( F_1^+ F_4^0 F_5^0 ) F_3^- ) F_2^+ F_6^-$ & \\
         & & & & $M^0\to I_2^0( I_1^+( F_1^+ F_4^0 ) F_3^- F_2^+ F_6^- ) F_5^0 $ & \\
         & & & & $M^0\to I_2^0( I_1^+( F_1^+ F_4^0 F_2^+ F_6^- ) F_3^- ) F_5^0 $ & \\
         \cline{4-6}
         & & & \multirow{13}{*}{$I_1^+$ \& $I_2^-$} & $M^0\to I_1^+( F_1^+ F_4^0 ) F_2^+ I_2^-( F_3^- F_5^0 ) F_6^- $ & \multirow{15}{*}{$13\cdot(2\cdot4^2\cdot4^2\cdot10^2\cdot4^2)=10{,}649{,}600$} \\%819{,}200
         & & & & $M^0\to I_1^+( F_1^+ F_2^+ F_3^- ) F_4^0 I_2^-( F_5^0 F_6^- ) $ & \\
         & & & & $M^0\to I_1^+( F_1^+ F_2^+ F_3^- F_4^0 ) I_2^-( F_5^0 F_6^- ) $ & \\
         & & & & $M^0\to I_1^+( F_1^+ F_2^+ F_3^- ) I_2^-( F_4^0 F_5^0 F_6^- ) $ & \\
         & & & & $M^0\to I_1^+( F_1^+ F_4^0 ) I_2^-( F_2^+ F_3^- F_6^- ) F_5^0 $ & \\
         & & & & $M^0\to I_1^+( F_1^+ F_4^0 F_5^0 ) I_2^-( F_2^+ F_3^- F_6^- ) $ & \\
         & & & & $M^0\to I_1^+( F_1^+ F_4^0 ) I_2^-( F_2^+ F_3^- F_5^0 F_6^- ) $ & \\
         & & & & $M^0\to I_1^+( F_1^+ F_2^+ I_2^-( F_3^- F_4^0 ) ) F_5^0 F_6^- $ & \\
         & & & & $M^0\to I_1^+( F_1^+ F_2^+ I_2^-( F_3^- F_4^0 ) F_5^0 ) F_6^- $ & \\
         & & & & $M^0\to I_1^+( F_1^+ F_2^+ I_2^-( F_3^- F_4^0 F_5^0  ) ) F_6^- $ & \\
         & & & & $M^0\to F_1^+ I_2^-( F_3^- F_6^- I_1^+( F_2^+ F_4^0 ) ) F_5^0 $ & \\
         & & & & $M^0\to F_1^+ I_2^-( F_3^- F_6^- I_1^+( F_2^+ F_4^0 ) F_5^0 ) $ & \\
         & & & & $M^0\to F_1^+ I_2^-( F_3^- F_6^- I_1^+( F_2^+ F_4^0 F_5^0 ) ) $ & \\
         \cline{4-6}
         & & & \multirow{14}{*}{$I_1^0$ \& $I_2^0$} & $M^0\to I_1^0( F_1^+ F_6^- ) I_2^0( F_2^+ F_3^- ) F_4^0 F_5^0 $ & \multirow{14}{*}{$14\cdot(2\cdot4^2\cdot4^2\cdot10^2\cdot10^2)=71{,}680{,}000$} \\%5.120.000
         & & & & $M^0\to I_1^0( F_1^+ F_6^- F_4^0 ) I_2^0( F_2^+ F_3^- ) F_5^0 $ & \\
         & & & & $M^0\to I_1^0( F_1^+ F_6^- ) I_2^0( F_2^+ F_3^- F_4^0 ) F_5^0 $ & \\
         & & & & $M^0\to I_1^0( F_1^+ F_6^- F_4^0 F_5^0 ) I_2^0( F_2^+ F_3^- ) $ & \\
         & & & & $M^0\to I_1^0( F_1^+ F_6^- F_4^0 ) I_2^0( F_2^+ F_3^- F_5^0 ) $ & \\
         & & & & $M^0\to I_1^0( F_1^+ F_6^- ) I_2^0( F_2^+ F_3^- F_4^0 F_5^0 ) $ & \\
         & & & & $M^0\to I_1^0( F_1^+ F_6^- ) F_2^+ F_3^- I_2^0( F_4^0 F_5^0 ) $ & \\
         & & & & $M^0\to I_1^0( F_1^+ F_2^+ F_3^- F_6^- ) I_2^0( F_4^0 F_5^0 ) $ & \\
         & & & & $M^0\to I_1^0( I_2^0( F_1^+ F_6^- ) F_2^+ F_3^- ) F_4^0 F_5^0 $ & \\
         & & & & $M^0\to I_1^0( I_2^0( F_1^+ F_6^- ) F_2^+ F_3^- F_4^0 ) F_5^0 $ & \\
         & & & & $M^0\to I_1^0( I_2^0( F_1^+ F_6^- F_4^0 ) F_2^+ F_3^- ) F_5^0 $ & \\
         & & & & $M^0\to I_1^0( I_2^0( F_1^+ F_6^- F_4^0 ) F_5^0 ) F_2^+ F_3^- $ & \\
         & & & & $M^0\to I_1^0( I_2^0( F_1^+ F_6^- ) F_4^0 F_5^0 ) F_2^+ F_3^- $ & \\
         & & & & $M^0\to I_1^0( I_2^0( F_4^0 F_5^0 ) F_1^+ F_6^- ) F_2^+ F_3^- $ & \\
         \cline{4-6}
         & & & $I_1^0$ \& $I_2^-$ & - & 30,720,000 (symmetry) \\
         \cline{4-6}
         & & & $I_1^-$ \& $I_2^-$ & - & 3,276,800 (symmetry) \\
         \cline{3-6}
         & & \multirow{48}{*}{$M^+$} & \multirow{10}{*}{$I_1^+$ \& $I_2^+$} & $M^+\to I_1^+( F_1^+ F_4^0 ) I_2^+( F_2^+ F_5^0 ) F_3^- F_6^0 $ & \multirow{10}{*}{$10\cdot(1\cdot4^2\cdot4\cdot10^3\cdot4^2)=10{,}240{,}000$} \\ %Adjusted:\multirow{64}{*}{$M^+$}
         & & & & $M^+\to I_1^+( F_1^+ F_4^0 F_6^0 ) I_2^+( F_2^+ F_5^0 ) F_3^- $ & \\
         & & & & $M^+\to I_1^+( F_1^+ F_4^0 ) I_2^+( F_2^+ F_5^0 F_6^0 ) F_3^- $ & \\
         & & & & $M^+\to I_1^+( F_4^0 I_2^+( F_2^+ F_5^0 ) ) F_1^+ F_3^- F_6^0 $ & \\
         & & & & $M^+\to I_1^+( F_1^+ F_3^- F_4^0 I_2^+( F_2^+ F_5^0 ) ) F_6^0 $ & \\
         & & & & $M^+\to I_1^+( F_4^0 I_2^+(F_1^+ F_3^- F_2^+ F_5^0 ) ) F_6^0 $ & \\
         & & & & $M^+\to I_1^+( F_4^0 F_6^0 I_2^+( F_2^+ F_5^0 ) ) F_1^+ F_3^-  $ & \\
         & & & & $M^+\to I_1^+( F_4^0 I_2^+( F_2^+ F_5^0 F_6^0 ) ) F_1^+ F_3^-  $ & \\
         & & & & $M^+\to I_1^+( F_1^+ F_3^- I_2^+( F_2^+ F_4^0 ) ) F_5^0 F_6^0 $ & \\
         & & & & $M^+\to I_1^+( F_1^+ F_3^- I_2^+( F_2^+ F_4^0 F_5^0 ) ) F_6^0 $ & \\
         \cline{4-6}
         & & & \multirow{23}{*}{$I_1^+$ \& $I_2^0$} & $M^+\to I_1^+( F_1^+ F_4^0 ) I_2^0( F_2^+ F_3^-) F_5^0 F_6^0 $ & \multirow{23}{*}{$23\cdot(1\cdot4^2\cdot4\cdot10^3\cdot4\cdot10)=58{,}880{,}000$} \\
         & & & & $M^+\to I_1^+( F_1^+ F_4^0 F_5^0 ) I_2^0( F_2^+ F_3^-)  F_6^0 $ & \\
         & & & & $M^+\to I_1^+( F_1^+ F_4^0 ) I_2^0( F_2^+ F_3^- F_5^0)  F_6^0 $ & \\
         & & & & $M^+\to I_1^+( F_1^+ F_4^0 F_5^0 ) I_2^0( F_2^+ F_3^- F_6^0 ) $ & \\
         & & & & $M^+\to I_1^+( F_1^+ F_4^0 F_5^0 F_6^0 ) I_2^0( F_2^+ F_3^- ) $ & \\
         & & & & $M^+\to I_1^+( F_1^+ F_4^0 ) I_2^0( F_2^+ F_3^- F_5^0 F_6^0) $ & \\
         & & & & $M^+\to I_1^+( F_1^+ F_4^0 ) I_2^0( F_5^0 F_6^0 ) F_2^+ F_3^- $ & \\
         & & & & $M^+\to I_1^+( F_1^+ F_4^0 F_2^+ F_3^- ) I_2^0( F_5^0 F_6^0 ) $ & \\
         & & & & $M^+\to I_1^+( F_1^+ I_2^0( F_2^+ F_3^- ) ) F_4^0 F_5^0 F_6^0  $ & \\
         & & & & $M^+\to I_1^+( F_1^+ F_4^0 I_2^0( F_2^+ F_3^- ) ) F_5^0 F_6^0  $ & \\
         & & & & $M^+\to I_1^+( F_1^+ I_2^0( F_2^+ F_3^- F_4^0 ) ) F_5^0 F_6^0  $ & \\
         & & & & $M^+\to I_1^+( F_1^+ F_4^0 F_5^0 I_2^0( F_2^+ F_3^- ) ) F_6^0  $ & \\
         & & & & $M^+\to I_1^+( F_1^+ F_4^0 I_2^0( F_2^+ F_3^- F_5^0 ) ) F_6^0  $ & \\
         & & & & $M^+\to I_1^+( F_1^+ I_2^0( F_2^+ F_3^- F_4^0 F_5^0 ) ) F_6^0  $ & \\
         & & & & $M^+\to I_1^+( F_1^+ I_2^0( F_4^0 F_5^0 ) ) F_2^+ F_3^- F_6^0 $ & \\
         & & & & $M^+\to I_1^+( F_1^+ F_6^0 I_2^0( F_4^0 F_5^0 ) ) F_2^+ F_3^- $ & \\
         & & & & $M^+\to I_1^+( F_1^+ I_2^0( F_4^0 F_5^0 F_6^0 ) ) F_2^+ F_3^- $ & \\
         & & & & $M^+\to I_1^+( F_1^+ F_2^+ F_3^- I_2^0( F_4^0 F_5^0 ) ) F_6^0 $ & \\
         & & & & $M^+\to I_2^0( I_1^+( F_1^+ F_4^0 ) F_3^- ) F_2^+ F_5^0 F_6^0 $ & \\
         & & & & $M^+\to I_2^0( I_1^+( F_1^+ F_4^0 ) F_3^- F_5^0 ) F_2^+ F_6^0 $ & \\
         & & & & $M^+\to I_2^0( I_1^+( F_1^+ F_4^0 F_5^0 ) F_3^- ) F_2^+ F_6^0 $ & \\
         & & & & $M^+\to I_2^0( I_1^+( F_1^+ F_4^0 F_5^0 ) F_3^- F_6^0 ) F_2^+ $ & \\
         & & & & $M^+\to I_2^0( I_1^+( F_1^+ F_4^0 F_5^0 F_6^0 ) F_3^- ) F_2^+ $ & \\
         \cline{4-6}
         & & & \multirow{6}{*}{$I_1^+$ \& $I_2^-$} & $M^+\to I_1^+( F_1^+ F_4^0 ) I_2^-( F_3^- F_5^0 )  F_2^+ F_6^0 $ & \multirow{6}{*}{$6\cdot(1\cdot4^2\cdot4\cdot10^3\cdot4\cdot4)=6{,}144{,}000$} \\ 
         & & & & $M^+\to I_1^+( F_1^+ F_4^0 F_6^0 ) I_2^-( F_3^- F_5^0 ) F_2^+ $ & \\
         & & & & $M^+\to I_1^+( F_1^+ F_4^0 ) I_2^-( F_3^- F_5^0 F_6^0 ) F_2^+ $ & \\
         & & & & $M^+\to I_1^+( F_1^+ F_2^+ I_2^-( F_3^- F_4^0 ) ) F_5^0 F_6^0 $ & \\
         & & & & $M^+\to I_1^+( F_1^+ F_2^+ F_5^0 I_2^-( F_3^- F_4^0 ) ) F_6^0 $ & \\
         & & & & $M^+\to I_1^+( F_1^+ F_2^+ I_2^-( F_3^- F_4^0 F_5^0 ) ) F_6^0 $ & \\
         \cline{4-6}
          & & & \multirow{12}{*}{$I_1^0$ \& $I_2^0$} & $M^+\to I_1^0( F_1^+ F_3^- ) I_2^0( F_4^0 F_5^0 ) F_2^+ F_6^0 $ & \multirow{12}{*}{$12\cdot(1\cdot4^2\cdot4\cdot10^3\cdot10^2)=76{,}800{,}000$} \\ 
         & & & & $M^+\to I_1^0( F_1^+ F_3^- F_6^0 ) I_2^0( F_4^0 F_5^0 ) F_2^+ $ & \\
         & & & & $M^+\to I_1^0( F_1^+ F_3^- ) I_2^0( F_4^0 F_5^0 F_6^0 ) F_2^+ $ & \\
         & & & & $M^+\to I_1^0( I_2^0( F_1^+ F_3^- ) F_4^0 ) F_2^+ F_5^0 F_6^0 $ & \\
         & & & & $M^+\to I_1^0( I_2^0( F_1^+ F_3^- ) F_4^0 F_5^0 ) F_2^+ F_6^0 $ & \\
         & & & & $M^+\to I_1^0( I_2^0( F_1^+ F_3^- F_5^0 ) F_4^0 ) F_2^+ F_6^0 $ & \\
         & & & & $M^+\to I_1^0( I_2^0( F_1^+ F_3^- ) F_4^0 F_5^0 F_6^0 ) F_2^+ $ & \\
         & & & & $M^+\to I_1^0( I_2^0( F_1^+ F_3^- F_5^0 F_6^0 ) F_4^0 ) F_2^+ $ & \\
         & & & & $M^+\to I_1^0( I_2^0( F_4^0 F_5^0 ) F_1^+ F_3^- ) F_2^+ F_6^0 $ & \\
         & & & & $M^+\to I_1^0( I_2^0( F_4^0 F_5^0 ) F_1^+ F_3^- F_6^0 ) F_2^+ $ & \\
         & & & & $M^+\to I_1^0( I_2^0( F_4^0 F_5^0 F_6^0 ) F_1^+ F_3^- ) F_2^+ $ & \\
         & & & & $M^+\to I_1^0( I_2^0( F_4^0 F_5^0 ) F_6^0 ) F_1^+ F_3^- F_2^+ $ & \\
        \cline{4-6}
        & & & \multirow{7}{*}{$I_1^0$ \& $I_2^-$} & $M^+\to F_1^+ F_2^+ I_2^-( F_3^- F_4^0 ) I_1^0( F_5^0 F_6^0 ) $ & \multirow{7}{*}{$7\cdot(1\cdot4^2\cdot4\cdot10^3\cdot4\cdot10)=17{,}920{,}000$} \\ 
         & & & & $M^+\to I_2^-( F_3^- I_1^0( F_4^0 F_5^0 ) ) F_1^+ F_2^+ F_6^0 $ & \\
         & & & & $M^+\to I_2^-( F_3^- F_6^0 I_1^0( F_4^0 F_5^0 ) ) F_1^+ F_2^+ $ & \\
         & & & & $M^+\to I_2^-( F_3^- I_1^0( F_4^0 F_5^0 F_6^0 ) ) F_1^+ F_2^+ $ & \\
         & & & & $M^+\to I_1^0( F_1^+ F_4^0 I_2^-( F_3^- F_5^0 ) ) F_2^+ F_6^0 $ & \\
         & & & & $M^+\to I_1^0( F_1^+ F_4^0 F_6^0 I_2^-( F_3^- F_5^0 ) ) F_2^+ $ & \\
         & & & & $M^+\to I_1^0( F_1^+ F_4^0 I_2^-( F_3^- F_5^0 F_6^0 ) ) F_2^+ $ & \\
         \cline{1-6}
    %\end{tabular}
    \caption{Categories of decay structures of the backgrounds for an example signal $M^+\to F_1^+ F_2^+ F_3^- F_4^0 F_5^0$, together with the corresponding GA space sizes. The decay structures are organised by the number of IRs, the number of non-reconstructed particles, the charge of the mother particle, and the charges of the IRs present in the background.}
    \label{space_size_chart}
\end{longtable}
}

\end{appendices}

%\bibliographystyle{plain}

%\begin{thebibliography}{99}
%\end{thebibliography}


\begin{thebibliography}{38}%
\makeatletter
\providecommand \@ifxundefined [1]{%
 \@ifx{#1\undefined}
}%
\providecommand \@ifnum [1]{%
 \ifnum #1\expandafter \@firstoftwo
 \else \expandafter \@secondoftwo
 \fi
}%
\providecommand \@ifx [1]{%
 \ifx #1\expandafter \@firstoftwo
 \else \expandafter \@secondoftwo
 \fi
}%
\providecommand \natexlab [1]{#1}%
\providecommand \enquote  [1]{``#1''}%
\providecommand \bibnamefont  [1]{#1}%
\providecommand \bibfnamefont [1]{#1}%
\providecommand \citenamefont [1]{#1}%
\providecommand \href@noop [0]{\@secondoftwo}%
\providecommand \href [0]{\begingroup \@sanitize@url \@href}%
\providecommand \@href[1]{\@@startlink{#1}\@@href}%
\providecommand \@@href[1]{\endgroup#1\@@endlink}%
\providecommand \@sanitize@url [0]{\catcode `\\12\catcode `\$12\catcode `\&12\catcode `\#12\catcode `\^12\catcode `\_12\catcode `\%12\relax}%
\providecommand \@@startlink[1]{}%
\providecommand \@@endlink[0]{}%
\providecommand \url  [0]{\begingroup\@sanitize@url \@url }%
\providecommand \@url [1]{\endgroup\@href {#1}{\urlprefix }}%
\providecommand \urlprefix  [0]{URL }%
\providecommand \Eprint [0]{\href }%
\providecommand \doibase [0]{https://doi.org/}%
\providecommand \selectlanguage [0]{\@gobble}%
\providecommand \bibinfo  [0]{\@secondoftwo}%
\providecommand \bibfield  [0]{\@secondoftwo}%
\providecommand \translation [1]{[#1]}%
\providecommand \BibitemOpen [0]{}%
\providecommand \bibitemStop [0]{}%
\providecommand \bibitemNoStop [0]{.\EOS\space}%
\providecommand \EOS [0]{\spacefactor3000\relax}%
\providecommand \BibitemShut  [1]{\csname bibitem#1\endcsname}%
\let\auto@bib@innerbib\@empty
%</preamble>
\bibitem [{\citenamefont {Alves~Jr}\ and\ \citenamefont {others (LHCb~Collaboration)}(2008)}]{LHCb}%
  \BibitemOpen
  \bibfield  {author} {\bibinfo {author} {\bibfnamefont {A.~A.}\ \bibnamefont {Alves~Jr}}\ and\ \bibinfo {author} {\bibnamefont {others (LHCb~Collaboration)}},\ }\href {https://doi.org/10.1088/1748-0221/3/08/S08005} {\bibfield  {journal} {\bibinfo  {journal} {Journal of Instrumentation}\ }\textbf {\bibinfo {volume} {3}},\ \bibinfo {pages} {S08005}}\BibitemShut {NoStop}%
\bibitem [{\citenamefont {Kou}\ and\ \citenamefont {others (Belle II~Collaboration)}(2019)}]{Belle2}%
  \BibitemOpen
  \bibfield  {author} {\bibinfo {author} {\bibfnamefont {E.}~\bibnamefont {Kou}}\ and\ \bibinfo {author} {\bibnamefont {others (Belle II~Collaboration)}},\ }\href {https://doi.org/10.1093/ptep/ptz106} {\bibfield  {journal} {\bibinfo  {journal} {Progress of Theoretical and Experimental Physics}\ }\textbf {\bibinfo {volume} {2019}},\ \bibinfo {pages} {123C01} (\bibinfo {year} {2019})}\BibitemShut {NoStop}%
\bibitem [{\citenamefont {Vaswani}\ \emph {et~al.}(2017)\citenamefont {Vaswani}, \citenamefont {Shazeer}, \citenamefont {Parmar}, \citenamefont {Uszkoreit}, \citenamefont {Jones}, \citenamefont {Gomez}, \citenamefont {Kaiser},\ and\ \citenamefont {Polosukhin}}]{transformer}%
  \BibitemOpen
  \bibfield  {author} {\bibinfo {author} {\bibfnamefont {A.}~\bibnamefont {Vaswani}}, \bibinfo {author} {\bibfnamefont {N.}~\bibnamefont {Shazeer}}, \bibinfo {author} {\bibfnamefont {N.}~\bibnamefont {Parmar}}, \bibinfo {author} {\bibfnamefont {J.}~\bibnamefont {Uszkoreit}}, \bibinfo {author} {\bibfnamefont {L.}~\bibnamefont {Jones}}, \bibinfo {author} {\bibfnamefont {A.~N.}\ \bibnamefont {Gomez}}, \bibinfo {author} {\bibfnamefont {{\L{}}.}~\bibnamefont {Kaiser}},\ and\ \bibinfo {author} {\bibfnamefont {I.}~\bibnamefont {Polosukhin}},\ }\href@noop {} {\bibinfo {title} {{Attention Is All You Need}}} (\bibinfo {year} {2017}),\ \Eprint {https://arxiv.org/abs/1706.03762} {arXiv:1706.03762} \BibitemShut {NoStop}%
\bibitem [{\citenamefont {Qasim}\ \emph {et~al.}(2025)\citenamefont {Qasim}, \citenamefont {Owen},\ and\ \citenamefont {Serra}}]{instrument}%
  \BibitemOpen
  \bibfield  {author} {\bibinfo {author} {\bibfnamefont {S.~R.}\ \bibnamefont {Qasim}}, \bibinfo {author} {\bibfnamefont {P.}~\bibnamefont {Owen}},\ and\ \bibinfo {author} {\bibfnamefont {N.}~\bibnamefont {Serra}},\ }\href {https://doi.org/10.1088/2632-2153/adf7ff} {\bibfield  {journal} {\bibinfo  {journal} {Mach. Learn.: Sci. Technol.}\ }\textbf {\bibinfo {volume} {6}},\ \bibinfo {pages} {035033} (\bibinfo {year} {2025})}\BibitemShut {NoStop}%
\bibitem [{\citenamefont {V{\aa}ge}(2022)}]{RLTracking}%
  \BibitemOpen
  \bibfield  {author} {\bibinfo {author} {\bibfnamefont {L.~H.}\ \bibnamefont {V{\aa}ge}},\ }in\ \href@noop {} {\emph {\bibinfo {booktitle} {7th Connecting the Dots Workshop (CTD), Princeton, United States}}}\ (\bibinfo {year} {2022})\BibitemShut {NoStop}%
\bibitem [{\citenamefont {Harvey}\ and\ \citenamefont {Lukas}(2021)}]{model_building}%
  \BibitemOpen
  \bibfield  {author} {\bibinfo {author} {\bibfnamefont {T.~R.}\ \bibnamefont {Harvey}}\ and\ \bibinfo {author} {\bibfnamefont {A.}~\bibnamefont {Lukas}},\ }\href@noop {} {\bibinfo {title} {{Particle Physics Model Building with Reinforcement Learning}}} (\bibinfo {year} {2021}),\ \Eprint {https://arxiv.org/abs/2103.04759} {arXiv:2103.04759} \BibitemShut {NoStop}%
\bibitem [{\citenamefont {Wojcik}\ \emph {et~al.}(2024)\citenamefont {Wojcik}, \citenamefont {Eu},\ and\ \citenamefont {Everett}}]{model_building2}%
  \BibitemOpen
  \bibfield  {author} {\bibinfo {author} {\bibfnamefont {G.~N.}\ \bibnamefont {Wojcik}}, \bibinfo {author} {\bibfnamefont {S.~T.}\ \bibnamefont {Eu}},\ and\ \bibinfo {author} {\bibfnamefont {L.~L.}\ \bibnamefont {Everett}},\ }\href@noop {} {\bibinfo {title} {{Towards Beyond Standard Model Model-Building with Reinforcement Learning on Graphs}}} (\bibinfo {year} {2024}),\ \Eprint {https://arxiv.org/abs/2407.07184} {arXiv:2407.07184} \BibitemShut {NoStop}%
\bibitem [{\citenamefont {Nishimura}\ \emph {et~al.}(2024)\citenamefont {Nishimura}, \citenamefont {Miyao},\ and\ \citenamefont {Otsuka}}]{Satsuki1}%
  \BibitemOpen
  \bibfield  {author} {\bibinfo {author} {\bibfnamefont {S.}~\bibnamefont {Nishimura}}, \bibinfo {author} {\bibfnamefont {C.}~\bibnamefont {Miyao}},\ and\ \bibinfo {author} {\bibfnamefont {H.}~\bibnamefont {Otsuka}},\ }\href@noop {} {\bibinfo {title} {{Reinforcement learning-based statistical search strategy for an axion model from flavor}}} (\bibinfo {year} {2024}),\ \Eprint {https://arxiv.org/abs/2409.10023} {arXiv:2409.10023} \BibitemShut {NoStop}%
\bibitem [{\citenamefont {Nishimura}\ \emph {et~al.}(2023)\citenamefont {Nishimura}, \citenamefont {Miyao},\ and\ \citenamefont {Otsuka}}]{Satsuki2}%
  \BibitemOpen
  \bibfield  {author} {\bibinfo {author} {\bibfnamefont {S.}~\bibnamefont {Nishimura}}, \bibinfo {author} {\bibfnamefont {C.}~\bibnamefont {Miyao}},\ and\ \bibinfo {author} {\bibfnamefont {H.}~\bibnamefont {Otsuka}},\ }\href {https://doi.org/10.1007/jhep12(2023)021} {\bibfield  {journal} {\bibinfo  {journal} {Journal of High Energy Physics}\ }\textbf {\bibinfo {volume} {2023}},\ \bibinfo {pages} {21} (\bibinfo {year} {2023})}\BibitemShut {NoStop}%
\bibitem [{\citenamefont {Golutvin}\ \emph {et~al.}(2023)\citenamefont {Golutvin}, \citenamefont {Iniukhin}, \citenamefont {Mauri}, \citenamefont {Owen}, \citenamefont {Serra},\ and\ \citenamefont {Ustyuzhanin}}]{DLAdvocate}%
  \BibitemOpen
  \bibfield  {author} {\bibinfo {author} {\bibfnamefont {A.}~\bibnamefont {Golutvin}}, \bibinfo {author} {\bibfnamefont {A.}~\bibnamefont {Iniukhin}}, \bibinfo {author} {\bibfnamefont {A.}~\bibnamefont {Mauri}}, \bibinfo {author} {\bibfnamefont {P.}~\bibnamefont {Owen}}, \bibinfo {author} {\bibfnamefont {N.}~\bibnamefont {Serra}},\ and\ \bibinfo {author} {\bibfnamefont {A.}~\bibnamefont {Ustyuzhanin}},\ }\href {https://doi.org/10.1140/epjc/s10052-023-11925-w} {\bibfield  {journal} {\bibinfo  {journal} {Eur. Phys. J. C}\ }\textbf {\bibinfo {volume} {83}},\ \bibinfo {pages} {779} (\bibinfo {year} {2023})}\BibitemShut {NoStop}%
\bibitem [{\citenamefont {Silver}\ \emph {et~al.}(2018)\citenamefont {Silver}, \citenamefont {Hubert}, \citenamefont {Schrittwieser}, \citenamefont {Antonoglou}, \citenamefont {Lai}, \citenamefont {Guez}, \citenamefont {Lanctot}, \citenamefont {Sifre}, \citenamefont {Kumaran}, \citenamefont {Graepel}, \citenamefont {Lillicrap}, \citenamefont {Simonyan},\ and\ \citenamefont {Hassabis}}]{AlphaZero}%
  \BibitemOpen
  \bibfield  {author} {\bibinfo {author} {\bibfnamefont {D.}~\bibnamefont {Silver}}, \bibinfo {author} {\bibfnamefont {T.}~\bibnamefont {Hubert}}, \bibinfo {author} {\bibfnamefont {J.}~\bibnamefont {Schrittwieser}}, \bibinfo {author} {\bibfnamefont {I.}~\bibnamefont {Antonoglou}}, \bibinfo {author} {\bibfnamefont {M.}~\bibnamefont {Lai}}, \bibinfo {author} {\bibfnamefont {A.}~\bibnamefont {Guez}}, \bibinfo {author} {\bibfnamefont {M.}~\bibnamefont {Lanctot}}, \bibinfo {author} {\bibfnamefont {L.}~\bibnamefont {Sifre}}, \bibinfo {author} {\bibfnamefont {D.}~\bibnamefont {Kumaran}}, \bibinfo {author} {\bibfnamefont {T.}~\bibnamefont {Graepel}}, \bibinfo {author} {\bibfnamefont {T.}~\bibnamefont {Lillicrap}}, \bibinfo {author} {\bibfnamefont {K.}~\bibnamefont {Simonyan}},\ and\ \bibinfo {author} {\bibfnamefont {D.}~\bibnamefont {Hassabis}},\ }\href {https://doi.org/10.1126/science.aar6404} {\bibfield  {journal} {\bibinfo  {journal} {Science}\ }\textbf {\bibinfo {volume} {362}},\ \bibinfo {pages} {6419} (\bibinfo
  {year} {2018})}\BibitemShut {NoStop}%
\bibitem [{\citenamefont {Coulom}(2007)}]{MCTS1}%
  \BibitemOpen
  \bibfield  {author} {\bibinfo {author} {\bibfnamefont {R.}~\bibnamefont {Coulom}}\ }(\bibinfo  {publisher} {Springer Berlin Heidelberg},\ \bibinfo {address} {Berlin, Heidelberg},\ \bibinfo {year} {2007})\ pp.\ \bibinfo {pages} {72--83}\BibitemShut {NoStop}%
\bibitem [{\citenamefont {Kocsis}\ and\ \citenamefont {Szepesv{\'a}ri}(2006)}]{MCTS2}%
  \BibitemOpen
  \bibfield  {author} {\bibinfo {author} {\bibfnamefont {L.}~\bibnamefont {Kocsis}}\ and\ \bibinfo {author} {\bibfnamefont {C.}~\bibnamefont {Szepesv{\'a}ri}}\ }(\bibinfo  {publisher} {Springer-Verlag},\ \bibinfo {address} {Berlin, Heidelberg},\ \bibinfo {year} {2006})\ pp.\ \bibinfo {pages} {282--29}\BibitemShut {NoStop}%
\bibitem [{\citenamefont {Silver}\ \emph {et~al.}(2016)\citenamefont {Silver}, \citenamefont {Huang}, \citenamefont {Maddison}, \citenamefont {Guez}, \citenamefont {Sifre}, \citenamefont {van~den Driessche}, \citenamefont {Schrittwieser}, \citenamefont {Antonoglou}, \citenamefont {Panneershelvam}, \citenamefont {Lanctot}, \citenamefont {Dieleman}, \citenamefont {Grewe}, \citenamefont {Nham}, \citenamefont {Kalchbrenner}, \citenamefont {Sutskever}, \citenamefont {Lillicrap}, \citenamefont {Leach}, \citenamefont {Kavukcuoglu}, \citenamefont {Graepel},\ and\ \citenamefont {Hassabis}}]{MCTS3}%
  \BibitemOpen
  \bibfield  {author} {\bibinfo {author} {\bibfnamefont {D.}~\bibnamefont {Silver}}, \bibinfo {author} {\bibfnamefont {A.}~\bibnamefont {Huang}}, \bibinfo {author} {\bibfnamefont {C.~J.}\ \bibnamefont {Maddison}}, \bibinfo {author} {\bibfnamefont {A.}~\bibnamefont {Guez}}, \bibinfo {author} {\bibfnamefont {L.}~\bibnamefont {Sifre}}, \bibinfo {author} {\bibfnamefont {G.}~\bibnamefont {van~den Driessche}}, \bibinfo {author} {\bibfnamefont {J.}~\bibnamefont {Schrittwieser}}, \bibinfo {author} {\bibfnamefont {I.}~\bibnamefont {Antonoglou}}, \bibinfo {author} {\bibfnamefont {V.}~\bibnamefont {Panneershelvam}}, \bibinfo {author} {\bibfnamefont {M.}~\bibnamefont {Lanctot}}, \bibinfo {author} {\bibfnamefont {S.}~\bibnamefont {Dieleman}}, \bibinfo {author} {\bibfnamefont {D.}~\bibnamefont {Grewe}}, \bibinfo {author} {\bibfnamefont {J.}~\bibnamefont {Nham}}, \bibinfo {author} {\bibfnamefont {N.}~\bibnamefont {Kalchbrenner}}, \bibinfo {author} {\bibfnamefont {I.}~\bibnamefont {Sutskever}}, \bibinfo {author}
  {\bibfnamefont {T.}~\bibnamefont {Lillicrap}}, \bibinfo {author} {\bibfnamefont {M.}~\bibnamefont {Leach}}, \bibinfo {author} {\bibfnamefont {K.}~\bibnamefont {Kavukcuoglu}}, \bibinfo {author} {\bibfnamefont {T.}~\bibnamefont {Graepel}},\ and\ \bibinfo {author} {\bibfnamefont {D.}~\bibnamefont {Hassabis}},\ }\href {https://doi.org/10.1038/nature16961} {\bibfield  {journal} {\bibinfo  {journal} {Nature}\ }\textbf {\bibinfo {volume} {529}},\ \bibinfo {pages} {484–489} (\bibinfo {year} {2016})}\BibitemShut {NoStop}%
\bibitem [{\citenamefont {Michishita}(2024)}]{symbolic_regression}%
  \BibitemOpen
  \bibfield  {author} {\bibinfo {author} {\bibfnamefont {Y.}~\bibnamefont {Michishita}},\ }\href {https://doi.org/10.7566/JPSJ.93.074005} {\bibfield  {journal} {\bibinfo  {journal} {J. Phys. Soc. Jpn.}\ }\textbf {\bibinfo {volume} {93}},\ \bibinfo {pages} {074005} (\bibinfo {year} {2024})}\BibitemShut {NoStop}%
\bibitem [{\citenamefont {Dalgaard}\ \emph {et~al.}(2020)\citenamefont {Dalgaard}, \citenamefont {Motzoi}, \citenamefont {S{\o}rensen},\ and\ \citenamefont {Sherson}}]{quantum_dynamics}%
  \BibitemOpen
  \bibfield  {author} {\bibinfo {author} {\bibfnamefont {M.}~\bibnamefont {Dalgaard}}, \bibinfo {author} {\bibfnamefont {F.}~\bibnamefont {Motzoi}}, \bibinfo {author} {\bibfnamefont {J.}~\bibnamefont {S{\o}rensen}},\ and\ \bibinfo {author} {\bibfnamefont {J.}~\bibnamefont {Sherson}},\ }\href {https://doi.org/10.1038/s41534-019-0241-0} {\bibfield  {journal} {\bibinfo  {journal} {npj Quantum Information}\ }\textbf {\bibinfo {volume} {6}},\ \bibinfo {pages} {6} (\bibinfo {year} {2020})}\BibitemShut {NoStop}%
\bibitem [{\citenamefont {Radford}\ \emph {et~al.}(2019)\citenamefont {Radford}, \citenamefont {Wu}, \citenamefont {Child}, \citenamefont {Luan}, \citenamefont {Amodei},\ and\ \citenamefont {Sutskever}}]{GPT2}%
  \BibitemOpen
  \bibfield  {author} {\bibinfo {author} {\bibfnamefont {A.}~\bibnamefont {Radford}}, \bibinfo {author} {\bibfnamefont {J.}~\bibnamefont {Wu}}, \bibinfo {author} {\bibfnamefont {R.}~\bibnamefont {Child}}, \bibinfo {author} {\bibfnamefont {D.}~\bibnamefont {Luan}}, \bibinfo {author} {\bibfnamefont {D.}~\bibnamefont {Amodei}},\ and\ \bibinfo {author} {\bibfnamefont {I.}~\bibnamefont {Sutskever}},\ }\href@noop {} {\bibfield  {journal} {\bibinfo  {journal} {OpenAI}\ }\textbf {\bibinfo {volume} {1}},\ \bibinfo {pages} {8} (\bibinfo {year} {2019})}\BibitemShut {NoStop}%
\bibitem [{\citenamefont {Brown}\ \emph {et~al.}(2020)\citenamefont {Brown}, \citenamefont {Mann}, \citenamefont {Ryder}, \citenamefont {Subbiah}, \citenamefont {Kaplan}, \citenamefont {Dhariwal}, \citenamefont {Neelakantan}, \citenamefont {Shyam}, \citenamefont {Sastry}, \citenamefont {Askell}, \citenamefont {Agarwal}, \citenamefont {Herbert-Voss}, \citenamefont {Krueger}, \citenamefont {Henighan}, \citenamefont {Child}, \citenamefont {Ramesh}, \citenamefont {Ziegler}, \citenamefont {Wu}, \citenamefont {Winter}, \citenamefont {Hesse}, \citenamefont {Chen}, \citenamefont {Sigler}, \citenamefont {Litwin}, \citenamefont {Gray}, \citenamefont {Chess}, \citenamefont {Clark}, \citenamefont {Berner}, \citenamefont {McCandlish}, \citenamefont {Radford}, \citenamefont {Sutskever},\ and\ \citenamefont {Amodei}}]{GPT3}%
  \BibitemOpen
  \bibfield  {author} {\bibinfo {author} {\bibfnamefont {T.}~\bibnamefont {Brown}}, \bibinfo {author} {\bibfnamefont {B.}~\bibnamefont {Mann}}, \bibinfo {author} {\bibfnamefont {N.}~\bibnamefont {Ryder}}, \bibinfo {author} {\bibfnamefont {M.}~\bibnamefont {Subbiah}}, \bibinfo {author} {\bibfnamefont {J.~D.}\ \bibnamefont {Kaplan}}, \bibinfo {author} {\bibfnamefont {P.}~\bibnamefont {Dhariwal}}, \bibinfo {author} {\bibfnamefont {A.}~\bibnamefont {Neelakantan}}, \bibinfo {author} {\bibfnamefont {P.}~\bibnamefont {Shyam}}, \bibinfo {author} {\bibfnamefont {G.}~\bibnamefont {Sastry}}, \bibinfo {author} {\bibfnamefont {A.}~\bibnamefont {Askell}}, \bibinfo {author} {\bibfnamefont {S.}~\bibnamefont {Agarwal}}, \bibinfo {author} {\bibfnamefont {A.}~\bibnamefont {Herbert-Voss}}, \bibinfo {author} {\bibfnamefont {G.}~\bibnamefont {Krueger}}, \bibinfo {author} {\bibfnamefont {T.}~\bibnamefont {Henighan}}, \bibinfo {author} {\bibfnamefont {R.}~\bibnamefont {Child}}, \bibinfo {author} {\bibfnamefont {A.}~\bibnamefont
  {Ramesh}}, \bibinfo {author} {\bibfnamefont {D.}~\bibnamefont {Ziegler}}, \bibinfo {author} {\bibfnamefont {J.}~\bibnamefont {Wu}}, \bibinfo {author} {\bibfnamefont {C.}~\bibnamefont {Winter}}, \bibinfo {author} {\bibfnamefont {C.}~\bibnamefont {Hesse}}, \bibinfo {author} {\bibfnamefont {M.}~\bibnamefont {Chen}}, \bibinfo {author} {\bibfnamefont {E.}~\bibnamefont {Sigler}}, \bibinfo {author} {\bibfnamefont {M.}~\bibnamefont {Litwin}}, \bibinfo {author} {\bibfnamefont {S.}~\bibnamefont {Gray}}, \bibinfo {author} {\bibfnamefont {B.}~\bibnamefont {Chess}}, \bibinfo {author} {\bibfnamefont {J.}~\bibnamefont {Clark}}, \bibinfo {author} {\bibfnamefont {C.}~\bibnamefont {Berner}}, \bibinfo {author} {\bibfnamefont {S.}~\bibnamefont {McCandlish}}, \bibinfo {author} {\bibfnamefont {A.}~\bibnamefont {Radford}}, \bibinfo {author} {\bibfnamefont {I.}~\bibnamefont {Sutskever}},\ and\ \bibinfo {author} {\bibfnamefont {D.}~\bibnamefont {Amodei}},\ }in\ \href@noop {} {\emph {\bibinfo {booktitle} {Proceedings of the \nth{33}
  Annual Conference on Neural Information Processing Systems}}}\ (\bibinfo {year} {2020})\BibitemShut {NoStop}%
\bibitem [{\citenamefont {Qu}\ \emph {et~al.}(2022)\citenamefont {Qu}, \citenamefont {Li},\ and\ \citenamefont {Qian}}]{tagging1}%
  \BibitemOpen
  \bibfield  {author} {\bibinfo {author} {\bibfnamefont {H.}~\bibnamefont {Qu}}, \bibinfo {author} {\bibfnamefont {C.}~\bibnamefont {Li}},\ and\ \bibinfo {author} {\bibfnamefont {S.}~\bibnamefont {Qian}},\ }in\ \href@noop {} {\emph {\bibinfo {booktitle} {Proceedings of the \nth{39} International Conference on Machine Learning, Baltimore, Maryland, USA}}}\ (\bibinfo {year} {2022})\BibitemShut {NoStop}%
\bibitem [{\citenamefont {Wu}\ \emph {et~al.}(2025)\citenamefont {Wu}, \citenamefont {Wang}, \citenamefont {Li}, \citenamefont {Qu},\ and\ \citenamefont {Zhu}}]{tagging2}%
  \BibitemOpen
  \bibfield  {author} {\bibinfo {author} {\bibfnamefont {Y.}~\bibnamefont {Wu}}, \bibinfo {author} {\bibfnamefont {K.}~\bibnamefont {Wang}}, \bibinfo {author} {\bibfnamefont {C.}~\bibnamefont {Li}}, \bibinfo {author} {\bibfnamefont {H.}~\bibnamefont {Qu}},\ and\ \bibinfo {author} {\bibfnamefont {J.}~\bibnamefont {Zhu}},\ }\href {https://doi.org/10.1088/1674-1137/ad7f3d} {\bibfield  {journal} {\bibinfo  {journal} {Chinese Physics C}\ }\textbf {\bibinfo {volume} {49}},\ \bibinfo {pages} {1} (\bibinfo {year} {2025})}\BibitemShut {NoStop}%
\bibitem [{\citenamefont {Stroud}\ \emph {et~al.}(2024)\citenamefont {Stroud}, \citenamefont {Duckett}, \citenamefont {Hart}, \citenamefont {Pond}, \citenamefont {Rettie}, \citenamefont {Facini},\ and\ \citenamefont {Scanlon}}]{track_reconstruction}%
  \BibitemOpen
  \bibfield  {author} {\bibinfo {author} {\bibfnamefont {S.~V.}\ \bibnamefont {Stroud}}, \bibinfo {author} {\bibfnamefont {P.}~\bibnamefont {Duckett}}, \bibinfo {author} {\bibfnamefont {M.}~\bibnamefont {Hart}}, \bibinfo {author} {\bibfnamefont {N.}~\bibnamefont {Pond}}, \bibinfo {author} {\bibfnamefont {S.}~\bibnamefont {Rettie}}, \bibinfo {author} {\bibfnamefont {G.}~\bibnamefont {Facini}},\ and\ \bibinfo {author} {\bibfnamefont {T.}~\bibnamefont {Scanlon}},\ }\href@noop {} {\bibinfo {title} {{Transformers for Charged Particle Track Reconstruction in High Energy Physics}}} (\bibinfo {year} {2024}),\ \Eprint {https://arxiv.org/abs/2411.07149} {arXiv:2411.07149} \BibitemShut {NoStop}%
\bibitem [{\citenamefont {Caron}\ \emph {et~al.}(2025)\citenamefont {Caron}, \citenamefont {Dobreva}, \citenamefont {Ferrer~Sánchez}, \citenamefont {Martín-Guerrero}, \citenamefont {Odyurt}, \citenamefont {Ruiz~de Austri~Bazan}, \citenamefont {Wolffs},\ and\ \citenamefont {Zhao}}]{track_reconstruction2}%
  \BibitemOpen
  \bibfield  {author} {\bibinfo {author} {\bibfnamefont {S.}~\bibnamefont {Caron}}, \bibinfo {author} {\bibfnamefont {N.}~\bibnamefont {Dobreva}}, \bibinfo {author} {\bibfnamefont {A.}~\bibnamefont {Ferrer~Sánchez}}, \bibinfo {author} {\bibfnamefont {J.~D.}\ \bibnamefont {Martín-Guerrero}}, \bibinfo {author} {\bibfnamefont {U.}~\bibnamefont {Odyurt}}, \bibinfo {author} {\bibfnamefont {R.}~\bibnamefont {Ruiz~de Austri~Bazan}}, \bibinfo {author} {\bibfnamefont {Z.}~\bibnamefont {Wolffs}},\ and\ \bibinfo {author} {\bibfnamefont {Y.}~\bibnamefont {Zhao}},\ }\href {https://doi.org/10.1140/epjc/s10052-025-14156-3} {\bibfield  {journal} {\bibinfo  {journal} {Eur. Phys. J. C}\ }\textbf {\bibinfo {volume} {85}},\ \bibinfo {pages} {460} (\bibinfo {year} {2025})}\BibitemShut {NoStop}%
\bibitem [{\citenamefont {Kasper}\ \emph {et~al.}(2025)\citenamefont {Kasper}, \citenamefont {Wrońska}, \citenamefont {Awal}, \citenamefont {Hetzel}, \citenamefont {Kołodziej}, \citenamefont {Rusiecka}, \citenamefont {Stahl},\ and\ \citenamefont {Wong}}]{detection_setup_optimisation}%
  \BibitemOpen
  \bibfield  {author} {\bibinfo {author} {\bibfnamefont {J.}~\bibnamefont {Kasper}}, \bibinfo {author} {\bibfnamefont {A.}~\bibnamefont {Wrońska}}, \bibinfo {author} {\bibfnamefont {A.}~\bibnamefont {Awal}}, \bibinfo {author} {\bibfnamefont {R.}~\bibnamefont {Hetzel}}, \bibinfo {author} {\bibfnamefont {M.}~\bibnamefont {Kołodziej}}, \bibinfo {author} {\bibfnamefont {K.}~\bibnamefont {Rusiecka}}, \bibinfo {author} {\bibfnamefont {A.}~\bibnamefont {Stahl}},\ and\ \bibinfo {author} {\bibfnamefont {M.-L.}\ \bibnamefont {Wong}},\ }\href {https://doi.org/10.1088/1361-6560/adec39} {\bibfield  {journal} {\bibinfo  {journal} {Phys. Med. Biol.}\ }\textbf {\bibinfo {volume} {70}},\ \bibinfo {pages} {145015} (\bibinfo {year} {2025})}\BibitemShut {NoStop}%
\bibitem [{\citenamefont {Wessen}\ and\ \citenamefont {Camargo-Molina}(2024)}]{efficient_exploration}%
  \BibitemOpen
  \bibfield  {author} {\bibinfo {author} {\bibfnamefont {J.}~\bibnamefont {Wessen}}\ and\ \bibinfo {author} {\bibfnamefont {E.}~\bibnamefont {Camargo-Molina}},\ }\href@noop {} {\bibinfo {title} {{A diversity-enhanced genetic algorithm for efficient exploration of parameter spaces}}} (\bibinfo {year} {2024}),\ \Eprint {https://arxiv.org/abs/2412.17104v1} {arXiv:2412.17104v1} \BibitemShut {NoStop}%
\bibitem [{\citenamefont {Abel}\ \emph {et~al.}(2024)\citenamefont {Abel}, \citenamefont {Constantin}, \citenamefont {Harvey}, \citenamefont {Lukas},\ and\ \citenamefont {Nutricati}}]{efficient_exploration2}%
  \BibitemOpen
  \bibfield  {author} {\bibinfo {author} {\bibfnamefont {S.~A.}\ \bibnamefont {Abel}}, \bibinfo {author} {\bibfnamefont {A.}~\bibnamefont {Constantin}}, \bibinfo {author} {\bibfnamefont {T.~R.}\ \bibnamefont {Harvey}}, \bibinfo {author} {\bibfnamefont {A.}~\bibnamefont {Lukas}},\ and\ \bibinfo {author} {\bibfnamefont {L.~A.}\ \bibnamefont {Nutricati}},\ }\href {https://doi.org/10.1002/prop.202300260} {\bibfield  {journal} {\bibinfo  {journal} {Fortschr. Phys.}\ }\textbf {\bibinfo {volume} {72}},\ \bibinfo {pages} {2300260} (\bibinfo {year} {2024})}\BibitemShut {NoStop}%
\bibitem [{\citenamefont {Luo}\ \emph {et~al.}(2020)\citenamefont {Luo}, \citenamefont {Feng},\ and\ \citenamefont {Zhang}}]{efficient_exploration3}%
  \BibitemOpen
  \bibfield  {author} {\bibinfo {author} {\bibfnamefont {X.-L.}\ \bibnamefont {Luo}}, \bibinfo {author} {\bibfnamefont {J.}~\bibnamefont {Feng}},\ and\ \bibinfo {author} {\bibfnamefont {H.-H.}\ \bibnamefont {Zhang}},\ }\href {https://doi.org/10.1016/j.cpc.2019.06.008} {\bibfield  {journal} {\bibinfo  {journal} {Computer Physics Communications}\ }\textbf {\bibinfo {volume} {250}},\ \bibinfo {pages} {106818} (\bibinfo {year} {2020})}\BibitemShut {NoStop}%
\bibitem [{\citenamefont {Khadka}\ and\ \citenamefont {Tumer}(2018)}]{ERL}%
  \BibitemOpen
  \bibfield  {author} {\bibinfo {author} {\bibfnamefont {S.}~\bibnamefont {Khadka}}\ and\ \bibinfo {author} {\bibfnamefont {K.}~\bibnamefont {Tumer}},\ }in\ \href@noop {} {\emph {\bibinfo {booktitle} {\nth{32} Conference on Neural Information Processing Systems (NeurIPS), Montréal, Canada}}}\ (\bibinfo {year} {2018})\BibitemShut {NoStop}%
\bibitem [{\citenamefont {Such}\ \emph {et~al.}(2018)\citenamefont {Such}, \citenamefont {Madhavan}, \citenamefont {Conti}, \citenamefont {Lehman}, \citenamefont {Stanley},\ and\ \citenamefont {Clune}}]{DeepNeuroevolution}%
  \BibitemOpen
  \bibfield  {author} {\bibinfo {author} {\bibfnamefont {F.~P.}\ \bibnamefont {Such}}, \bibinfo {author} {\bibfnamefont {V.}~\bibnamefont {Madhavan}}, \bibinfo {author} {\bibfnamefont {E.}~\bibnamefont {Conti}}, \bibinfo {author} {\bibfnamefont {J.}~\bibnamefont {Lehman}}, \bibinfo {author} {\bibfnamefont {K.~O.}\ \bibnamefont {Stanley}},\ and\ \bibinfo {author} {\bibfnamefont {J.}~\bibnamefont {Clune}},\ }\href@noop {} {\bibinfo {title} {{Deep Neuroevolution: Genetic Algorithms Are a Competitive Alternative for Training Deep Neural Networks for Reinforcement Learning}}} (\bibinfo {year} {2018}),\ \Eprint {https://arxiv.org/abs/1712.06567} {arXiv:1712.06567} \BibitemShut {NoStop}%
\bibitem [{\citenamefont {Ecoffet}\ \emph {et~al.}(2021)\citenamefont {Ecoffet}, \citenamefont {Huizinga}, \citenamefont {Lehman}, \citenamefont {Stanley},\ and\ \citenamefont {Clune}}]{Go_Explore}%
  \BibitemOpen
  \bibfield  {author} {\bibinfo {author} {\bibfnamefont {A.}~\bibnamefont {Ecoffet}}, \bibinfo {author} {\bibfnamefont {J.}~\bibnamefont {Huizinga}}, \bibinfo {author} {\bibfnamefont {J.}~\bibnamefont {Lehman}}, \bibinfo {author} {\bibfnamefont {K.~O.}\ \bibnamefont {Stanley}},\ and\ \bibinfo {author} {\bibfnamefont {J.}~\bibnamefont {Clune}},\ }\href {https://doi.org/10.1038/s41586-020-03157-9} {\bibfield  {journal} {\bibinfo  {journal} {Nature}\ }\textbf {\bibinfo {volume} {590}},\ \bibinfo {pages} {580–586} (\bibinfo {year} {2021})}\BibitemShut {NoStop}%
\bibitem [{\citenamefont {Zare}\ \emph {et~al.}(2024)\citenamefont {Zare}, \citenamefont {Kebria}, \citenamefont {Khosravi},\ and\ \citenamefont {Nahavandi}}]{ImitationLearning}%
  \BibitemOpen
  \bibfield  {author} {\bibinfo {author} {\bibfnamefont {M.}~\bibnamefont {Zare}}, \bibinfo {author} {\bibfnamefont {P.~M.}\ \bibnamefont {Kebria}}, \bibinfo {author} {\bibfnamefont {A.}~\bibnamefont {Khosravi}},\ and\ \bibinfo {author} {\bibfnamefont {S.}~\bibnamefont {Nahavandi}},\ }\href {https://doi.org/10.1109/TCYB.2024.3395626} {\bibfield  {journal} {\bibinfo  {journal} {IEEE Transactions on Cybernetics}\ }\textbf {\bibinfo {volume} {54}},\ \bibinfo {pages} {12} (\bibinfo {year} {2024})}\BibitemShut {NoStop}%
\bibitem [{\citenamefont {Salimans}\ and\ \citenamefont {Chen}(2018)}]{backward_algorithm}%
  \BibitemOpen
  \bibfield  {author} {\bibinfo {author} {\bibfnamefont {T.}~\bibnamefont {Salimans}}\ and\ \bibinfo {author} {\bibfnamefont {R.~J.}\ \bibnamefont {Chen}},\ }\href@noop {} {\bibinfo {title} {{Learning Montezuma's Revenge from a Single Demonstration}}} (\bibinfo {year} {2018}),\ \Eprint {https://arxiv.org/abs/1812.03381} {arXiv:1812.03381} \BibitemShut {NoStop}%
\bibitem [{\citenamefont {Gu}\ \emph {et~al.}(2025)\citenamefont {Gu}, \citenamefont {Li}, \citenamefont {Xing}, \citenamefont {Zhang},\ and\ \citenamefont {Cheng}}]{GA_RL_paper2}%
  \BibitemOpen
  \bibfield  {author} {\bibinfo {author} {\bibfnamefont {S.}~\bibnamefont {Gu}}, \bibinfo {author} {\bibfnamefont {K.}~\bibnamefont {Li}}, \bibinfo {author} {\bibfnamefont {J.}~\bibnamefont {Xing}}, \bibinfo {author} {\bibfnamefont {Y.}~\bibnamefont {Zhang}},\ and\ \bibinfo {author} {\bibfnamefont {J.}~\bibnamefont {Cheng}},\ }\href@noop {} {\bibinfo {title} {{Synergizing Reinforcement Learning and Genetic Algorithms for Neural Combinatorial Optimization}}} (\bibinfo {year} {2025}),\ \Eprint {https://arxiv.org/abs/2506.09404} {arXiv:2506.09404} \BibitemShut {NoStop}%
\bibitem [{\citenamefont {Maus}\ \emph {et~al.}(2025)\citenamefont {Maus}, \citenamefont {Atamna},\ and\ \citenamefont {Glasmachers}}]{GA_RL_paper}%
  \BibitemOpen
  \bibfield  {author} {\bibinfo {author} {\bibfnamefont {T.}~\bibnamefont {Maus}}, \bibinfo {author} {\bibfnamefont {A.}~\bibnamefont {Atamna}},\ and\ \bibinfo {author} {\bibfnamefont {T.}~\bibnamefont {Glasmachers}},\ }\href@noop {} {\bibinfo {title} {{Leveraging Genetic Algorithms for Efficient Demonstration Generation in Real-World Reinforcement Learning Environments}}} (\bibinfo {year} {2025}),\ \Eprint {https://arxiv.org/abs/2507.00762} {arXiv:2507.00762} \BibitemShut {NoStop}%
\bibitem [{\citenamefont {Navas}\ \emph {et~al.}(2024)\citenamefont {Navas} \emph {et~al.}}]{PDG}%
  \BibitemOpen
  \bibfield  {author} {\bibinfo {author} {\bibfnamefont {S.}~\bibnamefont {Navas}} \emph {et~al.},\ }\href {https://doi.org/10.1103/PhysRevD.110.030001} {\bibfield  {journal} {\bibinfo  {journal} {Phys. Rev. D}\ }\textbf {\bibinfo {volume} {110}},\ \bibinfo {pages} {030001} (\bibinfo {year} {2024})}\BibitemShut {NoStop}%
\bibitem [{\citenamefont {Hijano}\ \emph {et~al.}(2025)\citenamefont {Hijano}, \citenamefont {Lancierini}, \citenamefont {Marshall}, \citenamefont {Mauri}, \citenamefont {Owen}, \citenamefont {Patel}, \citenamefont {Petridis}, \citenamefont {Qasim}, \citenamefont {Serra}, \citenamefont {Sutcliffe},\ and\ \citenamefont {Tilquin}}]{Heterogeneous_GNNs}%
  \BibitemOpen
  \bibfield  {author} {\bibinfo {author} {\bibfnamefont {G.}~\bibnamefont {Hijano}}, \bibinfo {author} {\bibfnamefont {D.}~\bibnamefont {Lancierini}}, \bibinfo {author} {\bibfnamefont {A.~M.}\ \bibnamefont {Marshall}}, \bibinfo {author} {\bibfnamefont {A.}~\bibnamefont {Mauri}}, \bibinfo {author} {\bibfnamefont {P.}~\bibnamefont {Owen}}, \bibinfo {author} {\bibfnamefont {M.}~\bibnamefont {Patel}}, \bibinfo {author} {\bibfnamefont {K.}~\bibnamefont {Petridis}}, \bibinfo {author} {\bibfnamefont {S.~R.}\ \bibnamefont {Qasim}}, \bibinfo {author} {\bibfnamefont {N.}~\bibnamefont {Serra}}, \bibinfo {author} {\bibfnamefont {W.}~\bibnamefont {Sutcliffe}},\ and\ \bibinfo {author} {\bibfnamefont {H.}~\bibnamefont {Tilquin}},\ }\href@noop {} {\bibinfo {title} {{Replacing detector simulation with heterogeneous GNNs in flavour physics analyses}}} (\bibinfo {year} {2025}),\ \Eprint {https://arxiv.org/abs/2507.05069} {arXiv:2507.05069} \BibitemShut {NoStop}%
\bibitem [{\citenamefont {Förster}(2023)}]{AlphaZero_implementation}%
  \BibitemOpen
  \bibfield  {author} {\bibinfo {author} {\bibfnamefont {R.}~\bibnamefont {Förster}},\ }\href@noop {} {\bibinfo {title} {{AlphaZeroFromScratch}}},\ \bibinfo {howpublished} {\url{https://github.com/foersterrobert/AlphaZeroFromScratch}} (\bibinfo {year} {2023}),\ \bibinfo {note} {gitHub repository}\BibitemShut {NoStop}%
\bibitem [{\citenamefont {McInnes}\ and\ \citenamefont {Healy}(2024)}]{UMAP}%
  \BibitemOpen
  \bibfield  {author} {\bibinfo {author} {\bibfnamefont {L.}~\bibnamefont {McInnes}}\ and\ \bibinfo {author} {\bibfnamefont {J.}~\bibnamefont {Healy}},\ }\href {https://doi.org/10.1038/s43586-024-00363-x} {\bibfield  {journal} {\bibinfo  {journal} {Nat Rev Methods Primers}\ }\textbf {\bibinfo {volume} {4}},\ \bibinfo {pages} {82} (\bibinfo {year} {2024})}\BibitemShut {NoStop}%
\bibitem [{\citenamefont {Rodrigues}\ and\ \citenamefont {Schreiner}(2023)}]{DecayLanguage}%
  \BibitemOpen
  \bibfield  {author} {\bibinfo {author} {\bibfnamefont {E.}~\bibnamefont {Rodrigues}}\ and\ \bibinfo {author} {\bibfnamefont {H.}~\bibnamefont {Schreiner}},\ }\href {https://doi.org/10.5281/zenodo.7804926} {\bibinfo {title} {{DecayLanguage (v0.15.3)}}} (\bibinfo {year} {2023}),\ \bibinfo {note} {{Zenodo}. Available at: \url{https://doi.org/10.5281/zenodo.7804926}}\BibitemShut {NoStop}%
\end{thebibliography}
\end{document}